\documentclass[sn-mathphys]{sn-jnl}

\usepackage{tabularx}
\usepackage{graphicx}                                                           
\usepackage{float} 
\usepackage{booktabs}
\usepackage{makecell}

\usepackage{wrapfig}
\usepackage{bbding}
\usepackage{pifont}
\usepackage{expl3}

\hypersetup{
  colorlinks=true,
  linkcolor=blue,
  citecolor=blue,
  urlcolor=cyan
  }

\ExplSyntaxOn

\NewDocumentCommand{\circledstep}{O{RoyalBlue} m}{
  \cs_if_exist:cF { c@circstep_#1 } {
    \newcounter{circstep_#1}
  }
  \refstepcounter{circstep_#1}
  \textcolor{#1}{\ding{\numexpr171+\value{circstep_#1}\relax}}%
  \label{#2}
}

\NewDocumentCommand{\refcircstep}{O{RoyalBlue} m}{
  \hyperref[#2]{\textcolor{#1}{\ding{\numexpr171+\getrefnumber{#2}\relax}}}
}

\ExplSyntaxOff

\jyear{2021}%

\theoremstyle{thmstyleone}%
\theoremstyle{thmstyletwo}%

\theoremstyle{thmstylethree}%

\begin{document}

\title[Challenges and Trends in Egocentric Vision: A Survey]{Challenges and Trends in Egocentric Vision: A Survey}


\author[1]{\fnm{Xiang} \sur{Li}}
\author[1]{\fnm{Heqian} \sur{Qiu}}
\author[1]{\fnm{Lanxiao} \sur{Wang}}
\author[1]{\fnm{Hanwen} \sur{Zhang}}

\author[1]{\fnm{Chenghao} \sur{Qi}}
\author[1]{\fnm{Linfeng} \sur{Han}}
\author[1]{\fnm{Huiyu} \sur{Xiong}}

\author[1]{\fnm{Hongliang} \sur{Li}}

\affil[1]{\orgdiv{Department of Information and Communication Engineering}, \orgname{University of Electronic Science and Technology of China}, \orgaddress{\city{Chengdu} \postcode{611731},  \country{China}}}


\abstract{With the rapid development of artificial intelligence technologies and wearable devices, egocentric vision understanding has emerged as a new and challenging research direction, gradually attracting widespread attention from both academia and industry. Egocentric vision captures visual and multimodal data through cameras or sensors worn on the human body, offering a unique perspective that simulates human visual experiences. This paper provides a comprehensive survey of the research on egocentric vision understanding, systematically analyzing the components of egocentric scenes and categorizing the tasks into four main areas: subject understanding, object understanding, environment understanding, and hybrid understanding. We explore in detail the sub-tasks within each category. We also summarize the main challenges and trends currently existing in the field. Furthermore, this paper presents an overview of high-quality egocentric vision datasets, offering valuable resources for future research. By summarizing the latest advancements, we anticipate the broad applications of egocentric vision technologies in fields such as augmented reality, virtual reality, and embodied intelligence, and propose future research directions based on the latest developments in the field.}


\maketitle
\section{Introduction}\label{Introduction}

\begin{figure}[ht]%
  \centering
  \includegraphics[width=0.5\textwidth]{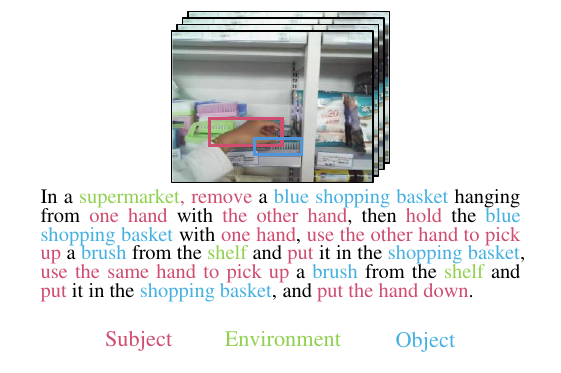}
  \caption{This figure illustrates an example of a egocentric scene, presenting a segment of video along with corresponding text. The red text denotes descriptions related to the subject, primarily focusing on actions. The blue text indicates the objects with which the subject interacts. The green text describes the environment in which the video is set.}\label{example}
\end{figure}

With the rapid advancement of artificial intelligence technologies, the field of computer vision has witnessed remarkable progress in recent years. Among these developments, egocentric visual understanding, as an emerging and challenging research direction, has increasingly garnered widespread attention from both academia and industry. Egocentric vision, derived from visual data captured by cameras or sensors worn on the human body, offers machines a unique perspective that emulates the human visual experience, an example shown in Fig. \ref{example}. Data obtained from this viewpoint exhibits a high degree of dynamism and interactivity, enabling the provision of richer and more authentic environmental information for a wide range of intelligent applications.

The study of egocentric visual understanding holds significant theoretical importance and broad application prospects. From a theoretical standpoint, it encompasses the interdisciplinary integration of computer vision, machine learning, and cognitive science, offering novel perspectives and methodologies for exploring human visual cognition mechanisms and advancing machine intelligence. In practical applications, egocentric visual understanding technology finds extensive use in domains such as augmented reality (AR), virtual reality (VR), intelligent surveillance, human-computer interaction, and robotics, thereby delivering substantial convenience to people’s lives and work.

\begin{figure}[ht]%
  \centering
  \includegraphics[width=0.9\textwidth]{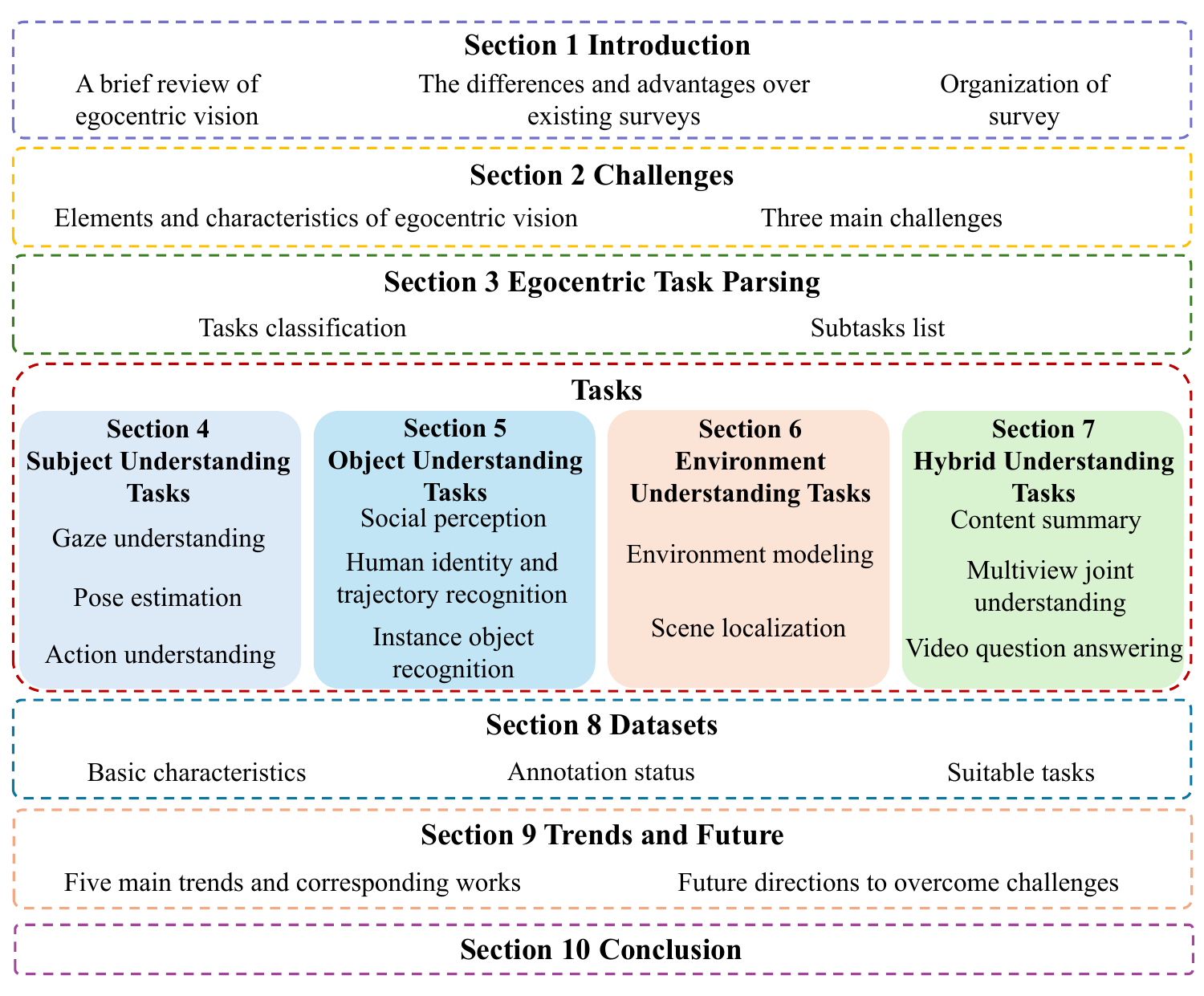}
  \caption{The structure of this paper is as follows. Nine sections are included. In the sect. \ref{Introduction}, egocentric vision is briefly reviewed, the differences between this survey and other existing surveys are highlighted, and the organizational structure is outlined. In sect. \ref{Challenges}, the challenges faced by recent work summarized. Sect. \ref{Egocentric Task Decomposition} breaks down egocentric scenario tasks into three key elements and characterizes each element’s properties. The subtasks are categorized into four major classes: subject, object, environment, and hybrid. In sect. \ref{Subject Understanding Related Tasks}, \ref{Object Understanding Related Tasks}, \ref{Environment Understanding Related Tasks}, and \ref{Hybrid Understanding Related Tasks}, the relevant subtasks under each major class are investigated. Sect. \ref{Datasets} provides a detailed summary of the datasets available for egocentric vision. In sect. \ref{Trends and Future}, major trends in recent works are concluded. Finally, in sect. \ref{Conclusion}, the contributions of this work are summarized, and insights into the future development of egocentric vision are offered.}\label{content}
\end{figure}

To better support future work and research in the egocentric domain, we conducted a comprehensive survey of recent related works mainly after 2020 in this field. Unlike previous studies, our review explicitly highlights the analysis of current challenges(sect. \ref{Challenges}), rooted in the complex nature of egocentric vision and pressing technological bottlenecks and emerging trends(sect. \ref{Trends and Future}), which represent valuable directions developed in recent work to tackle these challenges and advance the field. Betancourt et al. \cite{betancourt_evolution_2015} explored the development, characteristics, and challenges of early first-person vision (FPV) video analysis. Del Molino et al. \cite{molino_summarization_2016} summarized the literature and methodologies in the field of egocentric summarization. Rodin et al. \cite{rodin_predicting_2021} investigated various prediction tasks in egocentric vision, providing an overview of applications, devices, existing issues, datasets, and models. Núñez-Marcos et al. \cite{nunez-marcos_egocentric_2022} reviewed diverse methods for egocentric action recognition and proposed a detailed classification of available approaches. Bandini et al. \cite{bandini_analysis_2023} examined literature in egocentric vision focused on hands, with an emphasis on summarizing methods and datasets. Azam et al. \cite{azam_survey_2024} aimed to deliver a comprehensive overview of the current state of egocentric pose estimation research. Fan et al. \cite{fan_benchmarks_2024} surveyed benchmarks and methods related to egocentric hand-object interaction, outlining the challenges faced in this area. Plizzari et al. \cite{plizzari_outlook_2024} envisioned the future applications of egocentric technology across various scenarios through role-based narratives, followed by a survey of related work on egocentric tasks. In contrast to these existing studies, a key novelty of our paper lies in our systematic decomposition of egocentric scenarios. We categorize all tasks into four major classes, with each major class further subdivided into several subclasses based on the differing objectives of the tasks. As far as we know, this paper is the first survey to provide a hierarchical analysis of egocentric scenarios. The structure of this paper is outlined in Fig. \ref{content}. 

\section{Challenges}\label{Challenges}

In this section, we first analyze the inherent understanding challenges associated with each core component of egocentric vision, including the subject, interacting objects, and surrounding environment, based on their unique characteristics in egocentric scenarios. We then summarize several overarching technical challenges that have emerged as bottlenecks in the current development of egocentric vision methods.

In egocentric vision, the camera is typically attached to a specific subject—usually a human—leading to unique challenges and opportunities in visual analysis. The subject exhibits three notable characteristics: \textbf{(1) Rich Semantic Signals} across multiple modalities (e.g., actions, language, gaze, emotion); \textbf{(2) Frequent Viewpoint Changes} due to head/body motion, often causing occlusions; and \textbf{(3) Strong Personalization}, as individuals may perform similar actions differently.

Besides the subject, \textbf{interactive objects} play a critical role. These may be humans, animals, or physical items, characterized by \textbf{(1) Dynamic Visual Presence} affected by camera motion, \textbf{(2) Frequent Occlusions and Reappearances}, and \textbf{(3) Strong Subject-Object Interactivity}, which provides additional cues for understanding the subject's behavior.

The \textbf{environment} forms the spatial context for interaction, typically comprising semi-static structures (e.g., furniture, roads, buildings). It contributes via \textbf{(1) Background Stability} as a spatial reference, \textbf{(2) High Visual Dynamics} due to camera movement causing fragmented scene unfolding, and \textbf{(3) Behavioral Cues} such as doors or hallways that indicate possible intentions.

\begin{figure}[ht]%
  \centering
  \includegraphics[width=0.9\textwidth]{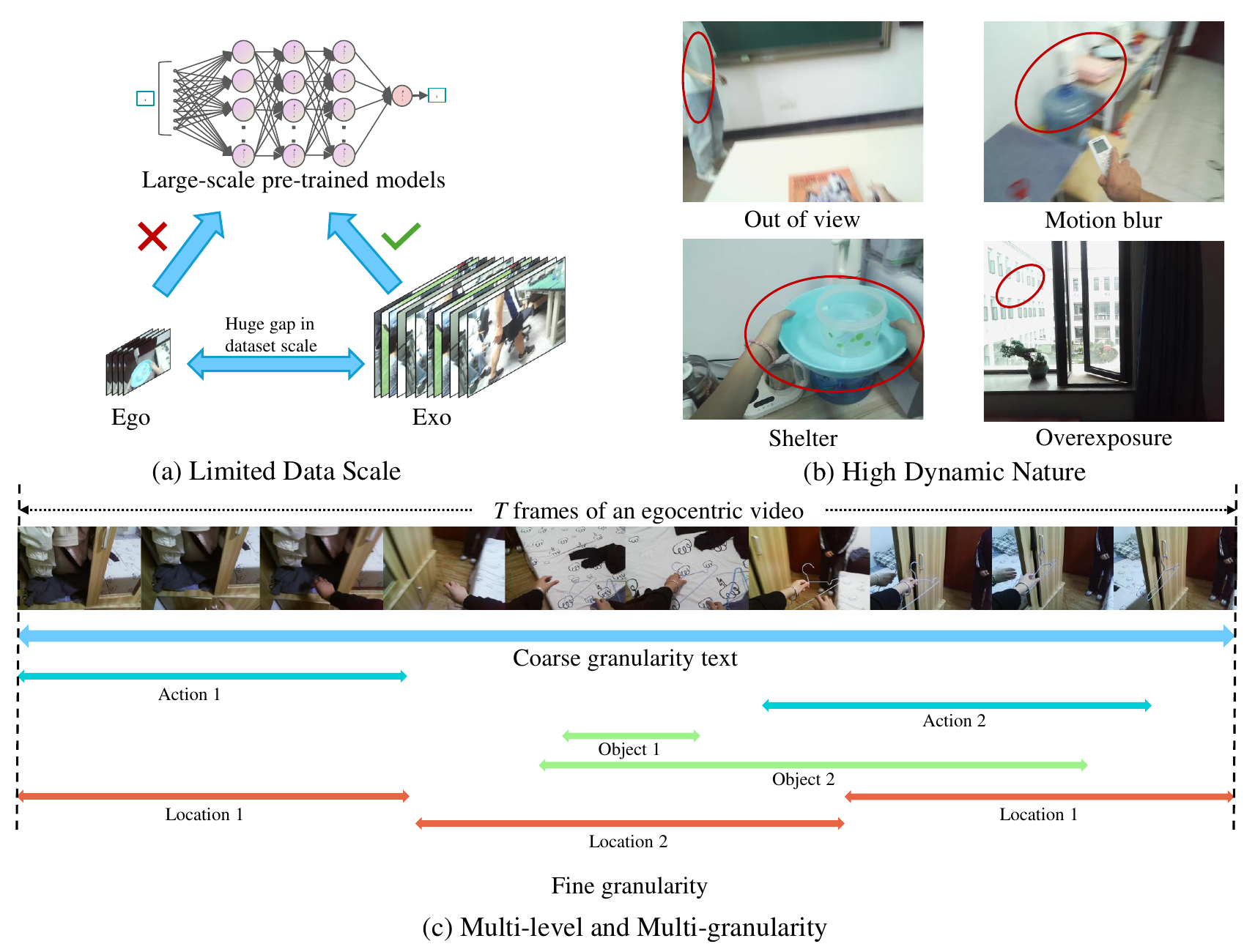}
  \caption{Three major challenges are currently faced in the field of egocentric vision. (a) Compared to exocentric vision, there is a lack of large-scale specialized datasets and high-quality evaluation benchmarks. (b) The high dynamic nature of egocentric vision introduces various difficulties in numerous visual tasks. (c) In a segment of egocentric video, the granularity of actions, objects, and environments varies significantly across different time scales. }\label{challenge}
\end{figure}

Moreover, we summarize the primary challenges in the egocentric vision domain based on the directions and content of recent related work in Fig. \ref{challenge}. Current works in egocentric vision predominantly leverage artificial intelligence-based approaches. Consequently, the scale and quality of datasets represent critical factors influencing the advancement of this field. At present, the largest image dataset in the exocentric perspective domain, WebLI-100B \cite{wang2025scaling}, introduced by Google’s DeepMind team, comprises 100 billion image-text pairs. In the video domain, the largest open-source text-video dataset, HD-VG-130M \cite{wang2023swap}, contains 130 million text-video pairs. By contrast, the largest dataset in the egocentric domain, Ego4D \cite{grauman_ego4d_2022}, includes only 3,670 hours of video, which is insufficient to support the development of various pre-training models and vision-language models for egocentric vision. Furthermore, in downstream tasks such as action understanding, pose estimation, and cross-perspective comprehension, the specialized datasets and evaluation benchmarks for egocentric vision suffer from limitations in both scale and quality. The absence of large-scale specialized datasets and high-quality evaluation benchmarks constitutes one of the most significant difficulties and challenges currently facing the egocentric vision domain.\circledstep[BurntOrange]{challenge:datasets}

In the previous content, we provide a detailed analysis of the characteristics of egocentric scenarios. Among these, the highly dynamic nature of egocentric vision, including rapid viewpoint changes \cite{yun_spherical_2024, tschernezki_epic_2024}, intense motion \cite{zhang_masked_2024}, and complex environmental interactions \cite{xu_weakly_2024}, represents the most significant distinction from conventional perspectives. This poses substantial challenges to egocentric understanding. Specifically, the frequent occurrence of occlusion in videos emerges as the most prominent issue \cite{zhao_fusing_2024, zhao_instance_2024, zhang_refa_2024, yang_egoposeformer_2024, wang_egocentric_2024}. For instance, when a user reaches to grasp an object, their hand may completely obscure the target, making it difficult for models to accurately recognize and interpret the scene’s content. This places stringent demands on the model’s capabilities for temporal understanding and memory retention.\circledstep[BurntOrange]{challenge:highly dynamic}

Additionally, in sect. \ref{Egocentric Task Decomposition}, we will further dissect the various components of egocentric vision. Overall, the semantic information in egocentric scenarios exhibits a multi-layered structure, with interactions occurring across these layers \cite{yuhanshen_learning_2024, wenqijia_audiovisual_2024, zhang_masked_2024, ivanrodin_action_2024}. Effectively representing these interactions and emphasizing the semantic content required for downstream tasks remains a significant difficulty in current research, mentioned in \cite{lai_eye_2022, ouyang_actionvos_2024, mur-labadia_affttention_2024}. In real-world egocentric scenarios, a notable characteristic is the extended temporal scale \cite{zhao_antgpt_2024, yuhanshen_progressaware_2024}. Wearable devices are typically in a default active state, continuously recording, while the content of interest often requires the model to autonomously identify it. This long temporal property further introduces a multi-granularity aspect to semantic information \cite{flaborea_prego_2024, yuhanshen_learning_2024, yang_egoposeformer_2024}: coarse-grained tasks demand that models extract abstract information from extended segments, whereas fine-grained tasks require models to decompose long sequences to provide more detailed content. However, most tasks necessitate the integration of information across different granularities, and this cross-granularity requirement likewise presents considerable challenges.\circledstep[BurntOrange]{challenge:multi-granularity}

\section{Egocentric Task Parsing}\label{Egocentric Task Decomposition}
\begin{figure}[ht]%
  \centering
  \includegraphics[width=0.9\textwidth]{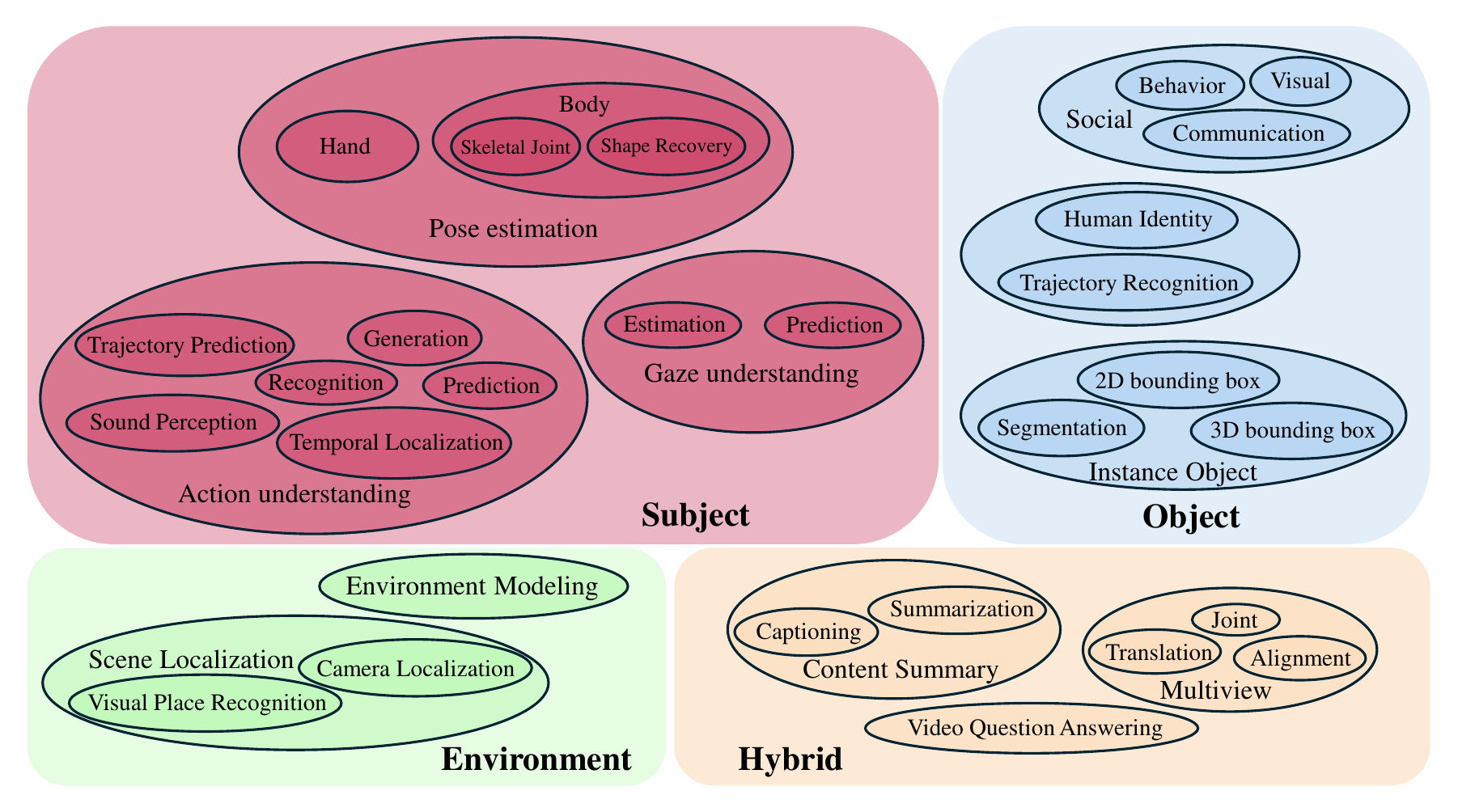}
  \caption{All egocentric tasks are categorized into four major groups, each containing multiple subtasks.}\label{tasks}
\end{figure}

In previous studies \cite{plizzari_outlook_2024,nunez-marcos_egocentric_2022}, egocentric vision tasks have typically been classified according to specific attributes such as objectives, methodologies, or technical frameworks. In contrast, our classification is grounded in an analysis of the fundamental elements that compose egocentric scenes. Specifically, we categorize tasks based on their primary focus on one of the three core scene elements, the subject, interacting objects, and the surrounding environment, or on their integrated understanding. This approach emphasizes what aspect of the egocentric scene each task aims to analyze or model. Accordingly, we divide egocentric vision tasks into four main categories (as illustrated in Fig. \ref{tasks}): \hyperref[Subject Understanding Related Tasks]{subject understanding}, \hyperref[Object Understanding Related Tasks]{object understanding}, \hyperref[Environment Understanding Related Tasks]{environment understanding}, and \hyperref[Hybrid Understanding Related Tasks]{hybrid understanding}. Each category corresponds to a distinct perspective on how information is extracted from egocentric visual data.

\textbf{Subject understanding tasks} primarily focus on analyzing the behavior, intentions, and actions of the subject with the scene in egocentric vision. Common subtasks include \hyperref[Gaze Understanding]{gaze understanding}, \hyperref[Pose Estimation]{pose estimation}, \hyperref[Action Understanding]{action recognition} each targeting different aspects of human activity and interaction. \textbf{Object understanding tasks} focus on identifying and analyzing external objects that the subject attends to within the scene, with subtasks such as \hyperref[Instance Object Recognition]{instance object recognition} and \hyperref[Social Perception]{social interaction modeling}, reflecting the dynamic relationship of objects. \textbf{Environmental understanding tasks} aim to model the structure and spatial layout of the scene to provide a more comprehensive contextual understanding. Subtasks include \hyperref[Scene Localization]{scene localization} and \hyperref[Environment Modeling]{environment modeling}, focusing on spatial awareness and navigation within egocentric environments. \textbf{Hybrid understanding tasks} integrate multi-dimensional information from the subject, objects, and environment to achieve a holistic understanding of complex scenes. Representative subtasks include \hyperref[Content Summary]{content summary}, \hyperref[Multiview Joint Understanding]{cross-view understanding} and \hyperref[Video Question Answering]{video question answering}, which typically require reasoning across multiple scene elements and modalities. The relationships and interdependencies between these subtasks are further discussed within the corresponding sections (sect. 4–7)

\section{Subject Understanding Tasks}\label{Subject Understanding Related Tasks}
Subject understanding tasks are divided based on the level and type of information extracted from the subject. These include gaze understanding (reflecting attention and intention), pose estimation (capturing physical state and interaction readiness), and action understanding (analyzing motion patterns across time and modalities). These subtasks are highly complementary: gaze supports intention inference, pose provides fine-grained cues for motion interpretation, and action understanding integrates these to analyze subject behavior holistically.

\subsection{Gaze Understanding}\label{Gaze Understanding}
Traditional gaze estimation methods explicitly infer users’ gaze directions using precise eye-tracking data, relying on 3D eye models, 2D regression, or appearance-based approaches, shown in \cite{cheng_appearancebased_2024}. These methods often achieve high accuracy but require dedicated hardware and complex pipelines.

\begin{figure}[ht]%
  \centering
  \includegraphics[width=0.5\textwidth]{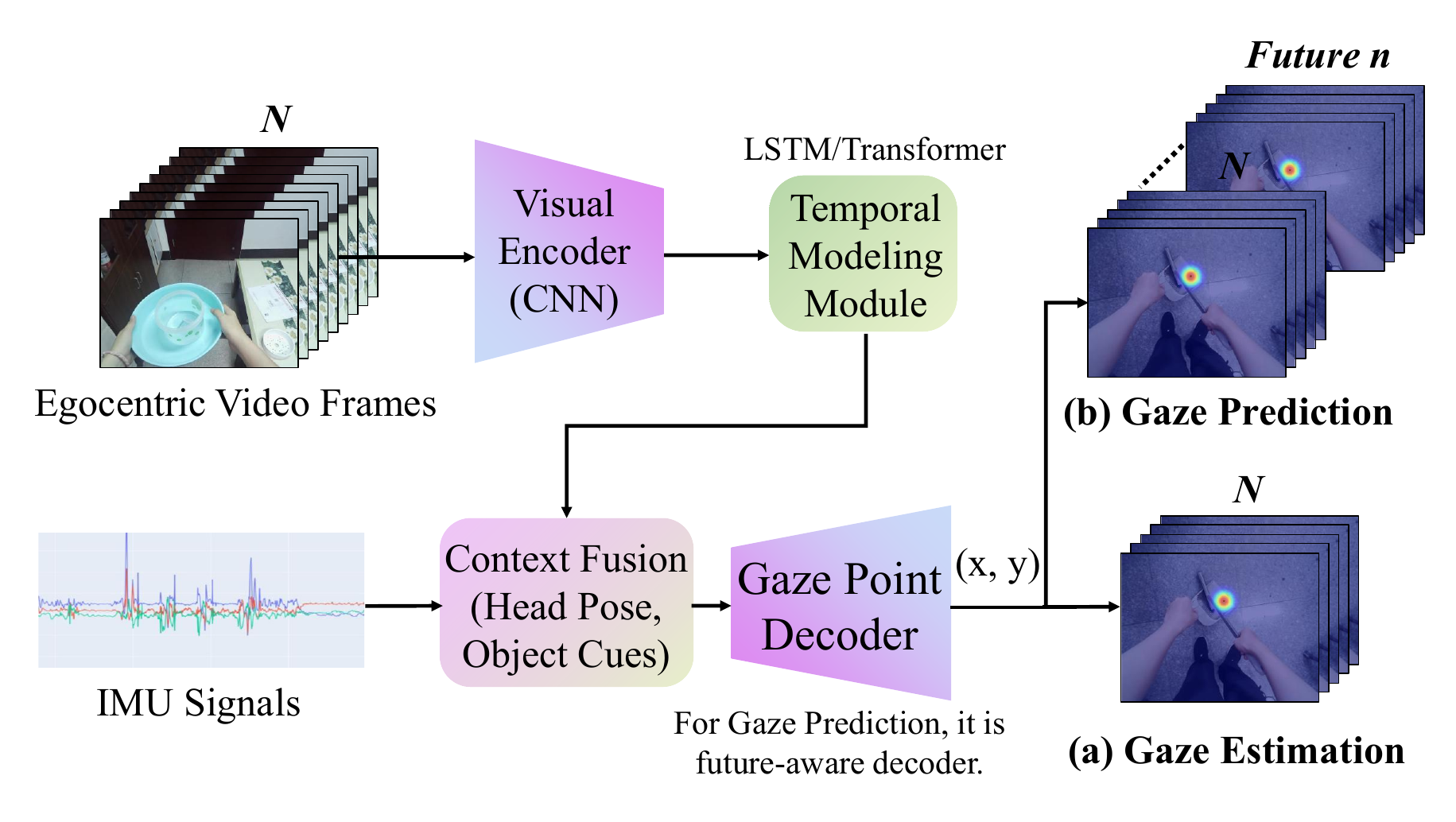}
  \caption{An overview of the gaze understanding pipeline in egocentric vision. The framework includes (a)gaze estimation and (b)prediction branches, leveraging visual features from egocentric frames and head poses information to infer the subject’s current gaze point and anticipate future gaze targets.}\label{Gaze}
\end{figure}

In contrast, this section focuses on implicit gaze understanding from egocentric videos, aiming to infer users’ attention via head motion, gaze shifts, and behavioral cues without direct eye movement data. Referring to works such as \cite{cazzato_when_2020, plizzari_outlook_2024, lai_listen_2024}, we explore two subtasks: \textit{gaze estimation} and \textit{gaze prediction}, illustrated in Fig. \ref{Gaze}. Common evaluation \textit{metrics} include Precision, Recall, and F1 Score. The performance of gaze estimation and prediction tasks are compared in Tab. \ref{tab:gaze_performance}.

\subsubsection{Gaze estimation}

Gaze estimation aims to infer the gaze point or generate attention heatmaps from egocentrical cues through deep learning or multimodal fusion to localize gaze regions.
\\ \textit{Seminal works:} Early representative methods focus on \textbf{contextual modeling}. Huang et al. \cite{huang_mutual_2020} proposed the Mutual Context Network (MCN), jointly modeling gaze and actions to exploit their mutual relevance. To incorporate multimodal signals, Thakur et al. \cite{thakur_predicting_2021} introduced a fusion network that combines egocentric video with inertial data (IMU) to enhance prediction robustness. 

Recent methods tend to adopt \textbf{transformer architectures} for global temporal reasoning. Lai et al. \cite{lai_eye_2022} proposed the Global-Local Correlation (GLC) module based on Transformer \cite{vaswani_attention_2017}, achieving a regression score of 60.8\% on EGTEA Gaze+ \cite{li_eye_2021}. Li et al. \cite{li_swingaze_2023} further extended this line by introducing SwinGaze, a Swin Transformer \cite{liu_swin_2021} based model that achieved 46.7\% accuracy and 37.8\% F1. On the large-scale EGO4D dataset \cite{grauman_ego4d_2022}, model of \cite{lai_eye_2022} also reported state-of-the-art performance.

\subsubsection{Gaze Prediction}
Gaze prediction aims to forecast users’ future gaze locations by analyzing current visual cues and implicit behavioral signals from egocentric videos. Unlike gaze estimation, this task requires reasoning over temporal dynamics, such as motion trajectories, head orientation, and intention patterns, to anticipate where attention will shift next.

Recent methods primarily fall into two directions: those (like \cite{yun_spherical_2024}) focusing on \textbf{compensating self-motion} and those (like \cite{lai_listen_2024}) \textbf{leveraging multimodal temporal fusion}. Yun et al. \cite{yun_spherical_2024} introduced the MuST framework, which transforms audiovisual signals into a head-pose-aligned space to reduce egocentric motion interference. Lai et al. \cite{lai_listen_2024} proposed a contrastive spatial-temporal separable (CSTS) fusion method that integrates both visual and audio streams, demonstrating its effectiveness across datasets such as Ego4D \cite{grauman_ego4d_2022} and AEA \cite{lv_aria_2024}. This line of work highlights the importance of temporal modeling and multimodal fusion for robust gaze anticipation in dynamic, first-person scenarios.

\begin{table}[ht]
  \footnotesize
  \centering
  \caption{Performance comparison of representative gaze understanding methods.}
  \label{tab:gaze_performance}
  \begin{tabular}{lcc}
    \toprule
    \textbf{Methods} & \multicolumn{2}{c}{\textbf{Datasets} F1($\uparrow$)}\\
    \Xhline{0.4pt}
    \addlinespace[3pt]
    \textit{Gaze Estimation} & \textbf{EGTEA Gaze+~\cite{li_eye_2021}} & \textbf{Ego4D~\cite{grauman_ego4d_2022}} \\
    GLC(2022)~\cite{lai_eye_2022} & 44.8\%  & \textbf{43.1\%} \\
    SwinGaze(2023)~\cite{li_swingaze_2023} & \textbf{46.7\%}  & -- \\
    \Xhline{0.4pt}
    \addlinespace[3pt]
    \textit{Gaze Prediction} & \textbf{AEA~\cite{lv_aria_2024}} & \textbf{Ego4D~\cite{grauman_ego4d_2022}} \\
    MuST(2024)~\cite{yun_spherical_2024} & 8.78 MAE($\downarrow$)\textsuperscript{†} & -- \\
    CSTS(2024)~\cite{lai_listen_2024} & \textbf{39.7\%} \textsuperscript{‡} & \textbf{59.9\%} \textsuperscript{‡} \\
    \bottomrule
  \end{tabular}
  \vspace{1mm}
  \begin{flushright}
  \textsuperscript{†} Mean Angular Error (MAE) with anticipation time 300ms. \\
  \textsuperscript{‡} F1 score with anticipation time 2 seconds.
  \end{flushright}
\end{table}

\subsection{Pose Estimation}\label{Pose Estimation}

Perceiving one's own position and posture is essential for interaction and motion in humans and animals. Egocentric pose estimation aims to generate accurate 2D or 3D representations of the body using multimodal inputs (e.g., RGB, depth, or IMU data), enabling deeper understanding of self-motion and supporting downstream tasks such as gesture recognition, AR/VR tracking, rehabilitation, and hand-object interaction.

\begin{figure}[ht]%
  \centering
  \includegraphics[width=0.5\textwidth]{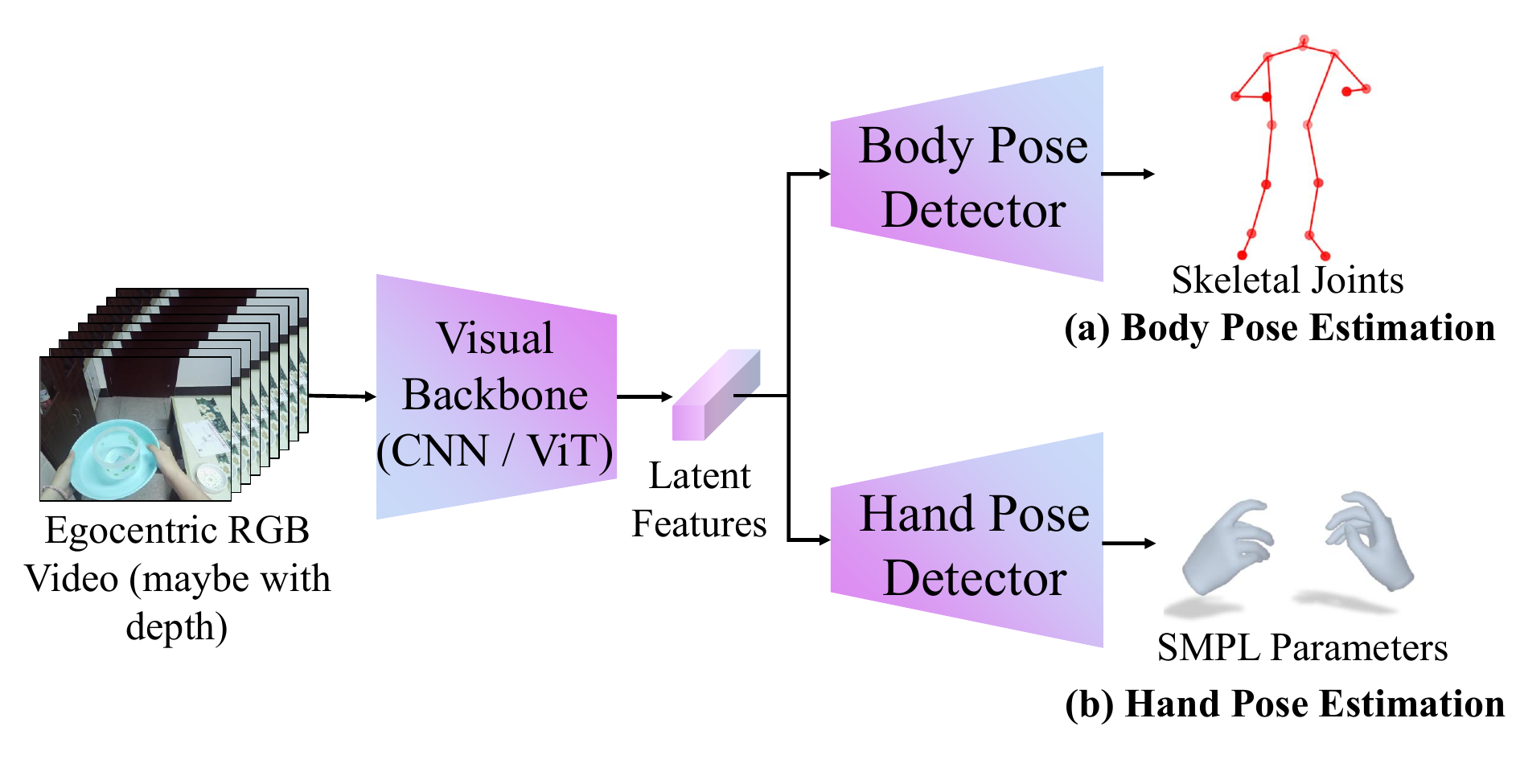}
  \caption{Illustration of a representative architecture for egocentric pose estimation. Visual features are extracted from first-person frames and used to locate (a)body or (b)hand regions.}\label{Pose}
\end{figure}

While extensive progress has been made in exocentric pose estimation~\cite{sarafianos_3d_2016, bengamra_review_2021, zheng_deep_2023}, the rise of wearable devices has spurred interest in egocentric settings~\cite{bandini_analysis_2023, azam_survey_2024, plizzari_outlook_2024, fan_benchmarks_2024}. Existing works mainly fall into two categories: full-body pose estimation and fine-grained hand pose estimation, as illustrated in Fig.~\ref{Pose}. Common \textit{metrics} for pose estimation tasks include MPJPE, PA-MPJPE, MPJRE and MPJVE, illustrated in \cite{azam_survey_2024}. Performance of under representative methods shown in Tab. \ref{tab:pose_comparison}.

\subsubsection{Body Pose Estimation}

In Human Pose Estimation (HPE), existing methods are typically categorized by output type: skeletal joint estimation and body mesh recovery. \textbf{Skeletal joint estimation} focuses on locating keypoints in 3D space to construct an abstract human skeleton, where each joint is represented by spatial coordinates.

\textit{Seminal works:} Early egocentric pose estimation methods often leveraged \textbf{specialized camera configurations} to mimic real-world usage. For instance, Tome et al.~\cite{tome_xregopose_2019} and Xu et al.~\cite{xu_mo2cap2_2019} adopted head-mounted fisheye cameras for mobile full-body pose capture, while Hori et al.~\cite{hori_silhouettebased_2022} explored wrist-mounted 360° panoramic cameras for single-camera 3D estimation.

To address \textbf{self-occlusion}, several works propose \textbf{indirect or uncertainty-aware strategies}. Ng et al.~\cite{ng_you2me_2020} inferred the subject’s pose via the observed pose of an interacting partner. Wang et al.~\cite{wang_estimating_2021} introduced an uncertainty-aware reprojection energy term, and Zhao et al.~\cite{zhao_egoglass_2021} leveraged partial body cues. Stereo-based and binocular approaches were explored in~\cite{akada_unrealego_2022, kang_ego3dpose_2023} using triangulation or multi-view fusion.

Other works utilize \textbf{scene context} to guide pose prediction and \textbf{enhance temporal reasoning} to improve robustness. Wang et al.~\cite{wang_sceneaware_2023} projected egocentric depth features into voxel space for scene-aware estimation. Kang et al.~\cite{kang_attentionpropagation_2024} and Akada et al.~\cite{akada_3d_2024} leveraged self-attention and structure-from-motion-based scene reconstruction to enhance temporal modeling. Zhao et al.~\cite{zhao_egobody3m_2024} fused latent multi-view features from VR settings, while Millerdurai et al.~\cite{millerdurai_eventego3d_2024} achieved high-speed (140Hz) real-time estimation using event-based cameras.

Recent advances adopt \textbf{transformer architectures and diffusion priors} for high-precision estimation. EgoPoseFormer~\cite{yang_egoposeformer_2024} follows a two-stage pipeline, achieving 33.4mm MPJPE on UnrealEgo~\cite{akada_unrealego_2022}. Wang et al.~\cite{wang_egocentric_2024} addressed fisheye distortion via a ViT-based feature extractor (FisheyeViT) and used a diffusion model to refine motion priors, reaching 57.59mm MPJPE on SceneEgo~\cite{wang_sceneaware_2023}.

\textbf{Body shape recovery} focuses on reconstructing full 3D human meshes—including joint locations and body surface—based on parameterized models such as SMPL~\cite{loper_smpl_2015} and SMPL-X~\cite{pavlakos_expressive_2019}. 

\textit{Seminal works:} \textbf{Sensor-fusion-based methods} aim to improve accuracy using multimodal input. Guzov et al.~\cite{guzov_human_2021} combined egocentric video and IMU data for smooth motion reconstruction. Jiang et al.~\cite{jiang_avatarposer_2022} and Zhang et al.~\cite{zhang_refa_2024} extended this idea to include motion signals from mixed reality headset with shandheld controllers or infrared cameras in head-mounted displays.

\textbf{Generative modeling approaches} tackle the challenge of sparse input by learning priors over body motion. Dittadi et al.~\cite{dittadi_fullbody_2021} used a VAE to recover pose from limited joint signals, while Aliakbarian et al.~\cite{aliakbarian_flag_2022} proposed a flow-based model to capture conditional pose distributions. Du et al.~\cite{du_avatars_2023} introduced a diffusion-based MLP architecture to estimate lower-body motion using only upper-body tracking.
\textbf{Personalized and long-time reconstruction methods} emphasize temporal consistency and user-specific adaptation. Jiang et al.~\cite{jiang_egoposer_2024} proposed EgoPoser, which decouples pose and global position, and integrates a SlowFast temporal module to support long-sequence inference. It also estimates individual body shape for accurate avatar anchoring, achieving SOTA performance on the HPS dataset~\cite{guzov_human_2021}.

\subsubsection{Hand Pose Estimation}

Egocentric hand pose estimation aims to regress 3D hand keypoints from first-person inputs such as RGB images, videos, depth maps, or 3D meshes. As hands are the primary medium for physical interaction, accurately modeling their poses is crucial for understanding human activities. From everyday gestures like picking up a phone to complex tasks like assembling components or operating appliances, hand-object interactions underpin most goal-directed behaviors.

\textit{Seminal works:} Existing egocentric hand pose estimation methods vary in model architecture, camera setup, and task formulation. For \textbf{direct hand-object modeling}, Kwon et al.~\cite{kwon_h2o_2021} introduced a fully convolutional network that predicts the 3D meshes of both hands and manipulated objects. Fan et al.~\cite{fan_arctic_2023} expanded the problem scope by proposing new tasks—consistent motion reconstruction and interaction field estimation, to better capture the physical dynamics during hand-object interaction. 

In \textbf{monocular scenarios}, Pavlakos et al.~\cite{pavlakos_reconstructing_2024} developed HaMeR, a Transformer-based hand mesh recovery model robust across benchmarks and in-the-wild data. For \textbf{multi-view settings}, Zhou et al.~\cite{zhou_1st_2023} adopted a ViT-based backbone to build a strong baseline for egocentric 3D keypoint prediction. To bridge the gap between single-view training and multi-camera VR setups, Liu et al.~\cite{liu_singletodualview_2024} proposed an unsupervised single-to-dual view adaptation framework (S2DHand). Wang et al.~\cite{wang_egocentric_2024} further integrated hand and body pose estimation, using a diffusion model to refine full-body predictions.

To improve \textbf{generalization to in-the-wild scenarios}, Prakash et al.~\cite{prakash_3d_2024} proposed WildHands, combining 3D supervision from lab-based datasets (e.g., ARCTIC~\cite{fan_arctic_2023}) with 2D auxiliary signals from large-scale egocentric video datasets (e.g., Ego4D~\cite{grauman_ego4d_2022}), achieving state-of-the-art accuracy on ARCTIC.

\begin{table}[]
  \centering
  \footnotesize
  \caption{Performance comparison of representative methods for egocentric body and hand pose estimation.}
  \label{tab:pose_comparison}
  \begin{tabular}{lcc}
  \toprule
  \textbf{Methods} & \multicolumn{2}{c}{\textbf{Datasets} MPJPE(mm $\downarrow$)} \\
  \Xhline{0.4pt}
  \addlinespace[3pt]
  \textit{Skeletal joint estimation} & \textbf{UnrealEgo~\cite{akada_unrealego_2022}} & \textbf{SceneEgo~\cite{wang_sceneaware_2023}} \\
  Ego3DPose(2023)   \cite{kang_ego3dpose_2023} & 60.82  & — \\
  EgoTAP(2024)   \cite{kang_attentionpropagation_2024} & 41.06 & — \\
  Akada et   al.(2024)~\cite{akada_3d_2024} & 46.20 & — \\
  EgoPoseFormer(2024)~\cite{yang_egoposeformer_2024} & \textbf{33.40} & 93.00 \\
  EgoWholeBody(2024)~\cite{wang_egocentric_2024} & — & \textbf{57.59} \\
  \Xhline{0.4pt}
  \addlinespace[3pt]
  \textit{Body shape recovery} & \textbf{AMASS~\cite{mahmood_amass_2019}} & \textbf{HPS~\cite{guzov_human_2021}} \\
  AvatarPoser(2022)~\cite{jiang_avatarposer_2022} & \textbf{26.10} & 201.78\textsuperscript{†} \\
  FLAG(2022)~\cite{aliakbarian_flag_2022} & 49.60 & — \\
  AGRoL(2023)~\cite{du_avatars_2023} & 26.60 & 202.18* \\
  EgoPoser(2024)~\cite{jiang_egoposer_2024} & 41.40 & \textbf{92.9} \\
  \Xhline{0.4pt} 
  \addlinespace[1pt]
  \Xhline{0.4pt}
  \addlinespace[3pt]
  \textit{Hand Pose Estimation} & \textbf{AssemblyHands~\cite{ohkawa_assemblyhands_2023}} & \textbf{ARCTIC~\cite{fan_arctic_2023}} \\
  ArcticNet-SF(2023)~\cite{fan_arctic_2023} & 110.76 \textsuperscript{‡} & 19.20 \\
  Zhou et al.(2023)~\cite{zhou_1st_2023} & \textbf{12.21} & — \\
  S2DHand(2024)~\cite{liu_singletodualview_2024} & 20.16 & — \\
  WildHands(2024)~\cite{prakash_3d_2024} & 80.40 & \textbf{15.72} \\
  \bottomrule
  \end{tabular}
  \vspace{1mm}
  \begin{flushright}
  \textsuperscript{†} Reported in \cite{jiang_egoposer_2024}
  \textsuperscript{‡} Reported in \cite{prakash_3d_2024}, zero shot config.
  \end{flushright}
\end{table}

\subsection{Action Understanding}\label{Action Understanding}
\begin{figure}[ht]%
  \centering
  \includegraphics[width=0.5\textwidth]{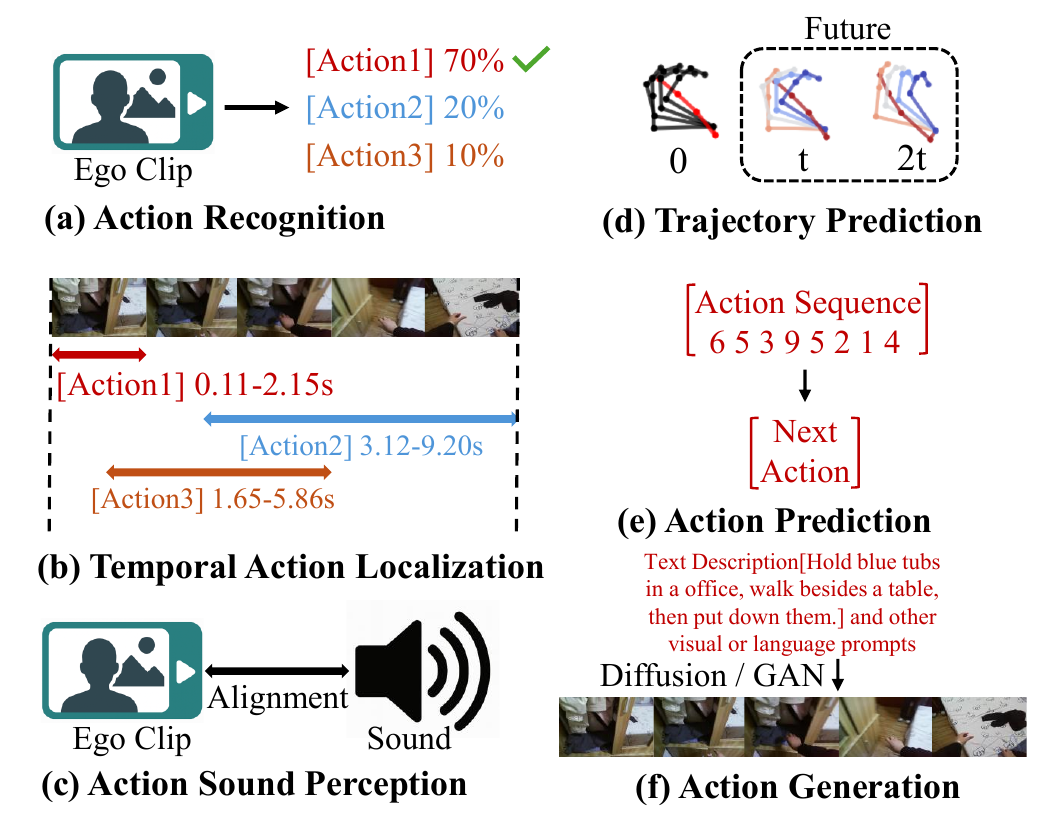}
  \caption{Representative forms of egocentric action understanding tasks: (a) action recognition, (b) temporal action detection, (c) action anticipation, (d) trajectory prediction, (e) action-sound association, and (f) action generation.}\label{Action}
\end{figure}

Egocentric action understanding aims to semantically abstract continuous body motions by modeling multimodal signals (e.g., vision, audio, gaze) and their contextual relationships with interactive objects and environments across time. This task is critical for real-time decision support and personalization in AR/VR, human-computer interaction, robotics, and intelligent surveillance. For instance, accurate action recognition facilitates imitation learning in robotics and enables intuitive interfaces in augmented reality. Building on existing surveys~\cite{nunez-marcos_egocentric_2022,kong_human_2022,hu_online_2022,lai_human_2024,plizzari_outlook_2024,ding_temporal_2024} and recent advances, we categorize egocentric action understanding tasks based on input-output forms, as illustrated in Fig.~\ref{Action}. Performance of representative methods for action
understanding is shown in Tab. \ref{tab:action_understanding}

\subsubsection{Action Recognition}
Action recognition aims to classify the specific actions performed by the subject in videos. In the egocentric setting, this involves analyzing visual data from the wearer's perspective to recognize ongoing actions, which may also be simplified into binary categories (e.g., correct vs.\ incorrect). This task is typically formulated as a classification problem, and evaluated using standard \textit{metrics} such as Top-k Accuracy, Precision, and Recall.

\begin{table}[ht]
  \centering
  \footnotesize
  \caption{Performance comparison of representative methods for egocentric action understanding.}
  \label{tab:action_understanding}
  \begin{tabular}{lcc}
  \toprule
  \textbf{Methods} & \multicolumn{2}{c}{\textbf{Datasets}} \\
  \Xhline{0.4pt}
  \addlinespace[3pt]
  \textit{Recognition} &   \textbf{EK-100~\cite{damen_rescaling_2022}} V\textsuperscript{*} Acc. (\% $\uparrow$)  &  N\textsuperscript{*} Acc. (\% $\uparrow$)\\
  ORViT(2022)~\cite{herzig_objectregion_2022} & 68.40  & 58.70 \\
  SUM-L(2023)\cite{wang_learning_2023} & 67.00  & 53.40 \\
  Shiota et al.(2024)~\cite{shiota_egocentric_2024} & 53.30  & 52.10 \\
  LVMAE(2024)~\cite{gundavarapu_extending_2024} & \textbf{75.00}  &  \textbf{61.80}\\
  X-MIC(2024)~\cite{annakukleva_xmic_2024} & 29.49\textsuperscript{†}  & 18.96\textsuperscript{†} \\
  
  \Xhline{0.4pt} 
  \addlinespace[1pt]
  \Xhline{0.4pt}
  \addlinespace[3pt]

  \textit{Temporal Localization} & \textbf{EK-100~\cite{damen_rescaling_2022}} V mAP (\% $\uparrow$) &  N mAP (\% $\uparrow$) \\
  Actionformer(2022)~\cite{zhang_actionformer_2022} &23.50  & 21.90 \\
  TriDet(2023)~\cite{shi_tridet_2023} & 25.40  & 23.80 \\
  Ego-Only(2023)~\cite{wang_egoonly_2023} & 29.00  & 28.10 \\
  AdaTAD(2024)~\cite{liu_endtoend_2024} & \textbf{29.30}  & \textbf{29.30} \\

  \Xhline{0.4pt} 
  \addlinespace[1pt]
  \Xhline{0.4pt}
  \addlinespace[3pt]

  \textit{Prediction} & \textbf{EK-100~\cite{damen_rescaling_2022}} V Recall@5 (\% $\uparrow$) &  N Recall@5 (\% $\uparrow$) \\
  UADT(2024) \cite{guo_uncertaintyaware_2024} & 43.50  & 46.60 \\
  InAViT(2024) \cite{roy_interaction_2024} & \textbf{52.54}  & \textbf{51.93} \\
  S-GEAR(2024) \cite{diko_semantically_2024} & 30.20  & 37.00 \\

  \Xhline{0.4pt} 
  \addlinespace[1pt]
  & \textbf{E4D-STA~\cite{grauman_ego4d_2022}} N mAP@5 (\% $\uparrow$) &  N+V mAP@5 (\% $\uparrow$) \\
  NAOGAT(2024) \cite{thakur_leveraging_2024} & 27.00  & 6.54 \\
  STAformer(2024) \cite{mur-labadia_affttention_2024} & \textbf{33.50}  & \textbf{17.25} \\

  \Xhline{0.4pt} 
  \addlinespace[1pt]
  & \textbf{E4D-LTA~\cite{grauman_ego4d_2022}} V ED@20 ($\downarrow$) &  N ED@20 ($\downarrow$) \\
  AntGPT(2024) \cite{zhao_antgpt_2024} & 0.724  & 0.744 \\
  PALM(2024) \cite{kim_palm_2024} & \textbf{0.656}  & \textbf{0.640} \\
  PlausiVL(2024) \cite{mittal_cant_2024} & 0.679  & 0.681\\

  \bottomrule
  \end{tabular}
  \vspace{1mm}
  \begin{flushright}
  \textsuperscript{*} N: Noun, V: Verb
  \textsuperscript{†} Cross-domain from Ego4D \cite{grauman_ego4d_2022}
  \end{flushright}
\end{table}

\textit{Seminal works:} Egocentric action recognition has explored \textbf{diverse multimodal cues and modeling strategies}. Early methods such as EPIC-Fusion~\cite{kazakos_epicfusion_2019} fused RGB, optical flow, and audio modalities, framing classification as separate verb-noun predictions. Gaze~\cite{min_integrating_2021}, gait~\cite{thapar_sharing_2020}, IMU~\cite{zhang_masked_2024}, and hand cues~\cite{shiota_egocentric_2024} were later incorporated to enhance action semantics. Modality-specific adaptations included event-camera usage~\cite{plizzari_e2_2022} and exocentric pretraining for cross-domain scale transfer~\cite{li_egoexo_2021}. Fusion-based methods such as EgoDistill~\cite{tan_egodistill_2023} and MMGEgo4D~\cite{gong_mmgego4d_2023} further improved cross-modal generalization through knowledge distillation and Transformer-based architectures.

To effectively model \textbf{complex temporal structures}, recent methods leveraged Vision Transformers and attention mechanisms for fine-grained temporal modeling. MeMViT~\cite{wu_memvit_2022} and MTV~\cite{yan_multiview_2022} explored multi-scale temporal aggregation, while Lu et al.~\cite{lu_mixed_2025} introduced a Mixed Attention Channel Shift Transformer for capturing both local and global temporal dependencies efficiently. Gundavarapu et al.~\cite{gundavarapu_extending_2024} proposed adaptive token reconstruction in masked autoencoders, achieving state-of-the-art performance on benchmarks like EPIC-Kitchens-100~\cite{damen_rescaling_2022}.

Recent advancements in \textbf{skeleton-based action recognition} have focused on modeling spatial-temporal dependencies. Chen et al.~\cite{chen_skeletonbased_2025} utilized the Hilbert-Schmidt Independence Criterion to effectively capture non-linear dependencies among skeletal joints. Complementarily, Geng et al.~\cite{geng_hierarchical_2024} presented hierarchical aggregated graph neural networks to enhance fine-grained action recognition via multi-level skeletal graph features.

Another promising direction employs \textbf{causal inference and knowledge-based reasoning} to improve robustness. Liu et al.~\cite{liu_knowledgebased_2024} proposed a hierarchical causal inference network, explicitly integrating domain knowledge to address biases and confounding effects. Further building on knowledge-driven approaches, compositional reasoning frameworks have been explored by Liu et al.~\cite{liu_knowledgedriven_2025}, decomposing complex actions into primitive components to facilitate interpretability and generalization. Jiao et al.~\cite{jiao_braininspired_2025} reviewed brain-inspired learning frameworks, proposing cognitive-inspired models as robust solutions for enhanced action perception and recognition.

To boost \textbf{cross-domain generalization}, recent efforts like CIR~\cite{plizzari_what_2023}, SUM-L~\cite{wang_learning_2023}, and ALGO~\cite{kundu_discovering_2024} have introduced domain-invariant representations and symbolic reasoning. Few-shot and cross-modal adaptation strategies such as MM-CDFSL~\cite{hatano_multimodal_2024} and X-MIC~\cite{annakukleva_xmic_2024} demonstrated effective knowledge transfer in limited-supervision scenarios.

Finally, specific \textbf{task-oriented applications} are gaining prominence. PREGO~\cite{flaborea_prego_2024}, for example, focuses on procedural error detection in real-time industrial environments, highlighting the practical relevance and deployment potential of egocentric action understanding methods.

\textit{Seminal works:} Egocentric action recognition has explored \textbf{diverse multimodal cues and modeling strategies}. Early methods such as EPIC-Fusion~\cite{kazakos_epicfusion_2019} fused RGB, optical flow, and audio modalities, and framed classification as separate verb-noun prediction. Gaze~\cite{min_integrating_2021}, gait~\cite{thapar_sharing_2020}, IMU~\cite{zhang_masked_2024}, and hand cues~\cite{shiota_egocentric_2024} were later incorporated to enhance action semantics. Several works focused on modality-specific adaptations, such as event cameras for low-power processing~\cite{plizzari_e2_2022} and exocentric pretraining for scale transfer~\cite{li_egoexo_2021}. More recently, fusion-based designs such as EgoDistill~\cite{tan_egodistill_2023} and MMGEgo4D~\cite{gong_mmgego4d_2023} improved cross-modal generalization through distillation and Transformer-based fusion. 

To model \textbf{complex temporal structures}, MeMViT~\cite{wu_memvit_2022} and MTV~\cite{yan_multiview_2022} leveraged Vision Transformers to represent long and multi-granularity temporal sequences. Object-centric designs like ORViT~\cite{herzig_objectregion_2022} embedded object features directly into video transformer pipelines. To capture longer temporal dependencies efficiently, Gundavarapu et al.~\cite{gundavarapu_extending_2024} proposed a masked autoencoder with adaptive token reconstruction. Their model achieved state-of-the-art performance on EPIC-Kitchens-100~\cite{damen_rescaling_2022}, reaching 75\% verb accuracy and 61.8\% noun accuracy. To boost \textbf{generalization across scenes and domains}, CIR~\cite{plizzari_what_2023}, SUM-L~\cite{wang_learning_2023}, and ALGO~\cite{kundu_discovering_2024} introduced domain-invariant representation learning and symbolic reasoning techniques. 

\textbf{Cross-domain adaptation} under limited supervision has also drawn attention. MM-CDFSL~\cite{hatano_multimodal_2024} addressed few-shot multimodal transfer via unlabeled target domain distillation. X-MIC~\cite{annakukleva_xmic_2024} injected egocentric knowledge into CLIP-style visual-language embeddings using a video adapter. Beyond general recognition, \textbf{task-specific applications} have emerged. PREGO~\cite{flaborea_prego_2024} introduced a one-class online detection model to identify procedural errors in real-time industrial settings.

\subsubsection{Temporal Action Localization}

Temporal Action Localization (also known as action detection or segmentation) aims to identify the start and end frames of actions in untrimmed videos and assign predefined labels to each segment. In embodied intelligence, it supports evaluating and optimizing robotic executions. Common evaluation \textit{metrics} include frame-level and segment-level measures such as Mean over Frames (MoF), Edit Score, F1@$\tau$, and mAP@$\tau$~\cite{ding_temporal_2024}.

\textit{Seminal works:} Unlike most temporal action localization models originally developed for exocentric data~\cite{ding_temporal_2024}, egocentric-specific research has focused on representing continuous actions and adapting to the unique challenges of first-person videos. To model \textbf{temporal dependencies and inter-segment relations}, GTRM~\cite{huang_improving_2020} introduces a graph-based reasoning module that can be attached to existing segmentation models. Lin et al.~\cite{lin_egocentric_2022} proposed EgoNCE for video-text contrastive pretraining. Then Zhang et al.~\cite{zhang_actionformer_2022} introduced ActionFormer, which integrates multi-scale features and self-attention to capture long-term context. TriDet~\cite{shi_tridet_2023} enhances boundary modeling by estimating relative probability distributions rather than direct offset regression. Liu et al.~\cite{liu_endtoend_2024} scaled both model size (1B parameters) and temporal input (1536 frames), and proposed TIA (Temporal-Informative Adapter) to reduce memory while preserving performance, achieving SOTA results with 29.3\% average mAP for verb and noun detection on EPIC-KITCHENS-100~\cite{damen_rescaling_2022} at IoU thresholds $\{0.1, 0.2, 0.3, 0.4, 0.5\}$.

To mitigate \textbf{domain gaps} between exocentric pretraining and egocentric testing, Wang et al.~\cite{wang_egoonly_2023} showed that such pretraining may degrade performance, while Quattrocchi et al.~\cite{quattrocchi_synchronization_2024} proposed an unsupervised distillation-based transfer method that significantly improved generalization to egocentric data.

For \textbf{real-time and procedural applications}, Shih-PoLee et al.~\cite{shih-polee_error_2024} detected process errors using graph-based prototypes trained on error-free videos. Shen et al.~\cite{yuhanshen_progressaware_2024} designed an online segmentation model with an Action Progress Prediction (APP) module, while Reza et al.~\cite{reza_hat_2024} introduced HAT, a memory-efficient transformer framework enabling long-term historical reasoning.

\subsubsection{Action Sound Perception}
We introduce the task of \textbf{action sound perception}, which differs from prior work that uses audio merely as a supplementary cue. Here, audio is treated as a primary modality—either for retrieving relevant sound clips or generating action-consistent pseudo-audio directly from video. This task is challenging due to the diversity and noise in egocentric audio signals. Action-related sounds are often short, subtle, and easily masked by ambient noise. Furthermore, many sound sources are off-screen, requiring models to reason beyond visible content and align audio with visual actions. Common \textit{metrics} include Top-1 and Top-5 Recall for retrieval, and Fréchet Audio Distance (FAD)~\cite{kilgour_frechet_2018} or learned similarity scores~\cite{wu_largescale_2023, luo_difffoley_2023} for generation.

\textit{Seminal works:} Research on egocentric action sound perception is still emerging. Huh et al.~\cite{huh_epicsounds_2023} introduced the EPIC-SOUNDS dataset with annotated action-related sound events and evaluated audio recognition baselines, revealing performance gaps in sound-based action recognition. To improve text-audio alignment, Oncescu et al.~\cite{oncescu_sound_2024} proposed generating audio-centric captions using large language models (LLMs) for better audio retrieval. Chen et al.~\cite{chen_action2sound_2024} addressed the challenge of disentangling foreground action sounds from background noise via audio conditioning, and released the Ego4D-Sounds dataset. Further, Chen et al.~\cite{chen_soundingactions_2024} proposed a self-supervised method to embed video, audio, and text into a shared representation space for multimodal alignment. 

\subsubsection{Trajectory Prediction}
Trajectory prediction aims to forecast the future movement of the body or hands based on historical and current multimodal egocentric data. Unlike other tasks, it requires both accurate perception of the present state and robust inference of future behaviors in dynamic environments. As a causal forecasting problem, it is inherently uncertain—small perturbations can lead to large deviations—demanding models to learn meaningful behavioral patterns from historical cues. A common \textit{metric} for evaluation is the Final Displacement Error (FDE), which measures the L2 distance between predicted and ground-truth positions at a future timestamp.

\textit{Seminal works:} Early methods focused on \textbf{2D hand movement forecasting} from video. Liu et al. \cite{liu_forecasting_2020} proposed a Motor Attention model to estimate future hand positions using probabilistic reasoning. Jia et al. \cite{jia_generative_2022} addressed future hand mask prediction with EgoGAN, which combines a 3D FCN and GAN for realistic sample generation. For \textbf{3D trajectory prediction}, Bao et al. \cite{bao_uncertaintyaware_2023} introduced USST, a state-space model that incorporates attention and uncertainty estimation to handle motion ambiguity. Recent work has leveraged \textbf{multitask learning and diffusion modeling} to improve prediction accuracy. Abilkassov et al. \cite{abilkassov_augmenting_2024} adopted a multitask approach using UniFormer \cite{li_uniformer_2023} to predict both hand position and hand-object interaction timing, achieving SOTA results on the Ego4D benchmark \cite{grauman_ego4d_2022}. To explore 3D skeleton-based hand motion forecasting, Tang et al. \cite{tang_prompting_2024} proposed PromptFDDM, a diffusion model guided by motion prompts, and achieved top performance on the FPHA dataset \cite{garcia-hernando_firstperson_2018}.

\subsubsection{Action Prediction}

The task of action prediction aims to infer the subject's future actions from historical multimodal egocentric video. It can be seen as a causal extension of action recognition or detection, sharing similarities with trajectory prediction but emphasizing higher-level semantic reasoning. The output is often expressed as action labels or natural language, aligning more closely with human cognition. Common evaluation \textit{metrics} include Top-k Accuracy, Precision, Recall, Mean over Frames (MoF), Edit Score, F1@$\tau$, and mAP@$\tau$, similar to those used in recognition and detection tasks.

\textit{Seminal works:} 
Recent works on egocentric action prediction can be broadly categorized along two key dimensions: \textbf{(1) Modeling Paradigm}, i.e., how future actions are inferred; and \textbf{(2) Prediction Granularity}, i.e., whether the goal is to predict action labels, durations, or structured semantics.

\textbf{(1) Modeling Paradigm.} \textit{Feature extrapolation models} forecast future representations directly. For instance, Wu et al. \cite{wu_learning_2021} decompose long-term prediction into short-term feature forecasts. \textit{Goal-conditioned models} use latent or symbolic goals for guidance, such as Roy et al. \cite{roy_action_2022}, who proposed latent goal-based RNNs. \textit{Transformer-based models} leverage temporal attention, e.g., Nawhal et al. \cite{nawhal_rethinking_2022} introduced a two-stage segment-based transformer; Girdhar et al. \cite{girdhar_anticipative_2021} presented an end-to-end video transformer. \textit{Multimodal \& structure-aware models} integrate object interactions and language cues, Ashutosh et al. \cite{ashutosh_hiervl_2023} used video-language embedding; Ivanrodin et al. \cite{ivanrodin_action_2024} modeled action-object relations via egocentric action scene graphs (EASGs); and NAOGAT \cite{thakur_leveraging_2024} introduced interaction-guided transformers.

\textbf{(2) Prediction Granularity.} Some methods only predict future \textit{action labels} (e.g., \cite{wu_learning_2021, girdhar_anticipative_2021}), while others also predict \textit{durations} (\cite{zhao_diverse_2020}) or full \textit{semantic descriptions} (\cite{abdelslam_gepsan_2023}). To capture higher-level semantics, several works utilize \textit{Language Models (LLMs)} for reasoning: Diko et al. \cite{diko_semantically_2024} and Mittal et al. \cite{mittal_cant_2024} generate plausible future sequences; Zhao et al. \cite{zhao_antgpt_2024} proposed AntGPT, incorporating procedural knowledge; and Kim et al. \cite{kim_palm_2024} designed PALM, leveraging LLMs with prompt selection for long-term forecasting.

Representative SOTA models fall into the above categories. InAViT \cite{roy_interaction_2024}, an interaction-region-aware visual transformer, achieved 25.89\% Top-5 recall on EPIC-Kitchens-100. STAformer \cite{mur-labadia_affttention_2024} excelled at short-term object interaction anticipation on Ego4D. UADT \cite{guo_uncertaintyaware_2024} improved generalization by decoupling verb and noun prediction, achieving 68.4\% Top-5 Recall on EGTEA Gaze+. PALM \cite{kim_palm_2024} advanced long-term action prediction using LLMs and example prompting, surpassing prior benchmarks on Ego4D-LTA.

\subsubsection{Action Generation}

With the rapid advancement of diffusion models \cite{ho_denoising_2020} in high-quality and diverse image generation, and inspired by the novel task definition in \cite{lai_lego_2024}, a promising direction emerges: \textbf{egocentric action generation}. This task aims to generate images or videos based on natural language inputs, possibly enriched with multimodal egocentric cues.

However, this task faces key challenges. First, ensuring temporal coherence, action accuracy, and controllability in generation remains difficult. Second, most generative models are trained on exocentric data with static scenes, which hinders adaptation to egocentric settings, leading to domain transfer issues. Common evaluation \textit{metrics} include PSNR, EgoVLP and EgoVLP+ scores \cite{lin_egocentric_2022}, and CLIP-based image-text similarity.

\textit{Seminal works:} Lai et al. \cite{lai_lego_2024} pioneered the task of egocentric action frame generation, aiming to synthesize context-aware images of actions. To tackle this, they proposed the LEGO model, which integrates visual instruction tuning and combines image-text embeddings from a vision-language model (VLLM) as conditions for a latent diffusion model \cite{rombach_highresolution_2022}. This approach enhances action semantics and mitigates the domain gap between egocentric data and traditional generative datasets.

\section{Object Understanding Tasks}\label{Object
Understanding Related Tasks}
Object understanding tasks focus on modeling external entities that the subject interacts with, including instance object recognition, human identity recognition, trajectory prediction, and social understanding. While object and identity recognition focus on identifying “what” and “who” is present, trajectory prediction addresses “where” others may move. Social understanding complements these by modeling “how” individuals interact—capturing attention alignment, language-vision correlation, and interaction dynamics—thus enabling richer interpretation of egocentric social scenes.
\subsection{Social Perception}\label{Social Perception}

In daily life, human interaction is pervasive and fundamental to social behavior, spanning activities such as family conversations, classroom discussions, workplace meetings, and group outings. With the rise of virtual and augmented reality (VR/AR), there is growing potential to model and analyze such interactions in immersive digital environments, where both human-human and human-AI interactions coexist. In fields like education and healthcare, egocentric analysis of social behavior can enable personalized support and early detection of conditions such as Autism Spectrum Disorder (ASD).

\begin{figure}[ht]%
  \centering
  \includegraphics[width=0.5\textwidth]{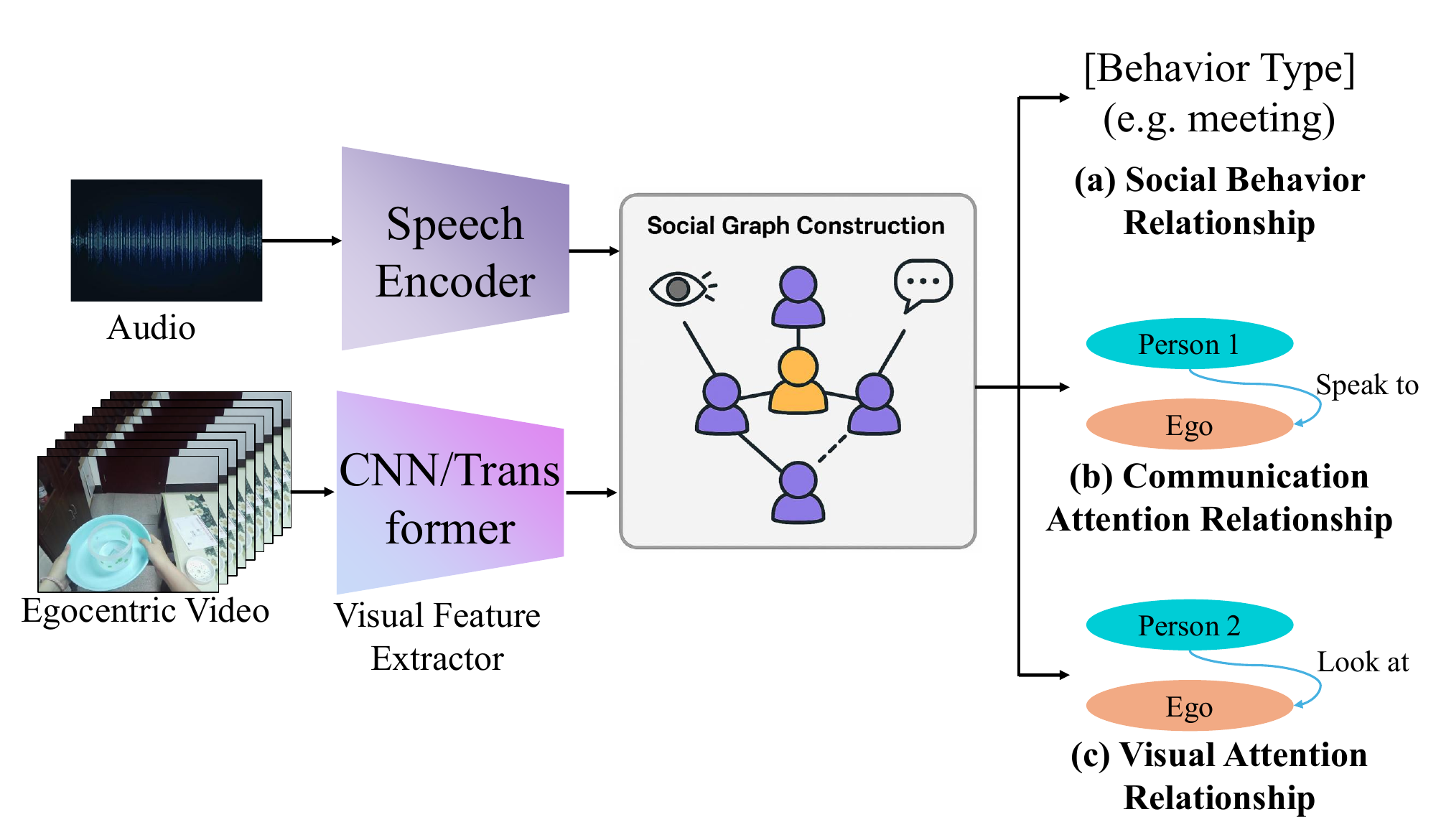}
  \caption{A typical framework for egocentric social understanding. Visual and optional speech features are extracted to model social relationships among individuals via a graph-based interaction module. The model outputs (a)social interaction categories and (b, c)inferred intentions within the scene.}\label{Social}
\end{figure}

The core goal of egocentric social scene understanding is to interpret diverse social behaviors and communication patterns from a first-person perspective. Based on task objectives, this domain can be categorized into social behavior relationship understanding, communication attention analysis, and visual attention modeling, as illustrated in Fig.~\ref{Social}. Performance comparison between representative methods is shown in Table~\ref{tab:social}.

\begin{table}[ht]
  \centering
  \footnotesize
  \caption{Performance comparison of representative methods for egocentric social perception.}
  \label{tab:social}
  \begin{tabular}{lcc}
  \toprule
  \textbf{Methods} & \multicolumn{2}{c}{\textbf{Datasets}} \\
  \Xhline{0.4pt}
  \addlinespace[3pt]
  \textit{Talk to Me} &   \textbf{EasyCom~\cite{donley_easycom_2021}} mAP (\% $\uparrow$)  &  \textbf{Ego 4D~\cite{grauman_ego4d_2022}} mAP (\% $\uparrow$)\\
  Ex2Eg-MAE(2024) \cite{tran_ex2egmae_2024} & \textbf{85.40}  & 68.10 \\
  SICNet(2024) \cite{kong_longterm_2024} & —  & \textbf{68.98}\\
  
  \Xhline{0.4pt} 
  \addlinespace[1pt]
  \Xhline{0.4pt}
  \addlinespace[3pt]

  \textit{Active Speaker Detection} & \textbf{EasyCom~\cite{donley_easycom_2021}} mAP (\% $\uparrow$) &  \textbf{EgoCom~\cite{northcutt_egocom_2023}} mAP (\% $\uparrow$) \\
  Ex2Eg-MAE(2024) \cite{tran_ex2egmae_2024} & \textbf{93.20}  & — \\
  MuST(2024) \cite{yun_spherical_2024} & 89.88  & — \\
  Majumder et al.(2024) \cite{majumder_learning_2024} & 70.20  & \textbf{65.60} \\

  \Xhline{0.4pt} 
  \addlinespace[1pt]
  \Xhline{0.4pt}
  \addlinespace[3pt]

  \textit{Look at Me} & \textbf{Ego 4D~\cite{grauman_ego4d_2022}} mAP (\% $\uparrow$) &  Acc. (\% $\uparrow$) \\
  PCIE LAM(2024) \cite{lertniphonphan_pcie_lam_2024} & \textbf{81.00}  & 93.00 \\
  Ex2Eg-MAE(2024) \cite{tran_ex2egmae_2024} & 78.30  & \textbf{93.50} \\

  \bottomrule
  \end{tabular}
  \vspace{1mm}
  \begin{flushright}
  \end{flushright}
\end{table}

\subsubsection{Social Behavior Relationship}

In egocentric social scenes, models must first capture the overall interaction dynamics before recognizing and categorizing social behaviors. This macro-level task, termed \textit{Social Behavior Relationship Understanding}, focuses on identifying salient behaviors and extracting social patterns across entire interaction sequences. Classification models are typically used for this task, and common evaluation \textit{metrics} include Area Under the ROC Curve (AUC), Top-k precision, and recall.

\textit{Seminal works:} These works can be broadly categorized into: (1) cue-based modeling for social signal perception, (2) relational modeling between agents and objects, and (3) task-specific frameworks for complex social scenarios like persuasion and multi-agent dynamics. Duarte et al. \cite{duarte_action_2018} explored how non-verbal cues like gaze, head pose, and gestures influence social behavior prediction. Li et al. \cite{li_deep_2019} emphasized dual relationships between subjects and objects using an Interactive LSTM, moving beyond isolated action modeling. Lai et al. \cite{lai_werewolf_2023} targeted persuasive behavior in social games by integrating textual and visual modalities. Grimaldi et al. \cite{grimaldi_am_2024} addressed egocentric engagement recognition in multi-user human-robot interaction.

\subsubsection{Communication Attention Relationship}
In egocentric social scenes, a key task is to model linguistic interactions between the subject and individuals in view. We define this as the \textbf{Communication Attention Relationship} task, which involves identifying active speakers, determining who is communicating with the subject, and analyzing interactions between others. This task supports deeper understanding of conversational dynamics in first-person views. Common evaluation \textit{metrics} include mAP and Top-1 Accuracy.

The reviewed works in communication attention relationship understanding can be grouped into four main categories based on their methodological focus and the specific challenges they address. Some approaches leverage\textbf{physical cues} such as head motion or facial expressions to infer auditory attention or hearing-related conditions~\cite{lu_sound_2022,yin_hearing_2024}, emphasizing the role of low-level physiological signals in speaker localization. \textbf{Multimodal fusion} techniques integrate audio and visual inputs to localize active speakers or model conversational dynamics~\cite{jiang_egocentric_2022,wenqijia_audiovisual_2024}, focusing on speaker activity recognition within egocentric social interactions. Third, \textbf{data augmentation and domain adaptation} strategies~\cite{tran_ex2egmae_2024} address the scarcity of large-scale egocentric datasets by synthesizing egocentric views from exocentric videos for pretraining. Finally, models incorporating \textbf{long-term temporal context}~\cite{kong_longterm_2024,yun_spherical_2024,majumder_learning_2024} aim to understand extended conversations or dynamic spatial configurations by employing attention mechanisms, spherical embeddings, or audio reconstruction objectives.

\subsubsection{Visual Attention Relationship}

In social interactions, gaze behavior serves as a crucial indicator of interpersonal attention and engagement. This \textbf{visual attention relationship} task focuses on modeling mutual gaze and social attention patterns among individuals in egocentric scenes. It aims to automatically detect who is looking at the subject and infer gaze relationships between other individuals, using computer vision techniques. This typically involves analyzing fine-grained facial cues—such as eye movements and expressions—from the egocentric perspective to estimate attention direction and social engagement. Common evaluation \textit{metrics} include mAP and Top-1 accuracy.

\textit{Seminal works:} Research in visual attention relationship modeling primarily centers on detecting eye contact and gaze behavior in egocentric videos. Early efforts, such as \cite{chong_detection_2020}, demonstrated that deep convolutional models could reach expert-level performance in automatic eye contact detection. More recent approaches expanded to multi-task learning frameworks, like EgoT2 \cite{xue_egocentric_2023}, which jointly optimize various egocentric tasks, including gaze recognition. State-of-the-art methods differ significantly in technical focus: Tran et al. \cite{tran_ex2egmae_2024} leveraged self-supervised masked autoencoding and face synthesis for pretraining, excelling in tasks like Look-at-me detection; while Lertniphonphan et al. \cite{lertniphonphan_pcie_lam_2024} integrated Bi-LSTM and gaze smoothing techniques within an ensemble architecture to refine temporal consistency and reduce noise. These approaches reflect two key methodological trends, \textbf{pretraining} via large-scale synthetic or augmented data versus \textbf{temporal modeling} with lightweight post-processing—for enhancing gaze interaction modeling in egocentric settings.

\subsection{Human Identity and Trajectory Recognition}
\begin{figure}[ht]%
  \centering
  \includegraphics[width=0.5\textwidth]{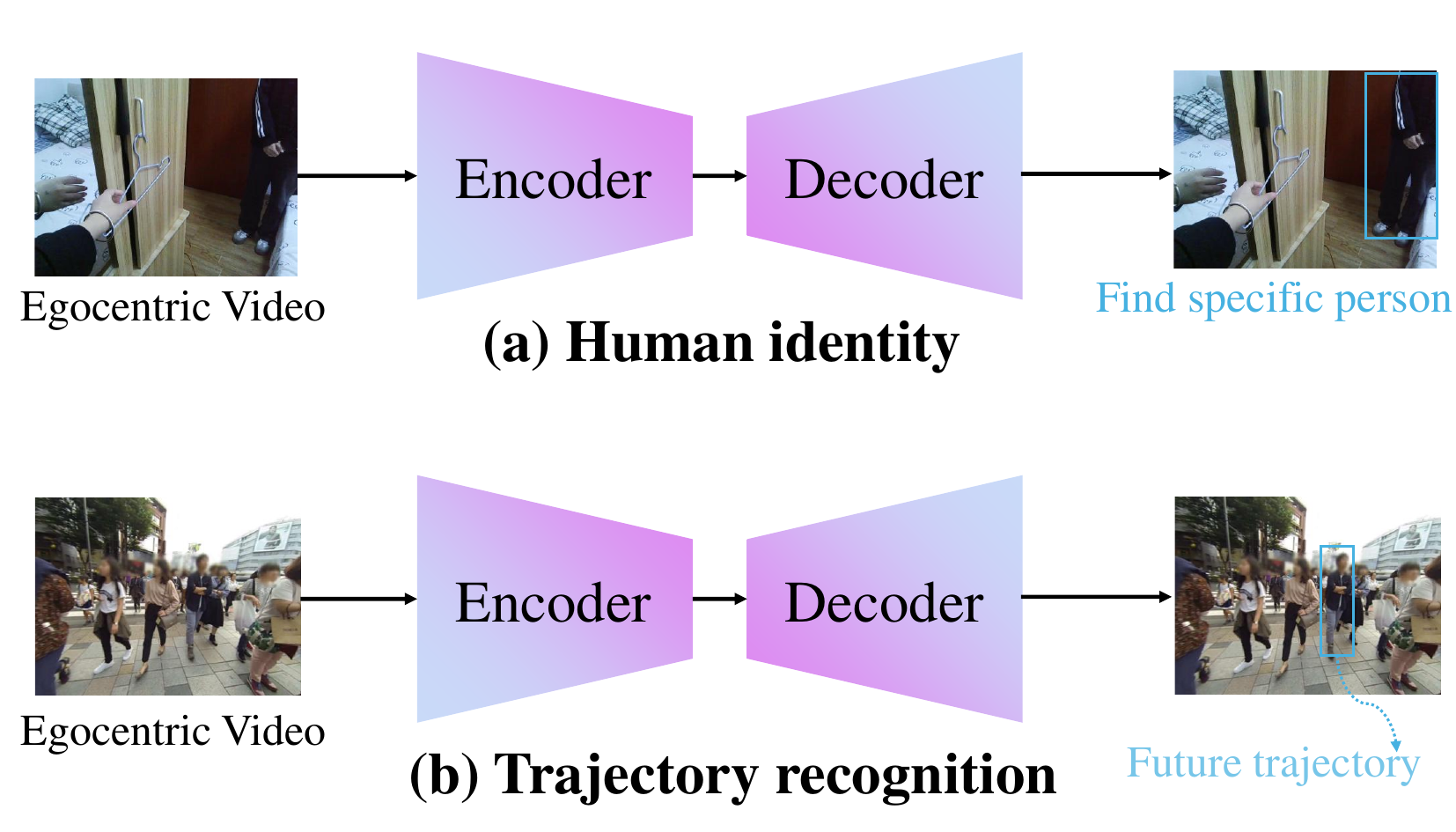}
  \caption{llustration of (a)human identity and (b)trajectory recognition in egocentric vision. The task involves recognizing individual identities and predicting their movement trajectories over time from the first-person perspective.}\label{Human}
\end{figure}

In egocentric vision, the identification and tracking of human objects in the subject’s field of view are essential for understanding social dynamics and enabling downstream tasks such as intention prediction and safety monitoring (Fig.~\ref{Human}). These tasks include facial recognition, person re-identification (Person Re-ID), and trajectory recognition. Facial recognition focuses on matching facial features against a known identity set, while Person Re-ID emphasizes full-body features across varying poses and appearances in open-world settings. In contrast, trajectory recognition aims to track and analyze the spatiotemporal motion of individuals over time to infer behavior patterns or predict potential collisions. These tasks rely on robust modeling of visual cues, spatial-temporal dependencies, and motion dynamics. Common \textit{evaluation metrics} include matching accuracy, rank-based retrieval metrics (e.g., Top-k accuracy), and trajectory similarity measures such as average displacement error.

\textit{Seminal works:} Existing research on egocentric human object understanding primarily focuses on two key directions: identity recognition and trajectory prediction. For identity recognition, Choudhary et al.~\cite{choudhary_domain_2021} addressed the domain gap between egocentric and fixed-camera datasets by introducing a Neural Style Transfer (NST)-based domain adaptation technique to generate hybrid-style data and reduce bias. In trajectory recognition, Yagi et al.~\cite{yagi_future_2018} proposed a multi-stream encoder-decoder framework that jointly models self-motion, scale, and pose to predict future person locations. Building on this, Chen et al.~\cite{chen_future_2023} incorporated pose, depth, and relational cues into a unified tensor representation to enhance learning richness and reduce prediction errors. These methods highlight the importance of integrating egocentric-specific motion and appearance cues to improve person-centric modeling.

\subsection{Instance Object Recognition}\label{Instance Object Recognition}

\begin{figure}[ht]%
  \centering
  \includegraphics[width=0.5\textwidth]{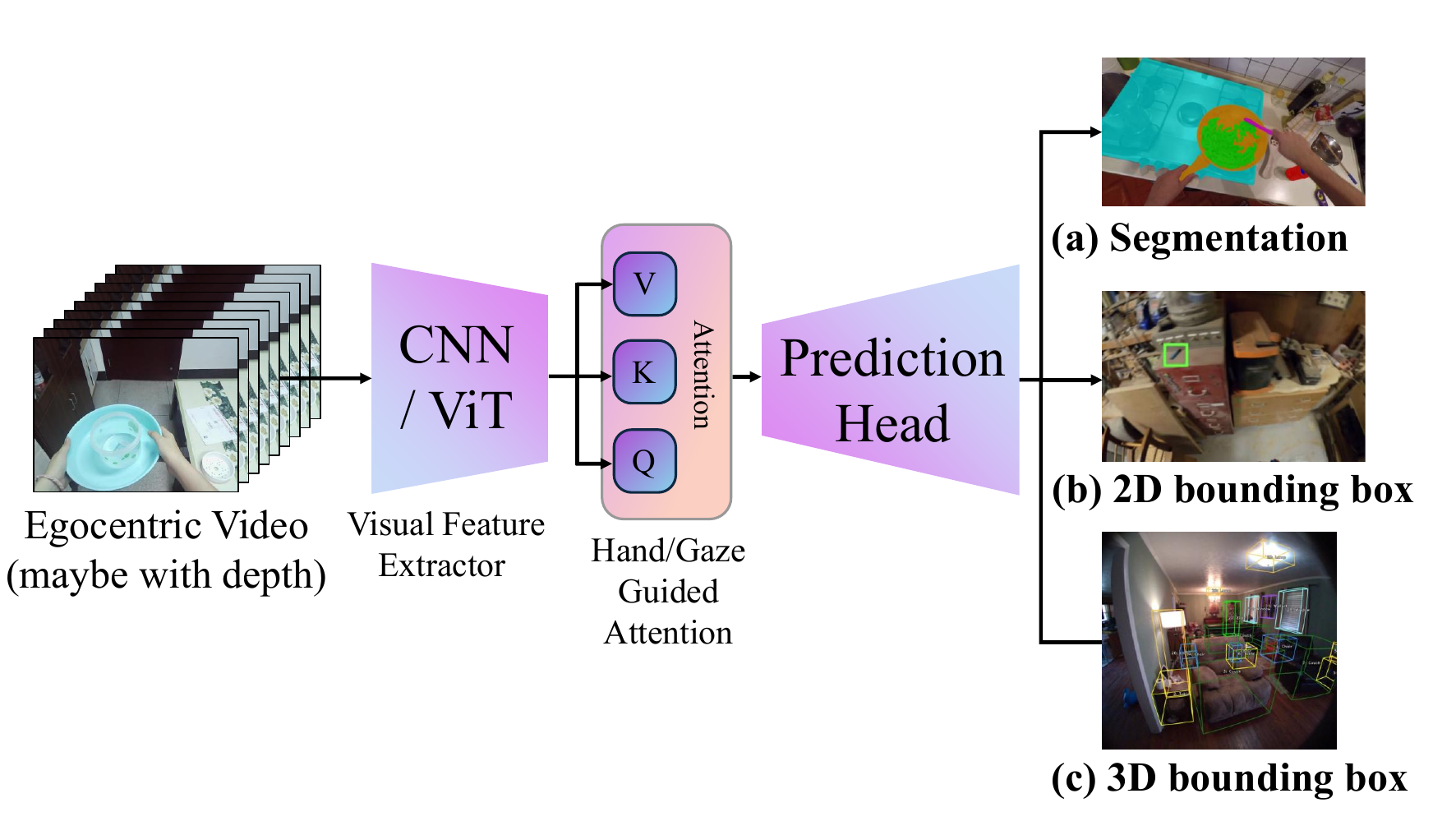}
  \caption{A typical pipeline for egocentric instance object identification. Visual features are extracted from first-person frames and enhanced via optional hand or gaze-guided attention for (a)instance segmentation and (b)2D/3D bounding box regression.}\label{Object}
\end{figure}

In egocentric visual scenes, instance objects exhibit significant diversity and contextual dependency. These include daily items like tableware, industrial tools, and household appliances, which often undergo dynamic spatial and temporal changes. Their interaction with the subject, along with properties such as material and state transitions, forms the core of recognition tasks, such as the vegetables a chef is about to pick up or the food in the pot that is about to burn.  Egocentric instance object recognition primarily aims to localize and identify objects—especially those under interaction—based on spatial-temporal cues, as shown in Fig.~\ref{Object}, performance comparison of representative methods is shown in Table~\ref{tab:object}. Tasks are commonly categorized into object detection using bounding boxes, semantic segmentation for precise boundaries, and further extended to 3D recognition and reconstruction in dynamic scenes. Evaluation \textit{metrics} include mIOU, spatio-temporal average precision (stAP), and temporal average precision (tAP). 

\begin{table}[ht]
  \centering
  \footnotesize
  \caption{Performance comparison of representative methods for egocentric object recognition.}
  \label{tab:object}
  \begin{tabular}{lcc}
  \toprule
  \textbf{Methods} & \multicolumn{2}{c}{\textbf{Datasets}} \\
  \Xhline{0.4pt}
  \addlinespace[3pt]
  \textit{2D Detection} &   \textbf{EgoObjects~\cite{zhu_egoobjects_2023}} AP50 (\% $\uparrow$)  & AR10 (\% $\uparrow$)\\
  DEVI(2023)\textsuperscript{†}  \cite{akiva_selfsupervised_2023} & 14.96\textsuperscript{†}  & 29.61\textsuperscript{†} \\
  Efficient-CLS(2023) \cite{wu_labelefficient_2023} & \textbf{61.01}  & —\\
  
  \Xhline{0.4pt} 
  \addlinespace[3pt]

  \textit{2D Segmentation} & \textbf{VISOR~\cite{darkhalil_epickitchens_2022}} mIOU (\% $\uparrow$) &  \textbf{VOST~\cite{tokmakov_breaking_2023}} mIOU (\% $\uparrow$) \\
  ActionVOS(2024) \cite{ouyang_actionvos_2024} & \textbf{68.20}  & \textbf{32.30} \\
  NVOS(2024)\textsuperscript{‡} \cite{yuhanshen_learning_2024} & 38.70\textsuperscript{‡}  & 23.20\textsuperscript{‡} \\

  \Xhline{0.4pt}
  \addlinespace[3pt]

  \textit{VQ2D} & \textbf{Ego 4D~\cite{grauman_ego4d_2022}} $\mathrm{tAP_{25}}$ (\% $\uparrow$) & $\mathrm{stAP_{25}}$ (\% $\uparrow$) \\
  VQLoC(2023) \cite{jiang_singlestage_2023} & 32.00  & 24.00 \\
  CocoFormer(2023) \cite{xu_where_2023} & 27.00 & 20.00 \\
  RELOCATE(2024) \cite{khosla_relocate_2024} & \textbf{35.00} & \textbf{43.00} \\

  \Xhline{0.4pt}
  \addlinespace[3pt]

  \textit{VQ3D} & \textbf{Ego 4D~\cite{grauman_ego4d_2022}} $\mathrm{L2}$ ($\downarrow$) & $\mathrm{angel}$ ( $\downarrow$) \\
  CocoFormer(2023) \cite{xu_where_2023} & 4.46 & 1.23 \\
  EgoLoc(2023) \cite{mai_egoloc_2023}  & \textbf{1.86} & \textbf{0.92} \\

  \bottomrule
  \end{tabular}
  \vspace{1mm}
  \begin{flushright}
    \textsuperscript{†} Self supervised\\
    \textsuperscript{‡} Weakly supervised, 38.70 is IOU, 23.20 is IOU Union.
  \end{flushright}
\end{table}

\textit{Seminal works:} Existing works on egocentric object recognition adopt diverse approaches to tackle the spatial-temporal complexity and multimodal nature of interactive instances. Some studies focus on self-supervised or label-efficient detection, such as DEVI~\cite{akiva_selfsupervised_2023} and Efficient-CLS~\cite{wu_labelefficient_2023}, which introduce biologically inspired mechanisms or leverage multi-view training. Other methods enhance cross-modal understanding through audio-visual alignment~\cite{huang_egocentric_2023, shi_crossmodal_2025}, or exploit hand-object interaction cues and action semantics for object localization~\cite{zhang_helping_2023, ouyang_actionvos_2024, xu_weakly_2024}. Furthermore, object state changes~\cite{yu_video_2023} and weakly supervised functional region recognition~\cite{xu_weakly_2024} represent finer-grained modeling perspectives.

In the context of language-conditioned object localization, VQL~\cite{grauman_ego4d_2022} and follow-ups like VQLoC~\cite{jiang_singlestage_2023} enable open-ended object queries from natural language. In parallel, 3D segmentation methods~\cite{tschernezki_neuraldiff_2021, xu_where_2023} focus on dynamic scene decomposition and bias reduction using advanced neural rendering and transformer-based architectures. Tracking-focused studies~\cite{huang_tracking_2023, zhao_instance_2024} address motion blur and viewpoint shift through patch-based memory or 3D coordinate-level registration protocols.

\textit{SOTA:} Recent efforts like RELOCATE~\cite{khosla_relocate_2024} introduce training-free region-based frameworks for efficient long-video object retrieval, while EgoLoc~\cite{mai_egoloc_2023} enhances 3D localization using SfM and multi-view weighting strategies. Open-world and zero-shot scenarios have also gained attention: EgoLifter~\cite{gu_egolifter_2024} utilizes SAM \cite{kirillov_segment_2023} masks and 3D Gaussian representation for promptable open-world segmentation, and ROSA~\cite{yuhanshen_learning_2024} aligns narration with object masks through contrastive learning, achieving zero-shot pixel-level localization on two egocentric video datasets (VISOR-NVOS and VOST \cite{tokmakov_breaking_2023}).

\section{Environment Understanding Tasks}\label{Environment Understanding Related Tasks}
Environmental understanding tasks focus on constructing spatial awareness and navigational context. Key subtasks include SLAM-based mapping and scene localization. These tasks are inherently connected: SLAM builds a dynamic map, while localization interprets the subject's position within it, supporting downstream tasks that require environment-aware modeling.
\subsection{Environment Modeling} \label{Environment Modeling}
\begin{figure}[ht]%
  \centering
  \includegraphics[width=0.5\textwidth]{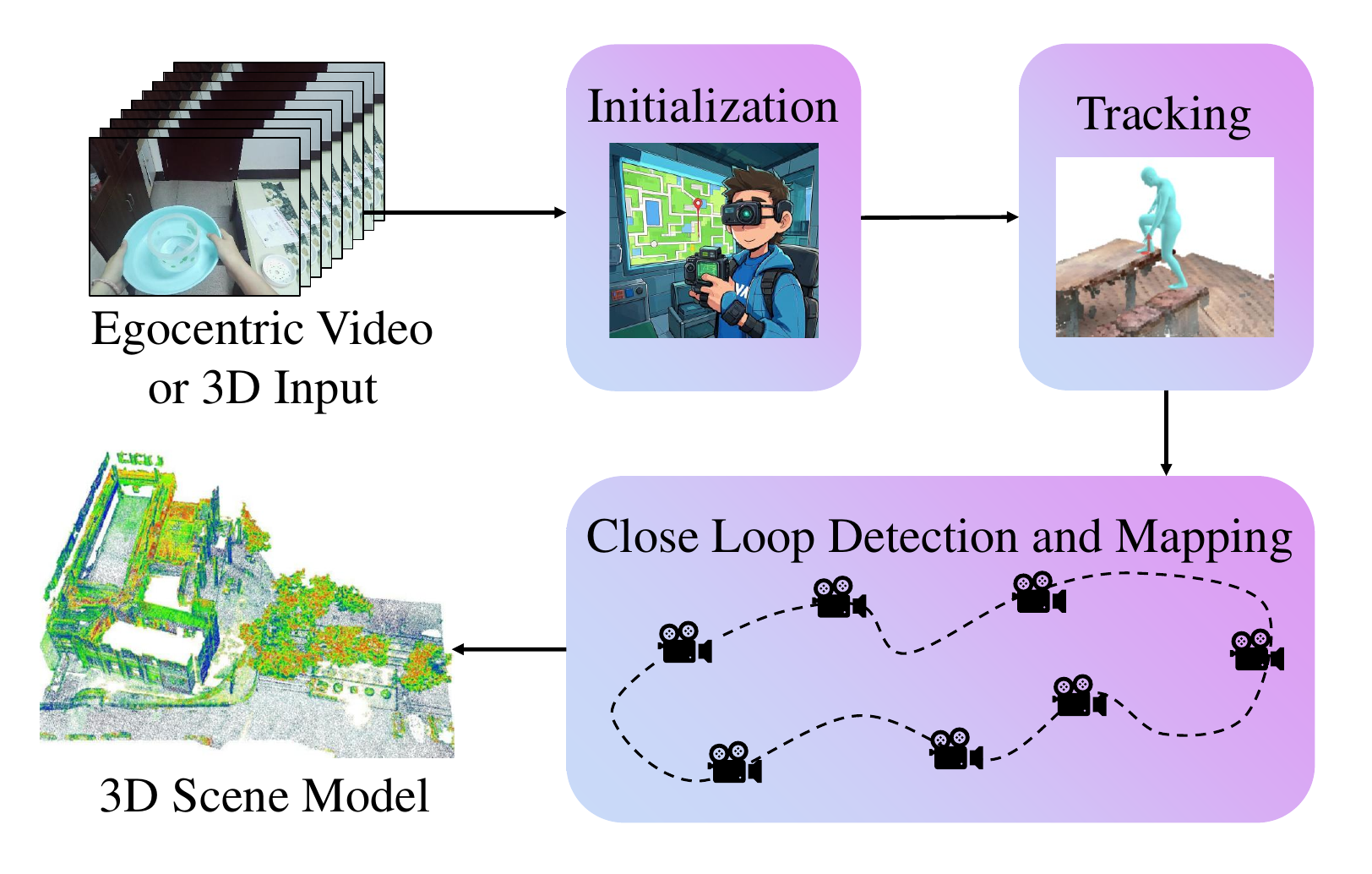}
  \caption{Environment modeling focuses on understanding spatial layouts and structures of environment from the first-person perspective.}\label{Modeling}
\end{figure}

In egocentric environmental understanding, \textbf{environment modeling} aims to build accurate representations of the surrounding scene, as illustrated in Fig.~\ref{Modeling}. Depending on the task, this can range from sparse local features to dense 3D reconstructions, often relying on Simultaneous Localization and Mapping (SLAM) techniques. SLAM provides foundational perception capabilities for tasks like autonomous navigation and path planning, especially in robotics.

Visual SLAM (V-SLAM) systems generally follow three stages: initialization (setting a global coordinate frame and initial map), tracking (real-time pose estimation), and mapping (constructing sparse/dense 3D representations via loop closure). 

While SLAM has achieved notable progress in static settings, dynamic environments—typical in egocentric data—pose challenges such as head motion-induced errors and scale drift. Additionally, wearable devices impose constraints on computational cost and real-time performance, requiring more lightweight and robust solutions.

\textit{Seminal works} have explored egocentric environment modeling from various perspectives. Hübner et al.~\cite{hubner_evaluation_2020} evaluated the mapping performance of commercial devices like Microsoft HoloLens. To achieve real-time localization and motion capture with minimal hardware, Yi et al.~\cite{yi_egolocate_2023} introduced EgoLocate, which fuses six IMUs and a body camera. Meanwhile, Rosinol et al.~\cite{rosinol_nerfslam_2023} addressed dense scene reconstruction by combining monocular SLAM with neural radiance fields (NeRF)~\cite{mildenhall_nerf_2021}, enabling volumetric modeling.

Recent advances further integrate human motion estimation with SLAM. Yin et al.~\cite{yin_egohdm_2024} proposed EgoHDM, an online egocentric SLAM system based on monocular RGB and inertial data. It jointly optimizes human pose and environment reconstruction via a novel visual-inertial motion (VIM) initialization that incorporates body shape constraints, achieving globally consistent motion and high accuracy on public benchmarks~\cite{trumble_total_2017,guzov_human_2021}. Performance is compared in Tab. \ref{tab:environment}.

\begin{table}[ht]
  \centering
  \footnotesize
  \caption{Performance comparison of representative methods for egocentric environment modeling.}
  \label{tab:environment}
  \begin{tabular}{lcc}
  \toprule
  \textbf{Methods} & \multicolumn{2}{c}{\textbf{Datasets}} \\
  \Xhline{0.4pt}
  \addlinespace[3pt]
  \textit{Modeling} & \textbf{TotalCapture~\cite{trumble_total_2017}} error ($\downarrow$) &  \textbf{HPS~\cite{guzov_human_2021}} error ($\downarrow$) \\
  EgoLocate(2023) \cite{yi_egolocate_2023} & 0.22  & 1.70 \\
  EgoHDM(2024) \cite{yin_egohdm_2024} & \textbf{0.13}  & \textbf{1.50} \\
  \bottomrule
  \end{tabular}
\end{table}

\subsection{Scene Localization} \label{Scene Localization}

\begin{figure}[ht]%
  \centering
  \includegraphics[width=0.5\textwidth]{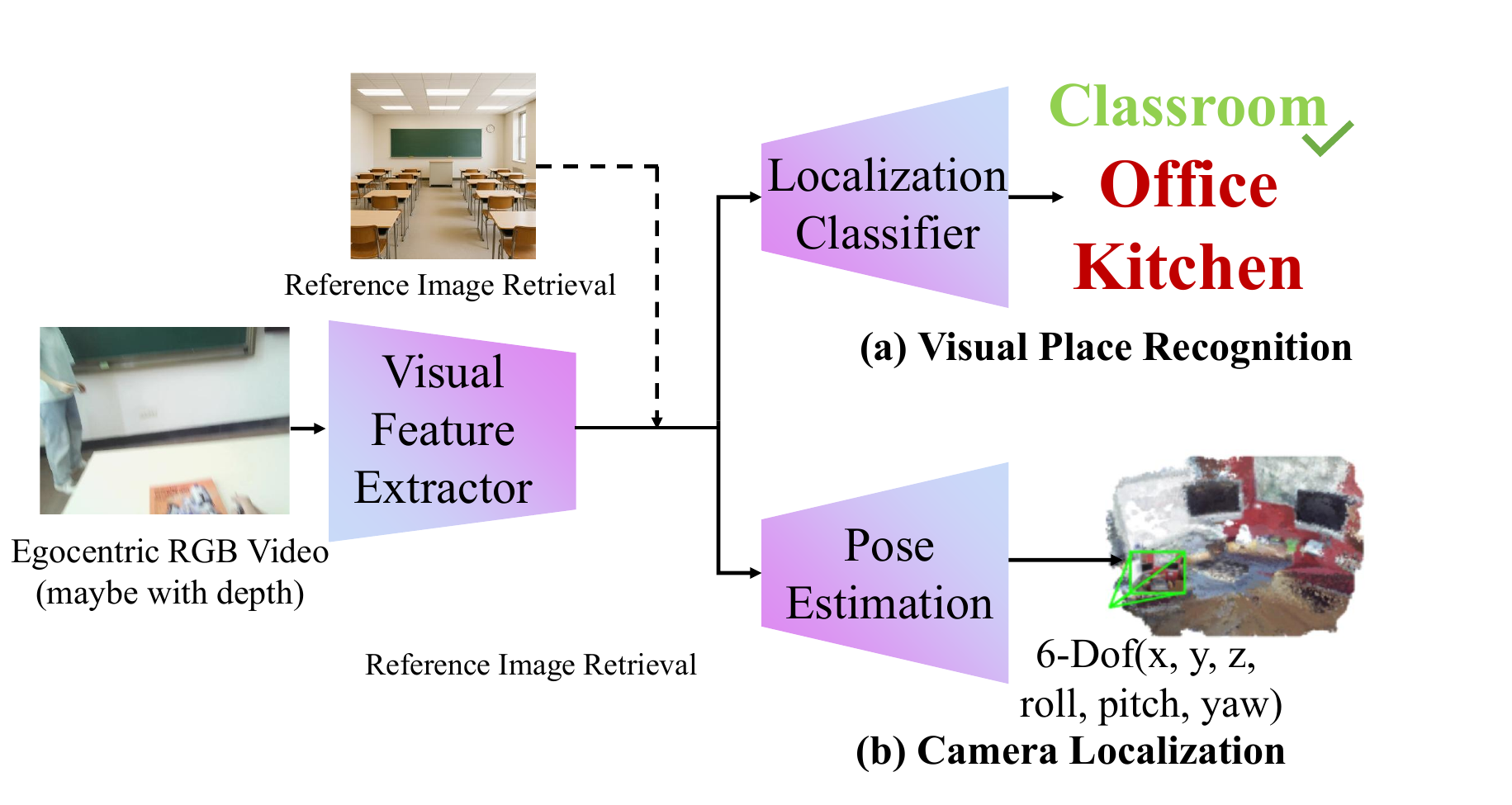}
  \caption{An example pipeline for egocentric scene localization. Visual features extracted from video frames are used to retrieve or match against a reference database or map, and fused with optional geometric to classify the place or estimate the subject’s position.}\label{Localization}
\end{figure}

Egocentric scene localization aims to understand environments by identifying and analyzing visual features, with the goal of mapping observations to known spatial locations (Fig.~\ref{Localization}). This task can be divided into two components: \textit{visual place recognition} (VPR), which focuses on identifying specific indoor or outdoor scenes, and \textit{camera localization}, which estimates the camera’s position and orientation based on visual and temporal cues. VPR is typically framed as a classification task, evaluated by Recall@N. Camera localization, by contrast, is a regression problem evaluated using \textbf{Median Position Error} (meters) and \textbf{Median Orientation Error} (degrees), measuring spatial and directional accuracy, respectively.

\textit{Seminal works: }Recent advances in egocentric scene localization span both visual place recognition (VPR) and camera pose estimation. For large-scale geo-localization, Berton et al.~\cite{berton_rethinking_2022} proposed the SF-XL dataset and CosPlace for scalable training, while Ali-Bey et al.~\cite{ali-bey_mixvpr_2023} focused on accelerating global feature aggregation. Suveges et al.~\cite{suveges_unsupervised_2025} introduced online unsupervised learning of semantic maps, and Huang et al.~\cite{huang_automatic_2024} explored food environment recognition as a novel egocentric VPR task.

In camera localization, researchers have addressed challenges such as scene scalability, storage, and generalization. Blanton et al.~\cite{blanton_extending_2020} and Shavit et al.~\cite{shavit_learning_2021} developed pose regression networks supporting multiple scenes using CNN and Transformer architectures. Do et al.~\cite{do_learning_2022} leveraged sparse implicit 3D scene encoding to reduce memory use, while Panek et al.~\cite{panek_meshloc_2022} proposed mesh-based localization for improved efficiency.

Recent state-of-the-art approaches further improve accuracy and efficiency. Zhu et al.~\cite{zhu_r2_2023} proposed R2Former, a Transformer-based model that unifies feature correlation and spatial attention for robust VPR, achieving top performance on major datasets. Lin et al.~\cite{lin_exploring_2024} introduced a dual-branch framework combining keypoint-guided and full-scene coordinate regression, significantly enhancing pose accuracy across diverse benchmarks.

\section{Hybrid Understanding Tasks}\label{Hybrid Understanding Related Tasks}
Hybrid understanding tasks integrate subject, object, and environmental cues to enable high-level semantic understanding. Subtasks such as video summarization, multi-view fusion, and video question answering require reasoning across modalities and scene elements. These tasks often build upon lower-level subtasks, forming a comprehensive pipeline for complex egocentric understanding.
\subsection{Content Summary}\label{Content Summary}
\begin{figure}[ht]%
  \centering
  \includegraphics[width=0.5\textwidth]{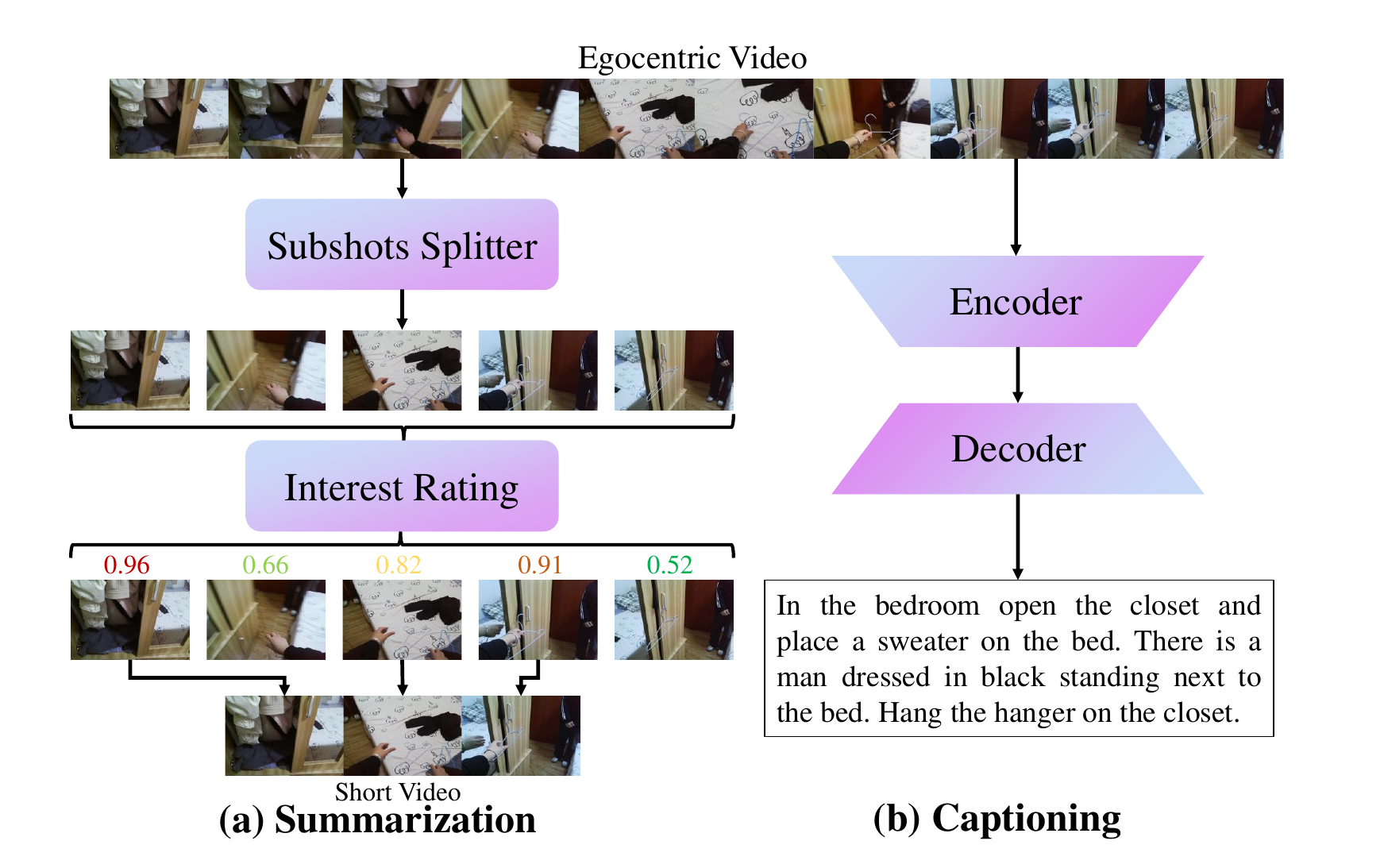}
  \caption{Content summary is to extract key segments or high-level semantics from long-form first-person video to enable efficient browsing, retrieval, or downstream understanding.}\label{Summary}
\end{figure}

Egocentric content summarization includes multiple task forms such as traditional video summarization and video captioning~\cite{xiong2025adaptively}, as illustrated in Fig.~\ref{Summary}. A key challenge is the absence of unified evaluation criteria due to the subjectivity of human annotations and personalized, dynamic user preferences, which complicate supervision and model generalization. In practice, summarization systems offer practical value in platforms like cloud storage and video sharing, where automatic summaries enhance data management and content retrieval. The F1-score is commonly used as the main evaluation \textit{metric}.

\textit{Seminal works:} Recent progress in egocentric video summarization includes unsupervised reinforcement learning approaches~\cite{nagar_generating_2021} and multistream summarization for dynamic camera inputs~\cite{elfeki_multistream_2022}. To address \textbf{personalization}, several works leverage multimodal user cues: audio-visual salience~\cite{furlan_fast_2018}, textual social data~\cite{ramos_personalizing_2020}, and interactive inputs such as text and video segments~\cite{wu_intentvizor_2022}. Co-summarization with auxiliary videos~\cite{sahu_egocentric_2023} further enhances temporal efficiency.

Beyond summarization, captioning tasks generate textual descriptions for egocentric data. Dai et al.~\cite{dai_egocap_2024} introduced the EgoCap dataset for first-person image captioning, while Qiu et al.~\cite{qiu_egocentric_2024} built the EgoDIMCAP dietary caption dataset to estimate food intake. Novel sensing strategies such as acoustic-guided smart glasses~\cite{parikh_echoguide_2024} improve recording efficiency in dietary monitoring.

Recent state-of-the-art methods combine \textbf{vision-language pretraining and contrastive learning}. He et al.~\cite{he_align_2023} proposed A2Summ, a Transformer-based multimodal summarization model using aligned embeddings. Xu et al.~\cite{xu_retrievalaugmented_2024} introduced a retrieval-augmented captioning framework that enhances egocentric descriptions with external third-person video references.

\subsection{Multiview Joint Understanding}\label{Multiview Joint Understanding}

\begin{figure}[ht]%
  \centering
  \includegraphics[width=0.5\textwidth]{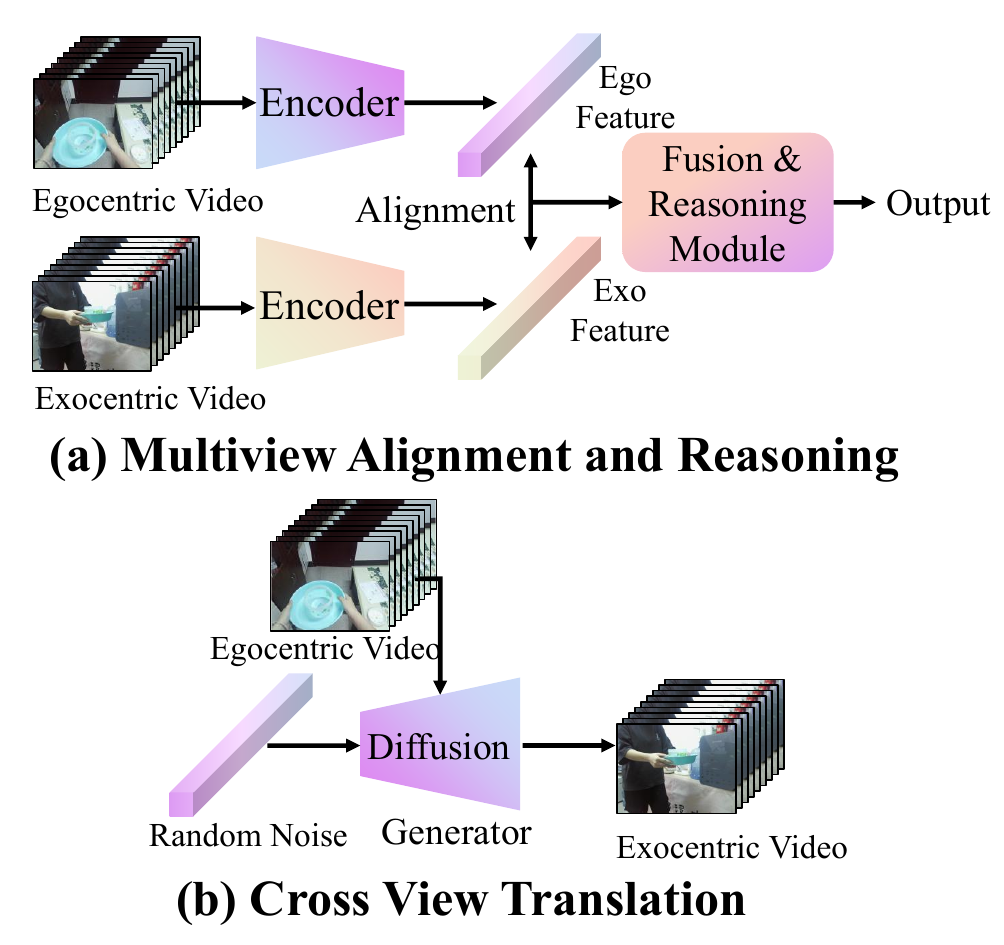}
  \caption{(a) Multiview alignment and reasoning aims to integrate and align information across multiple views for joint inference and relational understanding. (b) Cross-view translation focuses on generating or interpreting one view (e.g., third-person) from another (e.g., egocentric), enabling cross-perspective representation learning.}\label{Multiview}
\end{figure}

As egocentric scene understanding advances, researchers have increasingly explored multi-view joint understanding, as illustrated in Fig.~\ref{Multiview}. This includes both collaborative analysis of multiple egocentric views and integration with exocentric (third-person) data captured by static or aerial cameras.

A core challenge lies in aligning egocentric and exocentric perspectives to learn viewpoint-invariant features for downstream tasks such as cross-view matching and transformation. First, there are significant visual differences between Ego and Exo perspectives. Even when recording the same activity synchronously, the focus of attention varies greatly between perspectives. The egocentric view emphasizes the wearer's hand movements and object interactions, while the exocentric view provides more comprehensive background information and global context. Second, existing methods are mostly limited to coarse-grained action recognition, making it difficult to achieve fine-grained alignment of action stages. Additionally, obtaining precisely annotated synchronized Ego-Exo paired data is costly, severely limiting the development of multi-view joint understanding research.

\textit{Seminal works:} Recent efforts in multi-view egocentric understanding primarily focus on \textbf{bridging the gap} between egocentric (Ego) and exocentric (Exo) perspectives. Han et al.~\cite{han_benchmarking_2024} and Yang et al.~\cite{fanyang_yowo_2024} explored top-view surveillance and 6-DoF camera pose estimation to enable spatial alignment. To overcome the scarcity of egocentric data, many works transfer knowledge from large-scale exocentric datasets using unpaired training. Li et al.~\cite{li_egoexo_2021} used egocentric signal distillation for cross-view pretraining, while Wang et al.~\cite{wang_learning_2023} constructed pseudo Ego-Exo pairs via LLM-based captioning. Temporal misalignment is addressed through self-supervised temporal alignment~\cite{xue_learning_2023} and contrastive representation learning guided by vision-language models~\cite{jang_intra_2024,xu_weakly_2024}. 

Recent methods also exploit \textbf{cross-view geometric reasoning}~\cite{truong_crossview_2025} and knowledge distillation for action segmentation~\cite{quattrocchi_synchronization_2024}. In social scenes, pseudo-egocentric data generation enables egocentric reasoning without Ego data~\cite{tran_ex2egmae_2024}. Paired datasets such as Charades-Ego~\cite{sigurdsson_actor_2018} and TF2023~\cite{zhao_fusing_2024} enable weakly supervised alignment and camera wearer segmentation. With generative models, Cheng et al.~\cite{cheng_4diff_2024} and Luo et al.~\cite{miluo_put_2024} used paired Ego-Exo data to translate views via diffusion and two-stage generation frameworks.

\subsection{Video Question Answering}\label{Video Question Answering}

\begin{figure}[ht]%
  \centering
  \includegraphics[width=0.5\textwidth]{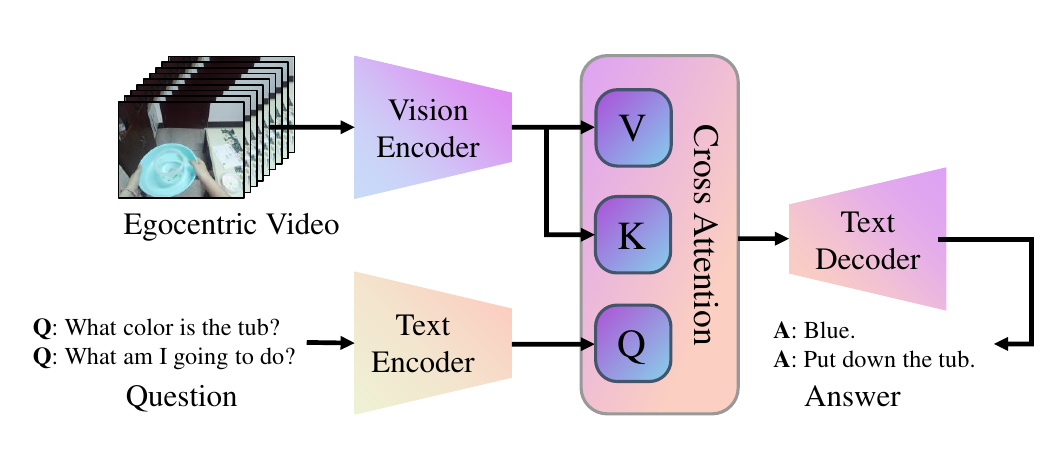}
  \caption{VQA system takes a first-person video and a natural language question as input, and generates an answer based on multimodal understanding of visual and temporal context.}\label{VQA}
\end{figure}

Egocentric Video Question Answering (Ego-VQA) is a cross-modal task that requires generating answers by jointly understanding egocentric videos and natural language questions, as shown in Fig.~\ref{VQA}. Input modalities include vision and text, and outputs are typically natural language or video-referenced responses.

Ego-VQA is commonly categorized into closed-ended QA—selecting answers from candidates (evaluated by accuracy)—and open-ended QA, which requires free-form answer generation, assessed using metrics such as ROUGE or human evaluation. The latter is more challenging due to its demand for both visual comprehension and language generation. This task has broad application potential, including embodied AI, AR, and real-time assistance in scenarios like industrial maintenance or skill training.

\textit{Seminal works:} Early efforts in Ego-VQA explored applications such as assisting visually impaired users, leading to \textbf{datasets} like EgoVQA \cite{fan_egovqa_2019} and Env-QA \cite{gao_envqa_2021}, which model diverse daily activities and interactions. EgoTaskQA \cite{jia_egotaskqa_2022} further expanded the scope to include descriptive, predictive, explanatory, and counterfactual questions, emphasizing higher-level reasoning needs.

To address the \textbf{complexity of long videos and rich temporal interactions}, recent work introduced more efficient attention mechanisms and memory structures. Gao et al. \cite{gao_mist_2023} proposed a selective attention model for dense spatiotemporal understanding, while AMEGO \cite{goletto_amego_2024} leveraged structured active memory for long-term egocentric video comprehension.

In \textbf{embodied intelligence}, researchers shifted from single-target questions \cite{yu_multitarget_2019} to multi-target reasoning and 3D environment understanding \cite{ma_sqa3d_2023,zhu_excalibur_2023}, aiming to evaluate agents' real-world scene understanding and planning ability. Beyond virtual simulations, AssistQ \cite{wong_assistq_2022} proposed affordance-centric task guidance based on egocentric views, bridging vision-language learning and actionable assistance.

Finally, as existing benchmarks focus mainly on fact recall, \textbf{newer tasks} such as Ego CVR \cite{hummel_egocvr_2024} and EgoThink \cite{cheng_egothink_2024} aim to assess higher-order capabilities like planning, reasoning, and perspective-aware retrieval, pushing VLMs toward deeper egocentric understanding.

\section{Datasets}\label{Datasets}
The rapid development of deep learning techniques has greatly advanced the field of computer vision, and this progress has largely been driven by the support of large-scale datasets. These datasets have provided rich data-driven support for model training, playing a key role in the development of pre-trained or self-supervised models \cite{radford_learning_2021}, Tong et al. \cite{tong_videomae_2022}, Lin et al. \cite{lin_egocentric_2022}, Pramanick et al. \cite{pramanick_egovlpv2_2023}, Gundavarapu et al. \cite{gundavarapu_extending_2024}. Through pre-training, we are able to systematically enhance the generalization ability of models and fine-tune them to adapt to various downstream tasks. It can be argued that datasets have become the core driving force behind computer vision research, offering a starting point for exploring new research problems and developing AI technologies that support human endeavors. The scale and diversity of these datasets directly impact their research value.

In the field of egocentric vision research, more realistic and specialized datasets have played a crucial role in technological advancements. However, compared to traditional exocentric perspective datasets, the number and scale of egocentric datasets remain relatively limited. This is primarily due to the relatively recent emergence of wearable camera technologies and the scarcity of egocentric content in online video resources. Nevertheless, in recent years, with the increasing demand for research, many high-quality, large-scale egocentric datasets have been released, providing important support for model training and research across various tasks.

In this section, we systematically review general-purpose egocentric datasets applicable to a variety of tasks and provide a detailed analysis of their characteristics. Specifically, we will focus on high-quality datasets released in recent years, which demonstrate exceptional performance in terms of scale, diversity, and task applicability. Furthermore, we present a comparative analysis of currently available public egocentric datasets in tables, covering aspects such as their application domains, data scale, and multimodal characteristics. In Table \ref{table1}, the basic characteristics of these datasets are listed. In Table \ref{table2}, we compare the available annotations across the datasets, along with the dataset homepage. In Table \ref{table3}, we list the tasks suitable for each dataset. Through this section, we aim to offer researchers a comprehensive reference on the latest egocentric datasets and provide support for the exploration of future research directions.

\begin{sidewaystable}
  \sidewaystablefn%
  \tiny
  \begin{center}
  \begin{minipage}{\textheight}
  \caption{Basic characteristics of datasets}\label{table1}
  \begin{tabularx}{\textwidth}{lllXlll}
  \toprule%
  Dataset                                                             & Scenarios               & Views        & Data Types                                               & Duration    & Sequences        & Participants     \\
  \midrule
  ADL\textcolor{red}{(2012)} \cite{pirsiavash_detecting_2012}             & Daily Life              & Ego          & RGB                                                      & 10 hours    & 20               & 20               \\
  EGTEA Gaze+\textcolor{red}{(2021)} \cite{li_eye_2021}                   & Cooking                 & Ego          & RGB, gaze                                                & 28  hours   & 86               & 32               \\
  EPIC-KITCHENS-100\textcolor{red}{(2022)}  \cite{damen_rescaling_2022}  & Cooking                 & Ego          & RGB, audio, flow                                         & 100  hours  & 700              & 37               \\
  Ego4D\textcolor{red}{(2022)} \cite{grauman_ego4d_2022}                  & Multiple                & Ego or Multi & RGB, audio, 3D, gaze, IMU                                & 3670  hours & 9650             & 931              \\
  HOI4D\textcolor{red}{(2022)} \cite{liu_hoi4d_2022}                      & Indoor                  & Ego          & RGB, depth                                               & 22.2  hours & 4000             & 9                \\
  AssemblyHands\textcolor{red}{(2023)}  \cite{ohkawa_assemblyhands_2023} & Toy Assembly            & Ego and Exo  & RGB                                                      & 490k frames & 62               & 34               \\
  UnrealEgo2\textcolor{red}{(2024)} \cite{akada_3d_2024}                  & Virtual Scene           & Stereo Ego   & RGB, depth                                               & 14 hours    & 15,207           & 17               \\
  EgoPet\textcolor{red}{(2024)} \cite{bar_egopet_2024}                    & Animal Activity         & Ego          & RGB                                                      & 84 hours    & 819              & Animals          \\
  WEAR\textcolor{red}{(2024)} \cite{bock_wear_2024}                       & Outdoor                 & Ego          & RGB, IMU                                                 & 15 hours    & 24               & 22               \\
  EgoThink\textcolor{red}{(2024)} \cite{cheng_egothink_2024}              & Multiple                & Ego          & RGB                                                      & 700 images  & \textbackslash{} & \textbackslash{} \\
  Ego-Exo4D\textcolor{red}{(2024)} \cite{grauman_egoexo4d_2024}           & Multiple                & Ego and Exo  & RGB, audio, 3D, gaze, IMU, camera poses                  & 1286 hours  & 5035             & 740              \\
  EgoExoLearn\textcolor{red}{(2024)} \cite{huang_egoexolearn_2024}      & Daily and lab activity  & Ego and Exo  & RGB, gaze                                                & 120 hours   & 747              & Several          \\
  EgoCVR\textcolor{red}{(2024)} \cite{hummel_egocvr_2024}                 & Multiple                & Ego          & RGB                                                      & 6 hours     & 2754             & Several          \\
  EgoExo-Fitness\textcolor{red}{(2024)} \cite{li_egoexofitness_2024}    & Fitness                 & Ego and Exo  & RGB                                                      & 32 hours    & 1276             & Several          \\
  TACO\textcolor{red}{(2024)} \cite{liu_taco_2024}                        & Activity involving tool & Ego and Exo  & RGB, depth                                               & 5.2M frames & 2500             & 14               \\
  AEA\textcolor{red}{(2024)} \cite{lv_aria_2024}                          & Daily Life              & Ego          & RGB, audio, 3D, gaze, IMU, camera poses, several sensors & 7.3 hours   & 143              & Several          \\
  Nymeria\textcolor{red}{(2024)} \cite{ma_nymeria_2024}                   & Daily Life              & Ego and Exo  & RGB, 3D, gaze, IMU, XSens mocap suit, several sensors    & 300 hours   & 1200             & 264              \\
  IndustReal\textcolor{red}{(2024)}  \cite{schoonbeek_industreal_2024}   & Toy Assembly            & Ego          & RGB, stereo views, depth, gaze, camera poses             & 5.8 hours   & 84               & 27               \\
  EgoTracks\textcolor{red}{(2024)} \cite{tang_egotracks_2024}             & Multiple                & Ego          & RGB                                                      & 603 hours   & 5708             & Several          \\
  EgoBody3M\textcolor{red}{(2024)} \cite{zhao_egobody3m_2024}             & Indoor                  & Ego and Exo  & RGB, depth                                               & 30 hours    & 2688             & 120              \\
  EgoMe\textcolor{red}{(2025)} \cite{qiu_egome_2025}                      & Multiple                & Ego and Exo  & RGB, depth, gaze, several sensors                        & 82 hours    & 15804            & 37              \\
  \botrule
  \end{tabularx}
  \end{minipage}
  \end{center}
\end{sidewaystable}

\begin{sidewaystable}
  \begin{center}
  \begin{minipage}{\textwidth}
  \caption{Tasks suitable for each dataset. Gaze refers to sect. \ref{Gaze Understanding}. Pose refers to sect. \ref{Pose Estimation}. Action refers to sect. \ref{Action Understanding}. Social refers to sect. \ref{Social Perception}. Object refers to sect. \ref{Instance Object Recognition}. Environment refers to sect. \ref{Environment Understanding Related Tasks}. Summary refers to sect. \ref{Content Summary}. Multi-View refers to sect. \ref{Multiview Joint Understanding}. VQA refers to sect. \ref{Video Question Answering}}\label{table3}
  \tiny
  \begin{tabular*}{\textwidth}{@{\extracolsep{\fill}}lccccccccc@{\extracolsep{\fill}}}
  \toprule%
  & \multicolumn{9}{@{}c@{}}{Task}  \\
  \cmidrule{2-10}
  Dataset & Gaze & Pose & Action & Social & Object & Environment & Summary & Multi-View & VQA \\
  \midrule
  ADL\textcolor{red}{(2012)}  \cite{pirsiavash_detecting_2012} &  &  & \Checkmark &  & \Checkmark &  & \Checkmark &  &  \\
  EGTEA Gaze+\textcolor{red}{(2021)} \cite{li_eye_2021} & \Checkmark &  & \Checkmark &  &  &  &  &  &  \\
  EPIC-KITCHENS-100\textcolor{red}{(2022)}   \cite{damen_rescaling_2022} &  &  & \Checkmark &  & \Checkmark &  & \Checkmark &  &  \\
  Ego4D\textcolor{red}{(2022)} \cite{grauman_ego4d_2022} & \Checkmark &  & \Checkmark & \Checkmark & \Checkmark &  & \Checkmark & \Checkmark &  \\
  HOI4D\textcolor{red}{(2022)} \cite{liu_hoi4d_2022} &  & \Checkmark & \Checkmark &  & \Checkmark &  &  &  &  \\
  AssemblyHands\textcolor{red}{(2023)}   \cite{ohkawa_assemblyhands_2023} &  & \Checkmark & \Checkmark &  &  &  &  & \Checkmark &  \\
  UnrealEgo2\textcolor{red}{(2024)} \cite{akada_3d_2024} &  & \Checkmark & \Checkmark &  &  &  &  &  &  \\
  EgoPet\textcolor{red}{(2024)} \cite{bar_egopet_2024} &  &  & \Checkmark &  &  & \Checkmark &  &  &  \\
  WEAR\textcolor{red}{(2024)} \cite{bock_wear_2024} &  &  & \Checkmark &  &  &  &  &  &  \\
  EgoThink\textcolor{red}{(2024)} \cite{cheng_egothink_2024} &  &  & \Checkmark &  & \Checkmark &  &  &  & \Checkmark \\
  Ego-Exo4D\textcolor{red}{(2024)} \cite{grauman_egoexo4d_2024} & \Checkmark & \Checkmark & \Checkmark &  & \Checkmark &  & \Checkmark & \Checkmark &  \\
  EgoExoLearn\textcolor{red}{(2024)}   \cite{huang_egoexolearn_2024} & \Checkmark &  & \Checkmark &  &  &  & \Checkmark & \Checkmark &  \\
  EgoCVR\textcolor{red}{(2024)} \cite{hummel_egocvr_2024} &  &  & \Checkmark &  &  &  & \Checkmark &  & \Checkmark \\
  EgoExo-Fitness\textcolor{red}{(2024)}   \cite{li_egoexofitness_2024} &  &  & \Checkmark &  &  &  &  & \Checkmark &  \\
  TACO\textcolor{red}{(2024)} \cite{liu_taco_2024} &  & \Checkmark & \Checkmark &  & \Checkmark &  &  & \Checkmark &  \\
  AEA\textcolor{red}{(2024)} \cite{lv_aria_2024} & \Checkmark &  & \Checkmark &  & \Checkmark & \Checkmark &  &  &  \\
  Nymeria\textcolor{red}{(2024)} \cite{ma_nymeria_2024} & \Checkmark & \Checkmark & \Checkmark &  &  & \Checkmark & \Checkmark & \Checkmark &  \\
  IndustReal\textcolor{red}{(2024)}   \cite{schoonbeek_industreal_2024} & \Checkmark &  & \Checkmark &  & \Checkmark &  &  &  &  \\
  EgoTracks\textcolor{red}{(2024)} \cite{tang_egotracks_2024} &  &  &  &  & \Checkmark &  &  &  &  \\
  EgoBody3M\textcolor{red}{(2024)} \cite{zhao_egobody3m_2024} &  & \Checkmark &  &  &  &  &  &  &  \\
  EgoMe\textcolor{red}{(2025)} \cite{qiu_egome_2025} & \Checkmark &  & \Checkmark &  &  &  & \Checkmark & \Checkmark &  \\
  \botrule
  \end{tabular*}
  \end{minipage}
  \end{center}
\end{sidewaystable}

\begin{table}[ht]
  \begin{center}
  \begin{minipage}{\textwidth}
  \tiny
  \caption{Available annotations and websites for datasets}\label{table2}
  \begin{tabularx}{\textwidth}{lXc}
  \toprule%
  Dataset & Annotations & Website \\
  \midrule
  ADL\textcolor{red}{(2012)}   \cite{pirsiavash_detecting_2012} & Temporal action segments, object bounding   boxes, active interacting object & \href{Link}{https://web.cs.ucdavis.edu/~hpirsiav/papers/ADLdataset/} \\
  EGTEA Gaze+\textcolor{red}{(2021)} \cite{li_eye_2021} & Temporal action segments, verb-noun action, hand mask & \href{Link}{https://cbs.ic.gatech.edu/fpv/} \\
  EPIC-KITCHENS-100\textcolor{red}{(2022)}   \cite{damen_rescaling_2022} & Temporal action segments, verb-noun action, narrations, hand and objects   masks, active interacting object & \href{Link}{https://epic-kitchens.github.io/2025} \\
  Ego4D\textcolor{red}{(2022)} \cite{grauman_ego4d_2022} & Temporal action segments,   verb-noun action, narrations, moment queries, speaker labels, diarisation,   hand bounding boxes, time to contact, active objects bounding boxes,   trajectories, next-active objects bounding boxes & \href{Link}{https://ego4d-data.org/} \\
  HOI4D\textcolor{red}{(2022)} \cite{liu_hoi4d_2022} & Temporal action segments,   verb-noun action, 3D hand poses and object poses, panoptic and motion   segmentation, object meshes, point clouds & \href{Link}{https://hoi4d.github.io/} \\
  AssemblyHands\textcolor{red}{(2023)}   \cite{ohkawa_assemblyhands_2023} & Temporal action segments, verb-noun action, 3D hand poses & \href{Link}{https://assemblyhands.github.io/} \\
  UnrealEgo2\textcolor{red}{(2024)} \cite{akada_3d_2024} & Action, 3D body and hand poses & \href{Link}{https://4dqv.mpi-inf.mpg.de/UnrealEgo2/} \\
  EgoPet\textcolor{red}{(2024)} \cite{bar_egopet_2024} & Temporal action segments, pseudo ground truth agent trajectories,   parameters of the terrain & \href{Link}{https://www.amirbar.net/egopet/} \\
  WEAR\textcolor{red}{(2024)} \cite{bock_wear_2024} & Temporal action segments & \href{Link}{https://mariusbock.github.io/wear/} \\
  EgoThink\textcolor{red}{(2024)} \cite{cheng_egothink_2024} & Question-answer pairs of action, object, localization, reasonning & \href{Link}{https://adacheng.github.io/EgoThink/} \\
  Ego-Exo4D\textcolor{red}{(2024)} \cite{grauman_egoexo4d_2024} & Temporal action segments,   verb-noun action, expert commentary of action, narrate-and-act descriptions   of action, atomic action descriptions, 3D hand poses and body poses, Ego-Exo   object mask & \href{Link}{https://ego-exo4d-data.org/} \\
  EgoExoLearn\textcolor{red}{(2024)}   \cite{huang_egoexolearn_2024} & Temporal action segments, coarse-level actions, fine-level actions & \href{Link}{https://github.com/OpenGVLab/EgoExoLearn} \\
  EgoCVR\textcolor{red}{(2024)} \cite{hummel_egocvr_2024} & Action narrations, composed video retrieval queries & \href{Link}{https://github.com/ExplainableML/EgoCVR} \\
  EgoExo-Fitness\textcolor{red}{(2024)}   \cite{li_egoexofitness_2024} & Two-level temporal action segments, technical keypoint verification,   natural language comments, action quality scores & \href{Link}{https://github.com/iSEE-Laboratory/EgoExo-Fitness} \\
  TACO\textcolor{red}{(2024)} \cite{liu_taco_2024} & Tool-action-object,  3D hand poses   and object poses, hand and objects masks & \href{Link}{https://taco2024.github.io/} \\
  AEA\textcolor{red}{(2024)} \cite{lv_aria_2024} & Temporal action segments,  3D   location, scene point cloud, speech narrations & \href{Link}{https://www.projectaria.com/datasets/aea/} \\
  Nymeria\textcolor{red}{(2024)} \cite{ma_nymeria_2024} & Temporal action segment, 3D hand poses, 3D location, scene point cloud,   motion narration, atomic action, activity summarization & \href{Link}{https://www.projectaria.com/datasets/nymeria/} \\
  IndustReal\textcolor{red}{(2024)}   \cite{schoonbeek_industreal_2024} & Temporal action segment, verb-noun action, hand key points, assembly   state, procedure step & \href{Link}{https://github.com/TimSchoonbeek/IndustReal} \\
  EgoTracks\textcolor{red}{(2024)} \cite{tang_egotracks_2024} & Objects bounding boxes, interaction state, transformed state,   recognizable state & \href{Link}{https://ego4d-data.org/docs/data/egotracks/} \\
  EgoBody3M\textcolor{red}{(2024)} \cite{zhao_egobody3m_2024} & Action, 3D kinematic body model, 2D body keypoint & \href{Link}{https://github.com/facebookresearch/EgoBody3M} \\
  EgoMe\textcolor{red}{(2025)} \cite{qiu_egome_2025} & Coarse-level video language annotation, fine-grained descriptions and   temporal segments & \href{Link}{https://huggingface.co/datasets/HeqianQiu/EgoMe} \\
  \botrule
  \end{tabularx}
  \end{minipage}
  \end{center}
\end{table}

\section{Trends and Future}\label{Trends and Future}
Building upon the challenges analyzed in sect. \ref{Challenges}, this section summarizes recent trends in egocentric vision research. These trends reflect how the community is actively developing new strategies to overcome the key difficulties inherent to egocentric data understanding and the technical bottlenecks faced by current methods.

\begin{figure}[ht]%
  \centering
  \includegraphics[width=0.8\textwidth]{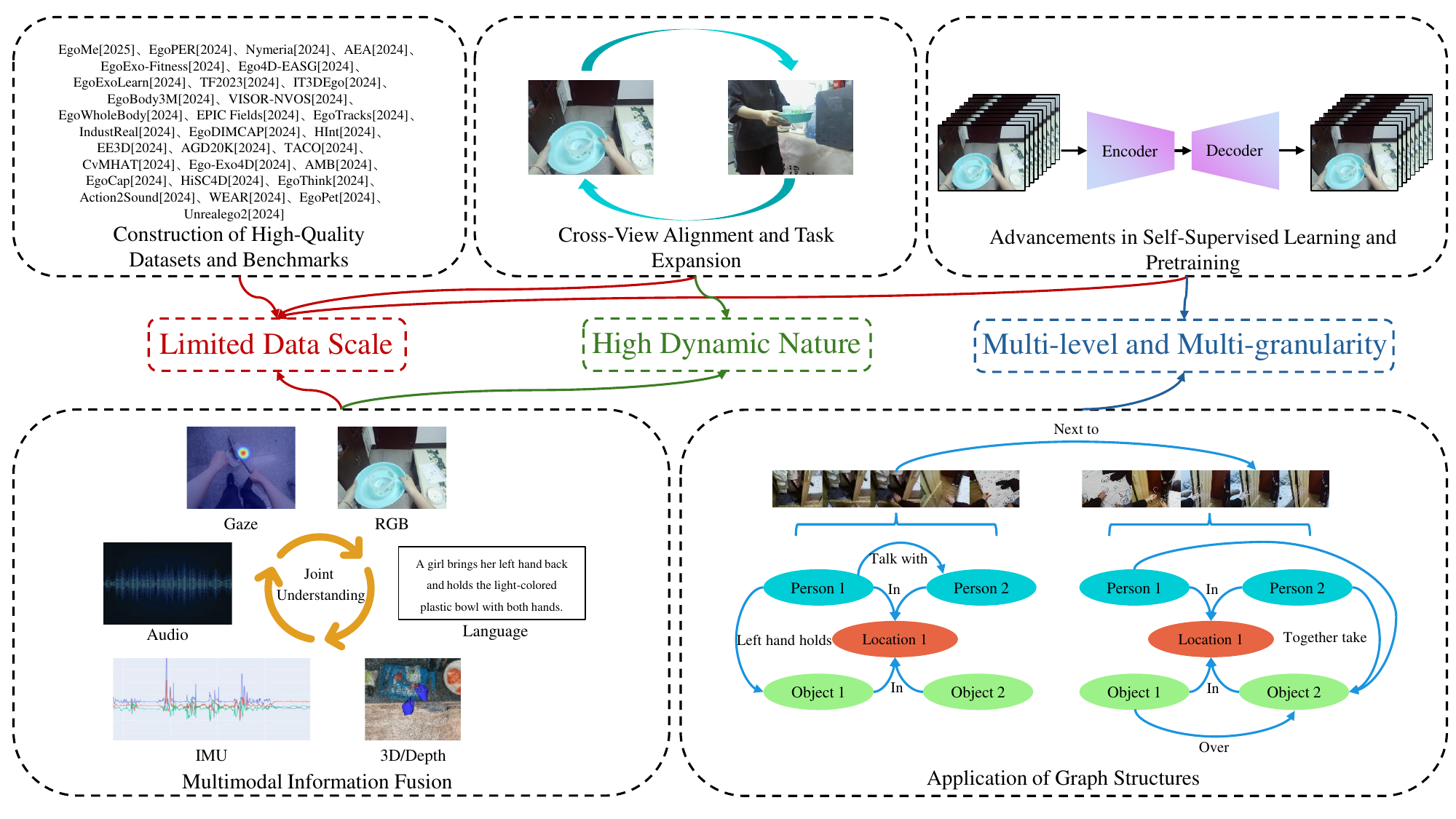}
  \caption{Five major trends emerges in recent work, connected with the challenges currently faced. A batch of high-quality datasets was proposed in 2024. Cross-view alignment and cross-view tasks have been explored more extensively. Pre-trained models with longer input frames (such as MAE and contrastive learning) are also becoming a trend. Multimodal information (including audio, gaze obtained from eye trackers, textual descriptions, body-worn IMUs, 3D point clouds, or depth information) is being utilized more thoroughly. The relationships among actions, objects, and environments can be structured using graph structures, which aids in better understanding.}\label{trend}
\end{figure}

In recent work within the egocentric vision domain, five major trends appear shown in Fig. \ref{trend}. The first is a substantial number of new datasets and benchmarks have been proposed, providing higher-quality and larger-scale data for various downstream tasks. These efforts include EgoMe \cite{qiu_egome_2025}, EgoPER \cite{shih-polee_error_2024}, Nymeria \cite{ma_nymeria_2024}, AEA \cite{lv_aria_2024}, EgoExo-Fitness \cite{li_egoexofitness_2024}, Ego4D-EASG \cite{ivanrodin_action_2024}, EgoExoLearn \cite{huang_egoexolearn_2024}, TF2023 \cite{zhao_fusing_2024}, IT3DEgo \cite{zhao_instance_2024}, EgoBody3M \cite{zhao_egobody3m_2024}, VISOR-NVOS \cite{yuhanshen_learning_2024}, EgoWholeBody \cite{wang_egocentric_2024}, EPIC Fields \cite{tschernezki_epic_2024}, EgoTracks \cite{tang_egotracks_2024}, IndustReal \cite{schoonbeek_industreal_2024}, EgoDIMCAP \cite{qiu_egocentric_2024}, HInt \cite{pavlakos_reconstructing_2024}, EE3D \cite{millerdurai_eventego3d_2024}, AGD20K \cite{luo_grounded_2024}, TACO \cite{liu_taco_2024}, CvMHAT \cite{han_benchmarking_2024}, Ego-Exo4D \cite{grauman_egoexo4d_2024}, AMB \cite{goletto_amego_2024}, EgoCap \cite{dai_egocap_2024}, HiSC4D \cite{dai_hisc4d_2024}, EgoThink \cite{cheng_egothink_2024}, Action2Sound \cite{chen_action2sound_2024}, WEAR \cite{bock_wear_2024}, EgoPet \cite{bar_egopet_2024}, and Unrealego2 \cite{akada_3d_2024}. These datasets effectively mitigate the data scarcity issue\refcircstep[BurntOrange]{challenge:datasets} in the egocentric vision domain. Among them, we have introduced a subset of these datasets in the sect. \ref{Datasets}.

Furthermore, research related to multi-view and cross-view understanding is gradually becoming a focal point in the egocentric vision domain. The specifics of these works will be elaborated in sect. \ref{Multiview Joint Understanding}. Notably, in the Exo-Ego direction, researchers have primarily focused on aligning Exo-Ego perspectives and learning viewpoint-invariant features. This research area has made significant progress in recent studies \cite{jang_intra_2024}, Xu et al. \cite{xu_weakly_2024}, Truong et al. \cite{truong_crossview_2025}, Quattrocchi et al. \cite{quattrocchi_synchronization_2024}, Tran et al. \cite{tran_ex2egmae_2024}. Through cross-view alignment, researchers can effectively leverage prior knowledge learned from large-scale conventional view data to enhance the performance of egocentric models and mitigate the issue of limited egocentric data\refcircstep[BurntOrange]{challenge:datasets}. Additionally, the introduction of multi-view alignment and joint understanding strategies helps models overcome the challenges of high dynamics\refcircstep[BurntOrange]{challenge:highly dynamic} in egocentric scenes by integrating stable and unobstructed visual information from conventional views, thereby significantly improving the performance of egocentric tasks. Although preliminary results have been achieved in cross-view alignment research, the field is still in its infancy, and many innovative tasks remain to be explored. Recently, research based on cross-view datasets \cite{grauman_egoexo4d_2024}, Huang et al. \cite{huang_egoexolearn_2024}, Li et al. \cite{li_egoexofitness_2024}, Qiu et al. \cite{qiu_egome_2025} has introduced a series of new cross-view task benchmarks, which are expected to become key research hotspots in the egocentric vision field.

In the research achieving SOTA performance in the egocentric vision domain, a notable trend is the increasing focus on combining pretraining and self-supervised learning methods, which have demonstrated significant advantages. Zhang et al. \cite{zhang_masked_2024} proposed a self-supervised pretraining method based on the multimodal Masked Autoencoder (MAE) framework \cite{he_masked_2022}, achieving strong multimodal representation capabilities in egocentric action understanding tasks. Tran et al. \cite{tran_ex2egmae_2024} utilized synthetic Exo-Ego data to design a self-supervised MAE architecture, achieving SOTA performance in social scene understanding tasks. Furthermore, Liu et al. \cite{liu_endtoend_2024} and \cite{gundavarapu_extending_2024} extended the maximum input length of video MAE by improving network design, achieving SOTA performance in temporal action segmentation and action recognition tasks, respectively. These findings highlight the critical role of self-supervised pretraining or fine-tuning in egocentric vision research. At the same time, there remains vast exploration potential in self-supervised tasks for egocentric or cross-view vision. Self-supervised learning methods effectively alleviate the issue of limited labeled data\refcircstep[BurntOrange]{challenge:datasets} in egocentric vision, while longer input lengths help the model better understand multi-granularity temporal information\refcircstep[BurntOrange]{challenge:multi-granularity} and contextual relationships, further enhancing performance. In the future, in-depth research combining self-supervised learning and pretraining techniques is expected to lead to more groundbreaking advancements in the egocentric vision field.

To overcome the high dynamic challenges faced by egocentric vision\refcircstep[BurntOrange]{challenge:highly dynamic}, the current trend is to leverage multimodal information to assist understanding. In terms of utilizing audio information, Yun et al. \cite{yun_spherical_2024} effectively used sound to compensate for the effects of self-motion, thereby improving the model's performance in gaze prediction and social understanding. Majumder et al. \cite{majumder_learning_2024} jointly reconstructed masked binaural audio with audio-video information, achieving more accurate perception of the active speaker in social scenes. Lai et al. \cite{lai_listen_2024} also employed audio information to assist in gaze prediction. Both \cite{chen_soundingactions_2024} and \cite{chen_action2sound_2024} explored the role of audio information in action understanding tasks. In action recognition tasks, Tan et al. \cite{tan_egodistill_2023}, Gong et al. \cite{gong_mmgego4d_2023}, and \cite{zhang_masked_2024} attempted to integrate IMU data into the network to mitigate the negative effects of occlusion and motion blur. These studies demonstrate the effectiveness of multimodal information in egocentric vision tasks, and future work may further address the limitations and high dynamics of wearable camera perspectives by incorporating more types of multimodal information.

In understanding multi-level and multi-granularity information in egocentric vision\refcircstep[BurntOrange]{challenge:multi-granularity}, many recent studies have attempted to incorporate graph data structures into models or to represent data through graph structures to achieve structured and hierarchical processing of egocentric content. Zhang et al. \cite{zhang_masked_2024} embedded the relative motion features of human joints into a graph structure, thereby enabling the effective fusion of multiple IMU data. Jia et al. \cite{wenqijia_audiovisual_2024} utilized graph representations to model the interactions among multiple individuals in social scenes. Rodin et al. \cite{ivanrodin_action_2024} employed graph structures to capture actions, interacting objects, their relationships, and the temporal progression of actions in egocentric videos. In egocentric vision, the focus is often on the subject's actions or the interactions between the subject and objects or the environment. Actions, particularly procedural activities, typically have temporal contextual relationships, and graph structures are inherently well-suited to represent relationships between different elements. As a result, graph data and graph networks may become a significant direction in egocentric vision research.

\section{Conclusion}\label{Conclusion}
In this paper, we conduct a comprehensive survey of the egocentric vision field. We provide a detailed summary of the 11 sub-tasks covered under the four primary aspects of the field: subject, object, environment, and hybrid. These sub-tasks include gaze understanding, pose estimation, action understanding, social perception, human identity and trajectory recognition, object recognition, environment modeling, scene localization, content summarization, multi-view joint understanding, and video question answering. For each sub-task, we first provide a brief introduction, followed by an in-depth analysis of the challenges it faces. We then summarize the key research works and methods, especially those that have achieved the current state-of-the-art (SOTA) performance. Additionally, we list the available datasets for each sub-task.

In reviewing the related works in the egocentric vision field, we particularly focus on the latest research advancements and, based on these works, summarize the three main challenges currently faced by the field. In response to these challenges, we distill five key development trends from recent significant research literature. We hope that the summary of these challenges and trends will provide direction and insight for future research in the egocentric vision field. Additionally, we systematically analyze the components of egocentric scenes, offering a clearer portrayal of the characteristics of egocentric vision and clarifying the research targets. Regarding egocentric datasets, we provide a detailed introduction to 21 different datasets, listing the basic features and annotations for each dataset. The tasks suitable for each dataset are also noted. With the continuous advancement of technology and the ongoing deepening of research, we believe that the egocentric vision field will undergo more comprehensive and in-depth exploration in the future. 

\bibliography{reference}


\begin{thebibliography}{238}
\ifx \bisbn   \undefined \def \bisbn  #1{ISBN #1}\fi
\ifx \binits  \undefined \def \binits#1{#1}\fi
\ifx \bauthor  \undefined \def \bauthor#1{#1}\fi
\ifx \batitle  \undefined \def \batitle#1{#1}\fi
\ifx \bjtitle  \undefined \def \bjtitle#1{#1}\fi
\ifx \bvolume  \undefined \def \bvolume#1{\textbf{#1}}\fi
\ifx \byear  \undefined \def \byear#1{#1}\fi
\ifx \bissue  \undefined \def \bissue#1{#1}\fi
\ifx \bfpage  \undefined \def \bfpage#1{#1}\fi
\ifx \blpage  \undefined \def \blpage #1{#1}\fi
\ifx \burl  \undefined \def \burl#1{\textsf{#1}}\fi
\ifx \doiurl  \undefined \def \doiurl#1{\url{https://doi.org/#1}}\fi
\ifx \betal  \undefined \def \betal{\textit{et al.}}\fi
\ifx \binstitute  \undefined \def \binstitute#1{#1}\fi
\ifx \binstitutionaled  \undefined \def \binstitutionaled#1{#1}\fi
\ifx \bctitle  \undefined \def \bctitle#1{#1}\fi
\ifx \beditor  \undefined \def \beditor#1{#1}\fi
\ifx \bpublisher  \undefined \def \bpublisher#1{#1}\fi
\ifx \bbtitle  \undefined \def \bbtitle#1{#1}\fi
\ifx \bedition  \undefined \def \bedition#1{#1}\fi
\ifx \bseriesno  \undefined \def \bseriesno#1{#1}\fi
\ifx \blocation  \undefined \def \blocation#1{#1}\fi
\ifx \bsertitle  \undefined \def \bsertitle#1{#1}\fi
\ifx \bsnm \undefined \def \bsnm#1{#1}\fi
\ifx \bsuffix \undefined \def \bsuffix#1{#1}\fi
\ifx \bparticle \undefined \def \bparticle#1{#1}\fi
\ifx \barticle \undefined \def \barticle#1{#1}\fi
\bibcommenthead
\ifx \bconfdate \undefined \def \bconfdate #1{#1}\fi
\ifx \botherref \undefined \def \botherref #1{#1}\fi
\ifx \url \undefined \def \url#1{\textsf{#1}}\fi
\ifx \bchapter \undefined \def \bchapter#1{#1}\fi
\ifx \bbook \undefined \def \bbook#1{#1}\fi
\ifx \bcomment \undefined \def \bcomment#1{#1}\fi
\ifx \oauthor \undefined \def \oauthor#1{#1}\fi
\ifx \citeauthoryear \undefined \def \citeauthoryear#1{#1}\fi
\ifx \endbibitem  \undefined \def \endbibitem {}\fi
\ifx \bconflocation  \undefined \def \bconflocation#1{#1}\fi
\ifx \arxivurl  \undefined \def \arxivurl#1{\textsf{#1}}\fi
\csname PreBibitemsHook\endcsname

\bibitem{betancourt_evolution_2015}
\begin{barticle}
\bauthor{\bsnm{Betancourt}, \binits{A.}},
\bauthor{\bsnm{Morerio}, \binits{P.}},
\bauthor{\bsnm{Regazzoni}, \binits{C.S.}},
\bauthor{\bsnm{Rauterberg}, \binits{M.}}:
\batitle{The evolution of first person vision methods: A survey}.
\bjtitle{IEEE Transactions on Circuits and Systems for Video Technology}
\bvolume{25}(\bissue{5}),
\bfpage{744}--\blpage{760}
(\byear{2015})
\end{barticle}
\endbibitem

\bibitem{molino_summarization_2016}
\begin{barticle}
\bauthor{\bsnm{Del~Molino}, \binits{A.G.}},
\bauthor{\bsnm{Tan}, \binits{C.}},
\bauthor{\bsnm{Lim}, \binits{J.-H.}},
\bauthor{\bsnm{Tan}, \binits{A.-H.}}:
\batitle{Summarization of egocentric videos: A comprehensive survey}.
\bjtitle{IEEE Transactions on Human-Machine Systems}
\bvolume{47}(\bissue{1}),
\bfpage{65}--\blpage{76}
(\byear{2016})
\end{barticle}
\endbibitem

\bibitem{rodin_predicting_2021}
\begin{barticle}
\bauthor{\bsnm{Rodin}, \binits{I.}},
\bauthor{\bsnm{Furnari}, \binits{A.}},
\bauthor{\bsnm{Mavroeidis}, \binits{D.}},
\bauthor{\bsnm{Farinella}, \binits{G.M.}}:
\batitle{Predicting the future from first person (egocentric) vision: A survey}.
\bjtitle{Computer Vision and Image Understanding}
\bvolume{211},
\bfpage{103252}
(\byear{2021})
\end{barticle}
\endbibitem

\bibitem{nunez-marcos_egocentric_2022}
\begin{barticle}
\bauthor{\bsnm{N{\'u}{\~n}ez-Marcos}, \binits{A.}},
\bauthor{\bsnm{Azkune}, \binits{G.}},
\bauthor{\bsnm{Arganda-Carreras}, \binits{I.}}:
\batitle{Egocentric vision-based action recognition: A survey}.
\bjtitle{Neurocomputing}
\bvolume{472},
\bfpage{175}--\blpage{197}
(\byear{2022})
\end{barticle}
\endbibitem

\bibitem{bandini_analysis_2023}
\begin{barticle}
\bauthor{\bsnm{Bandini}, \binits{A.}},
\bauthor{\bsnm{Zariffa}, \binits{J.}}:
\batitle{Analysis of the hands in egocentric vision: A survey}.
\bjtitle{IEEE transactions on pattern analysis and machine intelligence}
\bvolume{45}(\bissue{6}),
\bfpage{6846}--\blpage{6866}
(\byear{2020})
\end{barticle}
\endbibitem

\bibitem{azam_survey_2024}
\begin{bchapter}
\bauthor{\bsnm{Azam}, \binits{M.M.}},
\bauthor{\bsnm{Desai}, \binits{K.}}:
\bctitle{A survey on 3d egocentric human pose estimation}.
In: \bbtitle{Proceedings of the IEEE/CVF Conference on Computer Vision and Pattern Recognition},
pp. \bfpage{1643}--\blpage{1654}
(\byear{2024})
\end{bchapter}
\endbibitem

\bibitem{fan_benchmarks_2024}
\begin{bchapter}
\bauthor{\bsnm{Fan}, \binits{Z.}},
\bauthor{\bsnm{Ohkawa}, \binits{T.}},
\bauthor{\bsnm{Yang}, \binits{L.}},
\bauthor{\bsnm{Lin}, \binits{N.}},
\bauthor{\bsnm{Zhou}, \binits{Z.}},
\bauthor{\bsnm{Zhou}, \binits{S.}},
\bauthor{\bsnm{Liang}, \binits{J.}},
\bauthor{\bsnm{Gao}, \binits{Z.}},
\bauthor{\bsnm{Zhang}, \binits{X.}},
\bauthor{\bsnm{Zhang}, \binits{X.}}, \betal:
\bctitle{Benchmarks and challenges in pose estimation for egocentric hand interactions with objects}.
In: \bbtitle{European Conference on Computer Vision},
pp. \bfpage{428}--\blpage{448}
(\byear{2024}).
\bcomment{Springer}
\end{bchapter}
\endbibitem

\bibitem{plizzari_outlook_2024}
\begin{botherref}
\oauthor{\bsnm{Plizzari}, \binits{C.}},
\oauthor{\bsnm{Goletto}, \binits{G.}},
\oauthor{\bsnm{Furnari}, \binits{A.}},
\oauthor{\bsnm{Bansal}, \binits{S.}},
\oauthor{\bsnm{Ragusa}, \binits{F.}},
\oauthor{\bsnm{Farinella}, \binits{G.M.}},
\oauthor{\bsnm{Damen}, \binits{D.}},
\oauthor{\bsnm{Tommasi}, \binits{T.}}:
An outlook into the future of egocentric vision.
International Journal of Computer Vision,
1--57
(2024)
\end{botherref}
\endbibitem

\bibitem{wang2025scaling}
\begin{botherref}
\oauthor{\bsnm{Wang}, \binits{X.}},
\oauthor{\bsnm{Alabdulmohsin}, \binits{I.}},
\oauthor{\bsnm{Salz}, \binits{D.}},
\oauthor{\bsnm{Li}, \binits{Z.}},
\oauthor{\bsnm{Rong}, \binits{K.}},
\oauthor{\bsnm{Zhai}, \binits{X.}}:
Scaling pre-training to one hundred billion data for vision language models.
arXiv preprint arXiv:2502.07617
(2025)
\end{botherref}
\endbibitem

\bibitem{wang2023swap}
\begin{botherref}
\oauthor{\bsnm{Wang}, \binits{W.}},
\oauthor{\bsnm{Yang}, \binits{H.}},
\oauthor{\bsnm{Tuo}, \binits{Z.}},
\oauthor{\bsnm{He}, \binits{H.}},
\oauthor{\bsnm{Zhu}, \binits{J.}},
\oauthor{\bsnm{Fu}, \binits{J.}},
\oauthor{\bsnm{Liu}, \binits{J.}}:
Swap attention in spatiotemporal diffusions for text-to-video generation.
arXiv preprint arXiv:2305.10874
(2023)
\end{botherref}
\endbibitem

\bibitem{grauman_ego4d_2022}
\begin{bchapter}
\bauthor{\bsnm{Grauman}, \binits{K.}},
\bauthor{\bsnm{Westbury}, \binits{A.}},
\bauthor{\bsnm{Byrne}, \binits{E.}},
\bauthor{\bsnm{Chavis}, \binits{Z.}},
\bauthor{\bsnm{Furnari}, \binits{A.}},
\bauthor{\bsnm{Girdhar}, \binits{R.}},
\bauthor{\bsnm{Hamburger}, \binits{J.}},
\bauthor{\bsnm{Jiang}, \binits{H.}},
\bauthor{\bsnm{Liu}, \binits{M.}},
\bauthor{\bsnm{Liu}, \binits{X.}}, \betal:
\bctitle{Ego4d: Around the world in 3,000 hours of egocentric video}.
In: \bbtitle{Proceedings of the IEEE/CVF Conference on Computer Vision and Pattern Recognition},
pp. \bfpage{18995}--\blpage{19012}
(\byear{2022})
\end{bchapter}
\endbibitem

\bibitem{yun_spherical_2024}
\begin{bchapter}
\bauthor{\bsnm{Yun}, \binits{H.}},
\bauthor{\bsnm{Gao}, \binits{R.}},
\bauthor{\bsnm{Ananthabhotla}, \binits{I.}},
\bauthor{\bsnm{Kumar}, \binits{A.}},
\bauthor{\bsnm{Donley}, \binits{J.}},
\bauthor{\bsnm{Li}, \binits{C.}},
\bauthor{\bsnm{Kim}, \binits{G.}},
\bauthor{\bsnm{Ithapu}, \binits{V.K.}},
\bauthor{\bsnm{Murdock}, \binits{C.}}:
\bctitle{Spherical world-locking for audio-visual localization in egocentric videos}.
In: \bbtitle{European Conference on Computer Vision},
pp. \bfpage{256}--\blpage{274}
(\byear{2024}).
\bcomment{Springer}
\end{bchapter}
\endbibitem

\bibitem{tschernezki_epic_2024}
\begin{botherref}
\oauthor{\bsnm{Tschernezki}, \binits{V.}},
\oauthor{\bsnm{Darkhalil}, \binits{A.}},
\oauthor{\bsnm{Zhu}, \binits{Z.}},
\oauthor{\bsnm{Fouhey}, \binits{D.}},
\oauthor{\bsnm{Laina}, \binits{I.}},
\oauthor{\bsnm{Larlus}, \binits{D.}},
\oauthor{\bsnm{Damen}, \binits{D.}},
\oauthor{\bsnm{Vedaldi}, \binits{A.}}:
Epic fields: Marrying 3d geometry and video understanding.
Advances in Neural Information Processing Systems
\textbf{36}
(2024)
\end{botherref}
\endbibitem

\bibitem{zhang_masked_2024}
\begin{bchapter}
\bauthor{\bsnm{Zhang}, \binits{M.}},
\bauthor{\bsnm{Huang}, \binits{Y.}},
\bauthor{\bsnm{Liu}, \binits{R.}},
\bauthor{\bsnm{Sato}, \binits{Y.}}:
\bctitle{Masked video and body-worn imu autoencoder for egocentric action recognition}.
In: \bbtitle{European Conference on Computer Vision},
pp. \bfpage{312}--\blpage{330}
(\byear{2024}).
\bcomment{Springer}
\end{bchapter}
\endbibitem

\bibitem{xu_weakly_2024}
\begin{bchapter}
\bauthor{\bsnm{Xu}, \binits{L.}},
\bauthor{\bsnm{Gao}, \binits{Y.}},
\bauthor{\bsnm{Song}, \binits{W.}},
\bauthor{\bsnm{Hao}, \binits{A.}}:
\bctitle{Weakly supervised multimodal affordance grounding for egocentric images}.
In: \bbtitle{Proceedings of the AAAI Conference on Artificial Intelligence},
vol. \bseriesno{38},
pp. \bfpage{6324}--\blpage{6332}
(\byear{2024})
\end{bchapter}
\endbibitem

\bibitem{zhao_fusing_2024}
\begin{bchapter}
\bauthor{\bsnm{Zhao}, \binits{Z.}},
\bauthor{\bsnm{Wang}, \binits{Y.}},
\bauthor{\bsnm{Wang}, \binits{C.}}:
\bctitle{Fusing personal and environmental cues for identification and segmentation of first-person camera wearers in third-person views}.
In: \bbtitle{Proceedings of the IEEE/CVF Conference on Computer Vision and Pattern Recognition},
pp. \bfpage{16477}--\blpage{16487}
(\byear{2024})
\end{bchapter}
\endbibitem

\bibitem{zhao_instance_2024}
\begin{bchapter}
\bauthor{\bsnm{Zhao}, \binits{Y.}},
\bauthor{\bsnm{Ma}, \binits{H.}},
\bauthor{\bsnm{Kong}, \binits{S.}},
\bauthor{\bsnm{Fowlkes}, \binits{C.}}:
\bctitle{Instance tracking in 3d scenes from egocentric videos}.
In: \bbtitle{Proceedings of the IEEE/CVF Conference on Computer Vision and Pattern Recognition},
pp. \bfpage{21933}--\blpage{21944}
(\byear{2024})
\end{bchapter}
\endbibitem

\bibitem{zhang_refa_2024}
\begin{bchapter}
\bauthor{\bsnm{Zhang}, \binits{Q.}},
\bauthor{\bsnm{Xiao}, \binits{T.}},
\bauthor{\bsnm{Habeeb}, \binits{H.}},
\bauthor{\bsnm{Laich}, \binits{L.}},
\bauthor{\bsnm{Bouaziz}, \binits{S.}},
\bauthor{\bsnm{Snape}, \binits{P.}},
\bauthor{\bsnm{Zhang}, \binits{W.}},
\bauthor{\bsnm{Cioffi}, \binits{M.}},
\bauthor{\bsnm{Zhang}, \binits{P.}},
\bauthor{\bsnm{Pidlypenskyi}, \binits{P.}}, \betal:
\bctitle{Refa: Real-time egocentric facial animations for virtual reality}.
In: \bbtitle{Proceedings of the IEEE/CVF Conference on Computer Vision and Pattern Recognition},
pp. \bfpage{4793}--\blpage{4802}
(\byear{2024})
\end{bchapter}
\endbibitem

\bibitem{yang_egoposeformer_2024}
\begin{bchapter}
\bauthor{\bsnm{Yang}, \binits{C.}},
\bauthor{\bsnm{Tkach}, \binits{A.}},
\bauthor{\bsnm{Hampali}, \binits{S.}},
\bauthor{\bsnm{Zhang}, \binits{L.}},
\bauthor{\bsnm{Crowley}, \binits{E.J.}},
\bauthor{\bsnm{Keskin}, \binits{C.}}:
\bctitle{Egoposeformer: A simple baseline for stereo egocentric 3d human pose estimation}.
(\byear{2024}).
\bcomment{Springer}
\end{bchapter}
\endbibitem

\bibitem{wang_egocentric_2024}
\begin{bchapter}
\bauthor{\bsnm{Wang}, \binits{J.}},
\bauthor{\bsnm{Cao}, \binits{Z.}},
\bauthor{\bsnm{Luvizon}, \binits{D.}},
\bauthor{\bsnm{Liu}, \binits{L.}},
\bauthor{\bsnm{Sarkar}, \binits{K.}},
\bauthor{\bsnm{Tang}, \binits{D.}},
\bauthor{\bsnm{Beeler}, \binits{T.}},
\bauthor{\bsnm{Theobalt}, \binits{C.}}:
\bctitle{Egocentric whole-body motion capture with fisheyevit and diffusion-based motion refinement}.
In: \bbtitle{Proceedings of the IEEE/CVF Conference on Computer Vision and Pattern Recognition},
pp. \bfpage{777}--\blpage{787}
(\byear{2024})
\end{bchapter}
\endbibitem

\bibitem{yuhanshen_learning_2024}
\begin{bchapter}
\bauthor{\bsnm{Shen}, \binits{Y.}},
\bauthor{\bsnm{Wang}, \binits{H.}},
\bauthor{\bsnm{Yang}, \binits{X.}},
\bauthor{\bsnm{Feiszli}, \binits{M.}},
\bauthor{\bsnm{Elhamifar}, \binits{E.}},
\bauthor{\bsnm{Torresani}, \binits{L.}},
\bauthor{\bsnm{Mavroudi}, \binits{E.}}:
\bctitle{Learning to segment referred objects from narrated egocentric videos}.
In: \bbtitle{Proceedings of the IEEE/CVF Conference on Computer Vision and Pattern Recognition},
pp. \bfpage{14510}--\blpage{14520}
(\byear{2024})
\end{bchapter}
\endbibitem

\bibitem{wenqijia_audiovisual_2024}
\begin{bchapter}
\bauthor{\bsnm{Jia}, \binits{W.}},
\bauthor{\bsnm{Liu}, \binits{M.}},
\bauthor{\bsnm{Jiang}, \binits{H.}},
\bauthor{\bsnm{Ananthabhotla}, \binits{I.}},
\bauthor{\bsnm{Rehg}, \binits{J.M.}},
\bauthor{\bsnm{Ithapu}, \binits{V.K.}},
\bauthor{\bsnm{Gao}, \binits{R.}}:
\bctitle{The audio-visual conversational graph: From an egocentric-exocentric perspective}.
In: \bbtitle{Proceedings of the IEEE/CVF Conference on Computer Vision and Pattern Recognition},
pp. \bfpage{26396}--\blpage{26405}
(\byear{2024})
\end{bchapter}
\endbibitem

\bibitem{ivanrodin_action_2024}
\begin{bchapter}
\bauthor{\bsnm{Rodin}, \binits{I.}},
\bauthor{\bsnm{Furnari}, \binits{A.}},
\bauthor{\bsnm{Min}, \binits{K.}},
\bauthor{\bsnm{Tripathi}, \binits{S.}},
\bauthor{\bsnm{Farinella}, \binits{G.M.}}:
\bctitle{Action scene graphs for long-form understanding of egocentric videos}.
In: \bbtitle{Proceedings of the IEEE/CVF Conference on Computer Vision and Pattern Recognition},
pp. \bfpage{18622}--\blpage{18632}
(\byear{2024})
\end{bchapter}
\endbibitem

\bibitem{lai_eye_2022}
\begin{bchapter}
\bauthor{\bsnm{Lai}, \binits{B.}},
\bauthor{\bsnm{Liu}, \binits{M.}},
\bauthor{\bsnm{Ryan}, \binits{F.}},
\bauthor{\bsnm{Rehg}, \binits{J.M.}}:
\bctitle{In the eye of transformer: Global-local correlation for egocentric gaze estimation}.
In: \bbtitle{33rd British Machine Vision Conference 2022, {BMVC} 2022, London, UK, November 21-24, 2022}
(\byear{2022}).
\bcomment{{BMVA} Press}.
\burl{https://bmvc2022.mpi-inf.mpg.de/0227.pdf}
\end{bchapter}
\endbibitem

\bibitem{ouyang_actionvos_2024}
\begin{bchapter}
\bauthor{\bsnm{Ouyang}, \binits{L.}},
\bauthor{\bsnm{Liu}, \binits{R.}},
\bauthor{\bsnm{Huang}, \binits{Y.}},
\bauthor{\bsnm{Furuta}, \binits{R.}},
\bauthor{\bsnm{Sato}, \binits{Y.}}:
\bctitle{Actionvos: Actions as prompts for video object segmentation}.
In: \bbtitle{European Conference on Computer Vision},
pp. \bfpage{216}--\blpage{235}
(\byear{2024}).
\bcomment{Springer}
\end{bchapter}
\endbibitem

\bibitem{mur-labadia_affttention_2024}
\begin{bchapter}
\bauthor{\bsnm{Mur-Labadia}, \binits{L.}},
\bauthor{\bsnm{Martinez-Cantin}, \binits{R.}},
\bauthor{\bsnm{Guerrero}, \binits{J.J.}},
\bauthor{\bsnm{Farinella}, \binits{G.M.}},
\bauthor{\bsnm{Furnari}, \binits{A.}}:
\bctitle{Aff-ttention! affordances and attention models for short-term object interaction anticipation}.
In: \bbtitle{European Conference on Computer Vision},
pp. \bfpage{167}--\blpage{184}
(\byear{2024}).
\bcomment{Springer}
\end{bchapter}
\endbibitem

\bibitem{zhao_antgpt_2024}
\begin{bchapter}
\bauthor{\bsnm{Zhao}, \binits{Q.}},
\bauthor{\bsnm{Wang}, \binits{S.}},
\bauthor{\bsnm{Zhang}, \binits{C.}},
\bauthor{\bsnm{Fu}, \binits{C.}},
\bauthor{\bsnm{Do}, \binits{M.Q.}},
\bauthor{\bsnm{Agarwal}, \binits{N.}},
\bauthor{\bsnm{Lee}, \binits{K.}},
\bauthor{\bsnm{Sun}, \binits{C.}}:
\bctitle{Ant{GPT}: Can large language models help long-term action anticipation from videos?}
In: \bbtitle{The Twelfth International Conference on Learning Representations}
(\byear{2024}).
\burl{https://openreview.net/forum?id=Bb21JPnhhr}
\end{bchapter}
\endbibitem

\bibitem{yuhanshen_progressaware_2024}
\begin{bchapter}
\bauthor{\bsnm{Shen}, \binits{Y.}},
\bauthor{\bsnm{Elhamifar}, \binits{E.}}:
\bctitle{Progress-aware online action segmentation for egocentric procedural task videos}.
In: \bbtitle{Proceedings of the IEEE/CVF Conference on Computer Vision and Pattern Recognition},
pp. \bfpage{18186}--\blpage{18197}
(\byear{2024})
\end{bchapter}
\endbibitem

\bibitem{flaborea_prego_2024}
\begin{bchapter}
\bauthor{\bsnm{Flaborea}, \binits{A.}},
\bauthor{\bparticle{di} \bsnm{Melendugno}, \binits{G.M.D.}},
\bauthor{\bsnm{Plini}, \binits{L.}},
\bauthor{\bsnm{Scofano}, \binits{L.}},
\bauthor{\bsnm{De~Matteis}, \binits{E.}},
\bauthor{\bsnm{Furnari}, \binits{A.}},
\bauthor{\bsnm{Farinella}, \binits{G.M.}},
\bauthor{\bsnm{Galasso}, \binits{F.}}:
\bctitle{Prego: online mistake detection in procedural egocentric videos}.
In: \bbtitle{Proceedings of the IEEE/CVF Conference on Computer Vision and Pattern Recognition},
pp. \bfpage{18483}--\blpage{18492}
(\byear{2024})
\end{bchapter}
\endbibitem

\bibitem{cheng_appearancebased_2024}
\begin{botherref}
\oauthor{\bsnm{Cheng}, \binits{Y.}},
\oauthor{\bsnm{Wang}, \binits{H.}},
\oauthor{\bsnm{Bao}, \binits{Y.}},
\oauthor{\bsnm{Lu}, \binits{F.}}:
Appearance-based gaze estimation with deep learning: A review and benchmark.
IEEE Transactions on Pattern Analysis and Machine Intelligence
(2024)
\end{botherref}
\endbibitem

\bibitem{cazzato_when_2020}
\begin{barticle}
\bauthor{\bsnm{Cazzato}, \binits{D.}},
\bauthor{\bsnm{Leo}, \binits{M.}},
\bauthor{\bsnm{Distante}, \binits{C.}},
\bauthor{\bsnm{Voos}, \binits{H.}}:
\batitle{When i look into your eyes: A survey on computer vision contributions for human gaze estimation and tracking}.
\bjtitle{Sensors}
\bvolume{20}(\bissue{13}),
\bfpage{3739}
(\byear{2020})
\end{barticle}
\endbibitem

\bibitem{lai_listen_2024}
\begin{bchapter}
\bauthor{\bsnm{Lai}, \binits{B.}},
\bauthor{\bsnm{Ryan}, \binits{F.}},
\bauthor{\bsnm{Jia}, \binits{W.}},
\bauthor{\bsnm{Liu}, \binits{M.}},
\bauthor{\bsnm{Rehg}, \binits{J.M.}}:
\bctitle{Listen to look into the future: Audio-visual egocentric gaze anticipation}.
In: \bbtitle{European Conference on Computer Vision},
pp. \bfpage{192}--\blpage{210}
(\byear{2024}).
\bcomment{Springer}
\end{bchapter}
\endbibitem

\bibitem{huang_mutual_2020}
\begin{barticle}
\bauthor{\bsnm{Huang}, \binits{Y.}},
\bauthor{\bsnm{Cai}, \binits{M.}},
\bauthor{\bsnm{Li}, \binits{Z.}},
\bauthor{\bsnm{Lu}, \binits{F.}},
\bauthor{\bsnm{Sato}, \binits{Y.}}:
\batitle{Mutual context network for jointly estimating egocentric gaze and action}.
\bjtitle{IEEE Transactions on Image Processing}
\bvolume{29},
\bfpage{7795}--\blpage{7806}
(\byear{2020})
\end{barticle}
\endbibitem

\bibitem{thakur_predicting_2021}
\begin{bchapter}
\bauthor{\bsnm{Thakur}, \binits{S.K.}},
\bauthor{\bsnm{Beyan}, \binits{C.}},
\bauthor{\bsnm{Morerio}, \binits{P.}},
\bauthor{\bsnm{Del~Bue}, \binits{A.}}:
\bctitle{Predicting gaze from egocentric social interaction videos and imu data}.
In: \bbtitle{Proceedings of the 2021 International Conference on Multimodal Interaction},
pp. \bfpage{717}--\blpage{722}
(\byear{2021})
\end{bchapter}
\endbibitem

\bibitem{vaswani_attention_2017}
\begin{botherref}
\oauthor{\bsnm{Vaswani}, \binits{A.}},
\oauthor{\bsnm{Shazeer}, \binits{N.}},
\oauthor{\bsnm{Parmar}, \binits{N.}},
\oauthor{\bsnm{Uszkoreit}, \binits{J.}},
\oauthor{\bsnm{Jones}, \binits{L.}},
\oauthor{\bsnm{Gomez}, \binits{A.N.}},
\oauthor{\bsnm{Kaiser}, \binits{{\L}.}},
\oauthor{\bsnm{Polosukhin}, \binits{I.}}:
Attention is all you need.
Advances in neural information processing systems
\textbf{30}
(2017)
\end{botherref}
\endbibitem

\bibitem{li_eye_2021}
\begin{barticle}
\bauthor{\bsnm{Li}, \binits{Y.}},
\bauthor{\bsnm{Liu}, \binits{M.}},
\bauthor{\bsnm{Rehg}, \binits{J.M.}}:
\batitle{In the eye of the beholder: Gaze and actions in first person video}.
\bjtitle{IEEE transactions on pattern analysis and machine intelligence}
\bvolume{45}(\bissue{6}),
\bfpage{6731}--\blpage{6747}
(\byear{2021})
\end{barticle}
\endbibitem

\bibitem{li_swingaze_2023}
\begin{bchapter}
\bauthor{\bsnm{Li}, \binits{Y.}},
\bauthor{\bsnm{Wang}, \binits{X.}},
\bauthor{\bsnm{Ma}, \binits{Z.}},
\bauthor{\bsnm{Wang}, \binits{Y.}},
\bauthor{\bsnm{Meyer}, \binits{M.C.}}:
\bctitle{Swingaze: Egocentric gaze estimation with video swin transformer}.
In: \bbtitle{2023 IEEE 16th International Symposium on Embedded Multicore/Many-core Systems-on-Chip (MCSoC)},
pp. \bfpage{123}--\blpage{127}
(\byear{2023}).
\bcomment{IEEE}
\end{bchapter}
\endbibitem

\bibitem{liu_swin_2021}
\begin{bchapter}
\bauthor{\bsnm{Liu}, \binits{Z.}},
\bauthor{\bsnm{Lin}, \binits{Y.}},
\bauthor{\bsnm{Cao}, \binits{Y.}},
\bauthor{\bsnm{Hu}, \binits{H.}},
\bauthor{\bsnm{Wei}, \binits{Y.}},
\bauthor{\bsnm{Zhang}, \binits{Z.}},
\bauthor{\bsnm{Lin}, \binits{S.}},
\bauthor{\bsnm{Guo}, \binits{B.}}:
\bctitle{Swin transformer: Hierarchical vision transformer using shifted windows}.
In: \bbtitle{Proceedings of the IEEE/CVF International Conference on Computer Vision},
pp. \bfpage{10012}--\blpage{10022}
(\byear{2021})
\end{bchapter}
\endbibitem

\bibitem{lv_aria_2024}
\begin{botherref}
\oauthor{\bsnm{Lv}, \binits{Z.}},
\oauthor{\bsnm{Charron}, \binits{N.}},
\oauthor{\bsnm{Moulon}, \binits{P.}},
\oauthor{\bsnm{Gamino}, \binits{A.}},
\oauthor{\bsnm{Peng}, \binits{C.}},
\oauthor{\bsnm{Sweeney}, \binits{C.}},
\oauthor{\bsnm{Miller}, \binits{E.}},
\oauthor{\bsnm{Tang}, \binits{H.}},
\oauthor{\bsnm{Meissner}, \binits{J.}},
\oauthor{\bsnm{Dong}, \binits{J.}}, et al.:
Aria everyday activities dataset.
arXiv preprint arXiv:2402.13349
(2024)
\end{botherref}
\endbibitem

\bibitem{sarafianos_3d_2016}
\begin{barticle}
\bauthor{\bsnm{Sarafianos}, \binits{N.}},
\bauthor{\bsnm{Boteanu}, \binits{B.}},
\bauthor{\bsnm{Ionescu}, \binits{B.}},
\bauthor{\bsnm{Kakadiaris}, \binits{I.A.}}:
\batitle{3d human pose estimation: A review of the literature and analysis of covariates}.
\bjtitle{Computer Vision and Image Understanding}
\bvolume{152},
\bfpage{1}--\blpage{20}
(\byear{2016})
\end{barticle}
\endbibitem

\bibitem{bengamra_review_2021}
\begin{barticle}
\bauthor{\bsnm{Gamra}, \binits{M.B.}},
\bauthor{\bsnm{Akhloufi}, \binits{M.A.}}:
\batitle{A review of deep learning techniques for 2d and 3d human pose estimation}.
\bjtitle{Image and Vision Computing}
\bvolume{114},
\bfpage{104282}
(\byear{2021})
\end{barticle}
\endbibitem

\bibitem{zheng_deep_2023}
\begin{barticle}
\bauthor{\bsnm{Zheng}, \binits{C.}},
\bauthor{\bsnm{Wu}, \binits{W.}},
\bauthor{\bsnm{Chen}, \binits{C.}},
\bauthor{\bsnm{Yang}, \binits{T.}},
\bauthor{\bsnm{Zhu}, \binits{S.}},
\bauthor{\bsnm{Shen}, \binits{J.}},
\bauthor{\bsnm{Kehtarnavaz}, \binits{N.}},
\bauthor{\bsnm{Shah}, \binits{M.}}:
\batitle{Deep learning-based human pose estimation: A survey}.
\bjtitle{ACM Computing Surveys}
\bvolume{56}(\bissue{1}),
\bfpage{1}--\blpage{37}
(\byear{2023})
\end{barticle}
\endbibitem

\bibitem{tome_xregopose_2019}
\begin{bchapter}
\bauthor{\bsnm{Tome}, \binits{D.}},
\bauthor{\bsnm{Peluse}, \binits{P.}},
\bauthor{\bsnm{Agapito}, \binits{L.}},
\bauthor{\bsnm{Badino}, \binits{H.}}:
\bctitle{xr-egopose: Egocentric 3d human pose from an hmd camera}.
In: \bbtitle{Proceedings of the IEEE/CVF International Conference on Computer Vision},
pp. \bfpage{7728}--\blpage{7738}
(\byear{2019})
\end{bchapter}
\endbibitem

\bibitem{xu_mo2cap2_2019}
\begin{barticle}
\bauthor{\bsnm{Xu}, \binits{W.}},
\bauthor{\bsnm{Chatterjee}, \binits{A.}},
\bauthor{\bsnm{Zollhoefer}, \binits{M.}},
\bauthor{\bsnm{Rhodin}, \binits{H.}},
\bauthor{\bsnm{Fua}, \binits{P.}},
\bauthor{\bsnm{Seidel}, \binits{H.-P.}},
\bauthor{\bsnm{Theobalt}, \binits{C.}}:
\batitle{Mo 2 cap 2: Real-time mobile 3d motion capture with a cap-mounted fisheye camera}.
\bjtitle{IEEE transactions on visualization and computer graphics}
\bvolume{25}(\bissue{5}),
\bfpage{2093}--\blpage{2101}
(\byear{2019})
\end{barticle}
\endbibitem

\bibitem{hori_silhouettebased_2022}
\begin{barticle}
\bauthor{\bsnm{Hori}, \binits{R.}},
\bauthor{\bsnm{Hachiuma}, \binits{R.}},
\bauthor{\bsnm{Isogawa}, \binits{M.}},
\bauthor{\bsnm{Mikami}, \binits{D.}},
\bauthor{\bsnm{Saito}, \binits{H.}}:
\batitle{Silhouette-based 3d human pose estimation using a single wrist-mounted 360° camera}.
\bjtitle{IEEE Access}
\bvolume{10},
\bfpage{54957}--\blpage{54968}
(\byear{2022})
\end{barticle}
\endbibitem

\bibitem{ng_you2me_2020}
\begin{bchapter}
\bauthor{\bsnm{Ng}, \binits{E.}},
\bauthor{\bsnm{Xiang}, \binits{D.}},
\bauthor{\bsnm{Joo}, \binits{H.}},
\bauthor{\bsnm{Grauman}, \binits{K.}}:
\bctitle{You2me: Inferring body pose in egocentric video via first and second person interactions}.
In: \bbtitle{Proceedings of the IEEE/CVF Conference on Computer Vision and Pattern Recognition},
pp. \bfpage{9890}--\blpage{9900}
(\byear{2020})
\end{bchapter}
\endbibitem

\bibitem{wang_estimating_2021}
\begin{bchapter}
\bauthor{\bsnm{Wang}, \binits{J.}},
\bauthor{\bsnm{Liu}, \binits{L.}},
\bauthor{\bsnm{Xu}, \binits{W.}},
\bauthor{\bsnm{Sarkar}, \binits{K.}},
\bauthor{\bsnm{Theobalt}, \binits{C.}}:
\bctitle{Estimating egocentric 3d human pose in global space}.
In: \bbtitle{Proceedings of the IEEE/CVF International Conference on Computer Vision},
pp. \bfpage{11500}--\blpage{11509}
(\byear{2021})
\end{bchapter}
\endbibitem

\bibitem{zhao_egoglass_2021}
\begin{bchapter}
\bauthor{\bsnm{Zhao}, \binits{D.}},
\bauthor{\bsnm{Wei}, \binits{Z.}},
\bauthor{\bsnm{Mahmud}, \binits{J.}},
\bauthor{\bsnm{Frahm}, \binits{J.-M.}}:
\bctitle{Egoglass: Egocentric-view human pose estimation from an eyeglass frame}.
In: \bbtitle{2021 International Conference on 3D Vision (3DV)},
pp. \bfpage{32}--\blpage{41}
(\byear{2021}).
\bcomment{IEEE}
\end{bchapter}
\endbibitem

\bibitem{akada_unrealego_2022}
\begin{bchapter}
\bauthor{\bsnm{Akada}, \binits{H.}},
\bauthor{\bsnm{Wang}, \binits{J.}},
\bauthor{\bsnm{Shimada}, \binits{S.}},
\bauthor{\bsnm{Takahashi}, \binits{M.}},
\bauthor{\bsnm{Theobalt}, \binits{C.}},
\bauthor{\bsnm{Golyanik}, \binits{V.}}:
\bctitle{Unrealego: A new dataset for robust egocentric 3d human motion capture}.
In: \bbtitle{European Conference on Computer Vision},
pp. \bfpage{1}--\blpage{17}
(\byear{2022}).
\bcomment{Springer}
\end{bchapter}
\endbibitem

\bibitem{kang_ego3dpose_2023}
\begin{bchapter}
\bauthor{\bsnm{Kang}, \binits{T.}},
\bauthor{\bsnm{Lee}, \binits{K.}},
\bauthor{\bsnm{Zhang}, \binits{J.}},
\bauthor{\bsnm{Lee}, \binits{Y.}}:
\bctitle{Ego3dpose: Capturing 3d cues from binocular egocentric views}.
In: \bbtitle{SIGGRAPH Asia 2023 Conference Papers},
pp. \bfpage{1}--\blpage{10}
(\byear{2023})
\end{bchapter}
\endbibitem

\bibitem{wang_sceneaware_2023}
\begin{bchapter}
\bauthor{\bsnm{Wang}, \binits{J.}},
\bauthor{\bsnm{Luvizon}, \binits{D.}},
\bauthor{\bsnm{Xu}, \binits{W.}},
\bauthor{\bsnm{Liu}, \binits{L.}},
\bauthor{\bsnm{Sarkar}, \binits{K.}},
\bauthor{\bsnm{Theobalt}, \binits{C.}}:
\bctitle{Scene-aware egocentric 3d human pose estimation}.
In: \bbtitle{Proceedings of the IEEE/CVF Conference on Computer Vision and Pattern Recognition},
pp. \bfpage{13031}--\blpage{13040}
(\byear{2023})
\end{bchapter}
\endbibitem

\bibitem{kang_attentionpropagation_2024}
\begin{bchapter}
\bauthor{\bsnm{Kang}, \binits{T.}},
\bauthor{\bsnm{Lee}, \binits{Y.}}:
\bctitle{Attention-propagation network for egocentric heatmap to 3d pose lifting}.
In: \bbtitle{Proceedings of the IEEE/CVF Conference on Computer Vision and Pattern Recognition},
pp. \bfpage{842}--\blpage{851}
(\byear{2024})
\end{bchapter}
\endbibitem

\bibitem{akada_3d_2024}
\begin{bchapter}
\bauthor{\bsnm{Akada}, \binits{H.}},
\bauthor{\bsnm{Wang}, \binits{J.}},
\bauthor{\bsnm{Golyanik}, \binits{V.}},
\bauthor{\bsnm{Theobalt}, \binits{C.}}:
\bctitle{3d human pose perception from egocentric stereo videos}.
In: \bbtitle{Proceedings of the IEEE/CVF Conference on Computer Vision and Pattern Recognition},
pp. \bfpage{767}--\blpage{776}
(\byear{2024})
\end{bchapter}
\endbibitem

\bibitem{zhao_egobody3m_2024}
\begin{bchapter}
\bauthor{\bsnm{Zhao}, \binits{A.}},
\bauthor{\bsnm{Tang}, \binits{C.}},
\bauthor{\bsnm{Wang}, \binits{L.}},
\bauthor{\bsnm{Li}, \binits{Y.}},
\bauthor{\bsnm{Dave}, \binits{M.}},
\bauthor{\bsnm{Tao}, \binits{L.}},
\bauthor{\bsnm{Twigg}, \binits{C.D.}},
\bauthor{\bsnm{Wang}, \binits{R.Y.}}:
\bctitle{Egobody3m: Egocentric body tracking on a vr headset using a diverse dataset}.
In: \bbtitle{European Conference on Computer Vision},
pp. \bfpage{375}--\blpage{392}
(\byear{2024}).
\bcomment{Springer}
\end{bchapter}
\endbibitem

\bibitem{millerdurai_eventego3d_2024}
\begin{bchapter}
\bauthor{\bsnm{Millerdurai}, \binits{C.}},
\bauthor{\bsnm{Akada}, \binits{H.}},
\bauthor{\bsnm{Wang}, \binits{J.}},
\bauthor{\bsnm{Luvizon}, \binits{D.}},
\bauthor{\bsnm{Theobalt}, \binits{C.}},
\bauthor{\bsnm{Golyanik}, \binits{V.}}:
\bctitle{Eventego3d: 3d human motion capture from egocentric event streams}.
In: \bbtitle{Proceedings of the IEEE/CVF Conference on Computer Vision and Pattern Recognition},
pp. \bfpage{1186}--\blpage{1195}
(\byear{2024})
\end{bchapter}
\endbibitem

\bibitem{loper_smpl_2015}
\begin{bchapter}
\bauthor{\bsnm{Loper}, \binits{M.}},
\bauthor{\bsnm{Mahmood}, \binits{N.}},
\bauthor{\bsnm{Romero}, \binits{J.}},
\bauthor{\bsnm{Pons-Moll}, \binits{G.}},
\bauthor{\bsnm{Black}, \binits{M.J.}}:
\bctitle{Smpl: A skinned multi-person linear model}.
In: \bbtitle{Seminal Graphics Papers: Pushing the Boundaries, Volume 2},
pp. \bfpage{851}--\blpage{866}
(\byear{2023})
\end{bchapter}
\endbibitem

\bibitem{pavlakos_expressive_2019}
\begin{bchapter}
\bauthor{\bsnm{Pavlakos}, \binits{G.}},
\bauthor{\bsnm{Choutas}, \binits{V.}},
\bauthor{\bsnm{Ghorbani}, \binits{N.}},
\bauthor{\bsnm{Bolkart}, \binits{T.}},
\bauthor{\bsnm{Osman}, \binits{A.A.}},
\bauthor{\bsnm{Tzionas}, \binits{D.}},
\bauthor{\bsnm{Black}, \binits{M.J.}}:
\bctitle{Expressive body capture: 3d hands, face, and body from a single image}.
In: \bbtitle{Proceedings of the IEEE/CVF Conference on Computer Vision and Pattern Recognition},
pp. \bfpage{10975}--\blpage{10985}
(\byear{2019})
\end{bchapter}
\endbibitem

\bibitem{guzov_human_2021}
\begin{bchapter}
\bauthor{\bsnm{Guzov}, \binits{V.}},
\bauthor{\bsnm{Mir}, \binits{A.}},
\bauthor{\bsnm{Sattler}, \binits{T.}},
\bauthor{\bsnm{Pons-Moll}, \binits{G.}}:
\bctitle{Human poseitioning system (hps): 3d human pose estimation and self-localization in large scenes from body-mounted sensors}.
In: \bbtitle{Proceedings of the IEEE/CVF Conference on Computer Vision and Pattern Recognition},
pp. \bfpage{4318}--\blpage{4329}
(\byear{2021})
\end{bchapter}
\endbibitem

\bibitem{jiang_avatarposer_2022}
\begin{bchapter}
\bauthor{\bsnm{Jiang}, \binits{J.}},
\bauthor{\bsnm{Streli}, \binits{P.}},
\bauthor{\bsnm{Qiu}, \binits{H.}},
\bauthor{\bsnm{Fender}, \binits{A.}},
\bauthor{\bsnm{Laich}, \binits{L.}},
\bauthor{\bsnm{Snape}, \binits{P.}},
\bauthor{\bsnm{Holz}, \binits{C.}}:
\bctitle{Avatarposer: Articulated full-body pose tracking from sparse motion sensing}.
In: \bbtitle{European Conference on Computer Vision},
pp. \bfpage{443}--\blpage{460}
(\byear{2022}).
\bcomment{Springer}
\end{bchapter}
\endbibitem

\bibitem{dittadi_fullbody_2021}
\begin{bchapter}
\bauthor{\bsnm{Dittadi}, \binits{A.}},
\bauthor{\bsnm{Dziadzio}, \binits{S.}},
\bauthor{\bsnm{Cosker}, \binits{D.}},
\bauthor{\bsnm{Lundell}, \binits{B.}},
\bauthor{\bsnm{Cashman}, \binits{T.J.}},
\bauthor{\bsnm{Shotton}, \binits{J.}}:
\bctitle{Full-body motion from a single head-mounted device: Generating smpl poses from partial observations}.
In: \bbtitle{Proceedings of the IEEE/CVF International Conference on Computer Vision},
pp. \bfpage{11687}--\blpage{11697}
(\byear{2021})
\end{bchapter}
\endbibitem

\bibitem{aliakbarian_flag_2022}
\begin{bchapter}
\bauthor{\bsnm{Aliakbarian}, \binits{S.}},
\bauthor{\bsnm{Cameron}, \binits{P.}},
\bauthor{\bsnm{Bogo}, \binits{F.}},
\bauthor{\bsnm{Fitzgibbon}, \binits{A.}},
\bauthor{\bsnm{Cashman}, \binits{T.J.}}:
\bctitle{Flag: Flow-based 3d avatar generation from sparse observations}.
In: \bbtitle{Proceedings of the IEEE/CVF Conference on Computer Vision and Pattern Recognition},
pp. \bfpage{13253}--\blpage{13262}
(\byear{2022})
\end{bchapter}
\endbibitem

\bibitem{du_avatars_2023}
\begin{bchapter}
\bauthor{\bsnm{Du}, \binits{Y.}},
\bauthor{\bsnm{Kips}, \binits{R.}},
\bauthor{\bsnm{Pumarola}, \binits{A.}},
\bauthor{\bsnm{Starke}, \binits{S.}},
\bauthor{\bsnm{Thabet}, \binits{A.}},
\bauthor{\bsnm{Sanakoyeu}, \binits{A.}}:
\bctitle{Avatars grow legs: Generating smooth human motion from sparse tracking inputs with diffusion model}.
In: \bbtitle{Proceedings of the IEEE/CVF Conference on Computer Vision and Pattern Recognition},
pp. \bfpage{481}--\blpage{490}
(\byear{2023})
\end{bchapter}
\endbibitem

\bibitem{jiang_egoposer_2024}
\begin{bchapter}
\bauthor{\bsnm{Jiang}, \binits{J.}},
\bauthor{\bsnm{Streli}, \binits{P.}},
\bauthor{\bsnm{Meier}, \binits{M.}},
\bauthor{\bsnm{Holz}, \binits{C.}}:
\bctitle{Egoposer: Robust real-time egocentric pose estimation from sparse and intermittent observations everywhere}.
In: \bbtitle{European Conference on Computer Vision},
pp. \bfpage{277}--\blpage{294}
(\byear{2024}).
\bcomment{Springer}
\end{bchapter}
\endbibitem

\bibitem{kwon_h2o_2021}
\begin{bchapter}
\bauthor{\bsnm{Kwon}, \binits{T.}},
\bauthor{\bsnm{Tekin}, \binits{B.}},
\bauthor{\bsnm{St{\"u}hmer}, \binits{J.}},
\bauthor{\bsnm{Bogo}, \binits{F.}},
\bauthor{\bsnm{Pollefeys}, \binits{M.}}:
\bctitle{H2o: Two hands manipulating objects for first person interaction recognition}.
In: \bbtitle{Proceedings of the IEEE/CVF International Conference on Computer Vision},
pp. \bfpage{10138}--\blpage{10148}
(\byear{2021})
\end{bchapter}
\endbibitem

\bibitem{fan_arctic_2023}
\begin{bchapter}
\bauthor{\bsnm{Fan}, \binits{Z.}},
\bauthor{\bsnm{Taheri}, \binits{O.}},
\bauthor{\bsnm{Tzionas}, \binits{D.}},
\bauthor{\bsnm{Kocabas}, \binits{M.}},
\bauthor{\bsnm{Kaufmann}, \binits{M.}},
\bauthor{\bsnm{Black}, \binits{M.J.}},
\bauthor{\bsnm{Hilliges}, \binits{O.}}:
\bctitle{Arctic: A dataset for dexterous bimanual hand-object manipulation}.
In: \bbtitle{Proceedings of the IEEE/CVF Conference on Computer Vision and Pattern Recognition},
pp. \bfpage{12943}--\blpage{12954}
(\byear{2023})
\end{bchapter}
\endbibitem

\bibitem{pavlakos_reconstructing_2024}
\begin{bchapter}
\bauthor{\bsnm{Pavlakos}, \binits{G.}},
\bauthor{\bsnm{Shan}, \binits{D.}},
\bauthor{\bsnm{Radosavovic}, \binits{I.}},
\bauthor{\bsnm{Kanazawa}, \binits{A.}},
\bauthor{\bsnm{Fouhey}, \binits{D.}},
\bauthor{\bsnm{Malik}, \binits{J.}}:
\bctitle{Reconstructing hands in 3d with transformers}.
In: \bbtitle{Proceedings of the IEEE/CVF Conference on Computer Vision and Pattern Recognition},
pp. \bfpage{9826}--\blpage{9836}
(\byear{2024})
\end{bchapter}
\endbibitem

\bibitem{zhou_1st_2023}
\begin{botherref}
\oauthor{\bsnm{Zhou}, \binits{Z.}},
\oauthor{\bsnm{Lv}, \binits{Z.}},
\oauthor{\bsnm{Zhou}, \binits{S.}},
\oauthor{\bsnm{Zou}, \binits{M.}},
\oauthor{\bsnm{Wu}, \binits{T.}},
\oauthor{\bsnm{Yu}, \binits{M.}},
\oauthor{\bsnm{Tang}, \binits{Y.}},
\oauthor{\bsnm{Liang}, \binits{J.}}:
1st place solution of egocentric 3d hand pose estimation challenge 2023 technical report: A concise pipeline for egocentric hand pose reconstruction.
arXiv preprint arXiv:2310.04769
(2023)
\end{botherref}
\endbibitem

\bibitem{liu_singletodualview_2024}
\begin{bchapter}
\bauthor{\bsnm{Liu}, \binits{R.}},
\bauthor{\bsnm{Ohkawa}, \binits{T.}},
\bauthor{\bsnm{Zhang}, \binits{M.}},
\bauthor{\bsnm{Sato}, \binits{Y.}}:
\bctitle{Single-to-dual-view adaptation for egocentric 3d hand pose estimation}.
In: \bbtitle{Proceedings of the IEEE/CVF Conference on Computer Vision and Pattern Recognition},
pp. \bfpage{677}--\blpage{686}
(\byear{2024})
\end{bchapter}
\endbibitem

\bibitem{prakash_3d_2024}
\begin{bchapter}
\bauthor{\bsnm{Prakash}, \binits{A.}},
\bauthor{\bsnm{Tu}, \binits{R.}},
\bauthor{\bsnm{Chang}, \binits{M.}},
\bauthor{\bsnm{Gupta}, \binits{S.}}:
\bctitle{3d hand pose estimation in everyday egocentric images}.
In: \bbtitle{European Conference on Computer Vision},
pp. \bfpage{183}--\blpage{202}
(\byear{2024}).
\bcomment{Springer}
\end{bchapter}
\endbibitem

\bibitem{mahmood_amass_2019}
\begin{bchapter}
\bauthor{\bsnm{Mahmood}, \binits{N.}},
\bauthor{\bsnm{Ghorbani}, \binits{N.}},
\bauthor{\bsnm{Troje}, \binits{N.F.}},
\bauthor{\bsnm{Pons-Moll}, \binits{G.}},
\bauthor{\bsnm{Black}, \binits{M.J.}}:
\bctitle{Amass: Archive of motion capture as surface shapes}.
In: \bbtitle{Proceedings of the IEEE/CVF International Conference on Computer Vision},
pp. \bfpage{5442}--\blpage{5451}
(\byear{2019})
\end{bchapter}
\endbibitem

\bibitem{ohkawa_assemblyhands_2023}
\begin{bchapter}
\bauthor{\bsnm{Ohkawa}, \binits{T.}},
\bauthor{\bsnm{He}, \binits{K.}},
\bauthor{\bsnm{Sener}, \binits{F.}},
\bauthor{\bsnm{Hodan}, \binits{T.}},
\bauthor{\bsnm{Tran}, \binits{L.}},
\bauthor{\bsnm{Keskin}, \binits{C.}}:
\bctitle{Assemblyhands: Towards egocentric activity understanding via 3d hand pose estimation}.
In: \bbtitle{Proceedings of the IEEE/CVF Conference on Computer Vision and Pattern Recognition},
pp. \bfpage{12999}--\blpage{13008}
(\byear{2023})
\end{bchapter}
\endbibitem

\bibitem{kong_human_2022}
\begin{barticle}
\bauthor{\bsnm{Kong}, \binits{Y.}},
\bauthor{\bsnm{Fu}, \binits{Y.}}:
\batitle{Human action recognition and prediction: A survey}.
\bjtitle{International Journal of Computer Vision}
\bvolume{130}(\bissue{5}),
\bfpage{1366}--\blpage{1401}
(\byear{2022})
\end{barticle}
\endbibitem

\bibitem{hu_online_2022}
\begin{barticle}
\bauthor{\bsnm{Hu}, \binits{X.}},
\bauthor{\bsnm{Dai}, \binits{J.}},
\bauthor{\bsnm{Li}, \binits{M.}},
\bauthor{\bsnm{Peng}, \binits{C.}},
\bauthor{\bsnm{Li}, \binits{Y.}},
\bauthor{\bsnm{Du}, \binits{S.}}:
\batitle{Online human action detection and anticipation in videos: A survey}.
\bjtitle{Neurocomputing}
\bvolume{491},
\bfpage{395}--\blpage{413}
(\byear{2022})
\end{barticle}
\endbibitem

\bibitem{lai_human_2024}
\begin{botherref}
\oauthor{\bsnm{Lai}, \binits{B.}},
\oauthor{\bsnm{Toyer}, \binits{S.}},
\oauthor{\bsnm{Nagarajan}, \binits{T.}},
\oauthor{\bsnm{Girdhar}, \binits{R.}},
\oauthor{\bsnm{Zha}, \binits{S.}},
\oauthor{\bsnm{Rehg}, \binits{J.M.}},
\oauthor{\bsnm{Kitani}, \binits{K.}},
\oauthor{\bsnm{Grauman}, \binits{K.}},
\oauthor{\bsnm{Desai}, \binits{R.}},
\oauthor{\bsnm{Liu}, \binits{M.}}:
Human action anticipation: A survey.
arXiv preprint arXiv:2410.14045
(2024)
\end{botherref}
\endbibitem

\bibitem{ding_temporal_2024}
\begin{botherref}
\oauthor{\bsnm{Ding}, \binits{G.}},
\oauthor{\bsnm{Sener}, \binits{F.}},
\oauthor{\bsnm{Yao}, \binits{A.}}:
Temporal action segmentation: An analysis of modern techniques.
IEEE Transactions on Pattern Analysis and Machine Intelligence
(2024)
\end{botherref}
\endbibitem

\bibitem{damen_rescaling_2022}
\begin{botherref}
\oauthor{\bsnm{Damen}, \binits{D.}},
\oauthor{\bsnm{Doughty}, \binits{H.}},
\oauthor{\bsnm{Farinella}, \binits{G.M.}},
\oauthor{\bsnm{Furnari}, \binits{A.}},
\oauthor{\bsnm{Kazakos}, \binits{E.}},
\oauthor{\bsnm{Ma}, \binits{J.}},
\oauthor{\bsnm{Moltisanti}, \binits{D.}},
\oauthor{\bsnm{Munro}, \binits{J.}},
\oauthor{\bsnm{Perrett}, \binits{T.}},
\oauthor{\bsnm{Price}, \binits{W.}}, et al.:
Rescaling egocentric vision: Collection, pipeline and challenges for epic-kitchens-100.
International Journal of Computer Vision,
1--23
(2022)
\end{botherref}
\endbibitem

\bibitem{herzig_objectregion_2022}
\begin{bchapter}
\bauthor{\bsnm{Herzig}, \binits{R.}},
\bauthor{\bsnm{Ben-Avraham}, \binits{E.}},
\bauthor{\bsnm{Mangalam}, \binits{K.}},
\bauthor{\bsnm{Bar}, \binits{A.}},
\bauthor{\bsnm{Chechik}, \binits{G.}},
\bauthor{\bsnm{Rohrbach}, \binits{A.}},
\bauthor{\bsnm{Darrell}, \binits{T.}},
\bauthor{\bsnm{Globerson}, \binits{A.}}:
\bctitle{Object-region video transformers}.
In: \bbtitle{Proceedings of the Ieee/cvf Conference on Computer Vision and Pattern Recognition},
pp. \bfpage{3148}--\blpage{3159}
(\byear{2022})
\end{bchapter}
\endbibitem

\bibitem{wang_learning_2023}
\begin{bchapter}
\bauthor{\bsnm{Wang}, \binits{Q.}},
\bauthor{\bsnm{Zhao}, \binits{L.}},
\bauthor{\bsnm{Yuan}, \binits{L.}},
\bauthor{\bsnm{Liu}, \binits{T.}},
\bauthor{\bsnm{Peng}, \binits{X.}}:
\bctitle{Learning from semantic alignment between unpaired multiviews for egocentric video recognition}.
In: \bbtitle{Proceedings of the IEEE/CVF International Conference on Computer Vision},
pp. \bfpage{3307}--\blpage{3317}
(\byear{2023})
\end{bchapter}
\endbibitem

\bibitem{shiota_egocentric_2024}
\begin{bchapter}
\bauthor{\bsnm{Shiota}, \binits{T.}},
\bauthor{\bsnm{Takagi}, \binits{M.}},
\bauthor{\bsnm{Kumagai}, \binits{K.}},
\bauthor{\bsnm{Seshimo}, \binits{H.}},
\bauthor{\bsnm{Aono}, \binits{Y.}}:
\bctitle{Egocentric action recognition by capturing hand-object contact and object state}.
In: \bbtitle{Proceedings of the IEEE/CVF Winter Conference on Applications of Computer Vision},
pp. \bfpage{6541}--\blpage{6551}
(\byear{2024})
\end{bchapter}
\endbibitem

\bibitem{gundavarapu_extending_2024}
\begin{bchapter}
\bauthor{\bsnm{Gundavarapu}, \binits{N.B.}},
\bauthor{\bsnm{Friedman}, \binits{L.}},
\bauthor{\bsnm{Goyal}, \binits{R.}},
\bauthor{\bsnm{Hegde}, \binits{C.}},
\bauthor{\bsnm{Agustsson}, \binits{E.}},
\bauthor{\bsnm{Waghmare}, \binits{S.M.}},
\bauthor{\bsnm{Sirotenko}, \binits{M.}},
\bauthor{\bsnm{Yang}, \binits{M.-H.}},
\bauthor{\bsnm{Weyand}, \binits{T.}},
\bauthor{\bsnm{Gong}, \binits{B.}},
\bauthor{\bsnm{Sigal}, \binits{L.}}:
\bctitle{Extending video masked autoencoders to 128 frames}.
In: \bbtitle{The Thirty-eighth Annual Conference on Neural Information Processing Systems}
(\byear{2024}).
\burl{https://openreview.net/forum?id=bFrNPlWchg}
\end{bchapter}
\endbibitem

\bibitem{annakukleva_xmic_2024}
\begin{bchapter}
\bauthor{\bsnm{Kukleva}, \binits{A.}},
\bauthor{\bsnm{Sener}, \binits{F.}},
\bauthor{\bsnm{Remelli}, \binits{E.}},
\bauthor{\bsnm{Tekin}, \binits{B.}},
\bauthor{\bsnm{Sauser}, \binits{E.}},
\bauthor{\bsnm{Schiele}, \binits{B.}},
\bauthor{\bsnm{Ma}, \binits{S.}}:
\bctitle{X-mic: Cross-modal instance conditioning for egocentric action generalization}.
In: \bbtitle{Proceedings of the IEEE/CVF Conference on Computer Vision and Pattern Recognition},
pp. \bfpage{26364}--\blpage{26373}
(\byear{2024})
\end{bchapter}
\endbibitem

\bibitem{zhang_actionformer_2022}
\begin{bchapter}
\bauthor{\bsnm{Zhang}, \binits{C.}},
\bauthor{\bsnm{Wu}, \binits{J.}},
\bauthor{\bsnm{Li}, \binits{Y.}}:
\bctitle{Actionformer: Localizing moments of actions with transformers}.
In: \bbtitle{Proceedings of the European Conference on Computer Vision (ECCV)},
pp. \bfpage{492}--\blpage{510}
(\byear{2022}).
\bcomment{Springer}
\end{bchapter}
\endbibitem

\bibitem{shi_tridet_2023}
\begin{bchapter}
\bauthor{\bsnm{Shi}, \binits{D.}},
\bauthor{\bsnm{Zhong}, \binits{Y.}},
\bauthor{\bsnm{Cao}, \binits{Q.}},
\bauthor{\bsnm{Ma}, \binits{L.}},
\bauthor{\bsnm{Li}, \binits{J.}},
\bauthor{\bsnm{Tao}, \binits{D.}}:
\bctitle{Tridet: Temporal action detection with relative boundary modeling}.
In: \bbtitle{Proceedings of the IEEE/CVF Conference on Computer Vision and Pattern Recognition},
pp. \bfpage{18857}--\blpage{18866}
(\byear{2023})
\end{bchapter}
\endbibitem

\bibitem{wang_egoonly_2023}
\begin{bchapter}
\bauthor{\bsnm{Wang}, \binits{H.}},
\bauthor{\bsnm{Singh}, \binits{M.K.}},
\bauthor{\bsnm{Torresani}, \binits{L.}}:
\bctitle{Ego-only: Egocentric action detection without exocentric transferring}.
In: \bbtitle{Proceedings of the IEEE/CVF International Conference on Computer Vision},
pp. \bfpage{5250}--\blpage{5261}
(\byear{2023})
\end{bchapter}
\endbibitem

\bibitem{liu_endtoend_2024}
\begin{bchapter}
\bauthor{\bsnm{Liu}, \binits{S.}},
\bauthor{\bsnm{Zhang}, \binits{C.-L.}},
\bauthor{\bsnm{Zhao}, \binits{C.}},
\bauthor{\bsnm{Ghanem}, \binits{B.}}:
\bctitle{End-to-end temporal action detection with 1b parameters across 1000 frames}.
In: \bbtitle{Proceedings of the IEEE/CVF Conference on Computer Vision and Pattern Recognition},
pp. \bfpage{18591}--\blpage{18601}
(\byear{2024})
\end{bchapter}
\endbibitem

\bibitem{guo_uncertaintyaware_2024}
\begin{bchapter}
\bauthor{\bsnm{Guo}, \binits{H.}},
\bauthor{\bsnm{Agarwal}, \binits{N.}},
\bauthor{\bsnm{Lo}, \binits{S.-Y.}},
\bauthor{\bsnm{Lee}, \binits{K.}},
\bauthor{\bsnm{Ji}, \binits{Q.}}:
\bctitle{Uncertainty-aware action decoupling transformer for action anticipation}.
In: \bbtitle{Proceedings of the IEEE/CVF Conference on Computer Vision and Pattern Recognition},
pp. \bfpage{18644}--\blpage{18654}
(\byear{2024})
\end{bchapter}
\endbibitem

\bibitem{roy_interaction_2024}
\begin{bchapter}
\bauthor{\bsnm{Roy}, \binits{D.}},
\bauthor{\bsnm{Rajendiran}, \binits{R.}},
\bauthor{\bsnm{Fernando}, \binits{B.}}:
\bctitle{Interaction region visual transformer for egocentric action anticipation}.
In: \bbtitle{Proceedings of the IEEE/CVF Winter Conference on Applications of Computer Vision},
pp. \bfpage{6740}--\blpage{6750}
(\byear{2024})
\end{bchapter}
\endbibitem

\bibitem{diko_semantically_2024}
\begin{bchapter}
\bauthor{\bsnm{Diko}, \binits{A.}},
\bauthor{\bsnm{Avola}, \binits{D.}},
\bauthor{\bsnm{Prenkaj}, \binits{B.}},
\bauthor{\bsnm{Fontana}, \binits{F.}},
\bauthor{\bsnm{Cinque}, \binits{L.}}:
\bctitle{Semantically guided representation learning for action anticipation}.
In: \bbtitle{European Conference on Computer Vision},
pp. \bfpage{448}--\blpage{466}
(\byear{2024}).
\bcomment{Springer}
\end{bchapter}
\endbibitem

\bibitem{thakur_leveraging_2024}
\begin{bchapter}
\bauthor{\bsnm{Thakur}, \binits{S.}},
\bauthor{\bsnm{Beyan}, \binits{C.}},
\bauthor{\bsnm{Morerio}, \binits{P.}},
\bauthor{\bsnm{Murino}, \binits{V.}},
\bauthor{\bsnm{Del~Bue}, \binits{A.}}:
\bctitle{Leveraging next-active objects for context-aware anticipation in egocentric videos}.
In: \bbtitle{Proceedings of the IEEE/CVF Winter Conference on Applications of Computer Vision},
pp. \bfpage{8657}--\blpage{8666}
(\byear{2024})
\end{bchapter}
\endbibitem

\bibitem{kim_palm_2024}
\begin{bchapter}
\bauthor{\bsnm{Kim}, \binits{S.}},
\bauthor{\bsnm{Huang}, \binits{D.}},
\bauthor{\bsnm{Xian}, \binits{Y.}},
\bauthor{\bsnm{Hilliges}, \binits{O.}},
\bauthor{\bsnm{Van~Gool}, \binits{L.}},
\bauthor{\bsnm{Wang}, \binits{X.}}:
\bctitle{Palm: Predicting actions through language models}.
In: \bbtitle{European Conference on Computer Vision},
pp. \bfpage{140}--\blpage{158}
(\byear{2024}).
\bcomment{Springer}
\end{bchapter}
\endbibitem

\bibitem{mittal_cant_2024}
\begin{bchapter}
\bauthor{\bsnm{Mittal}, \binits{H.}},
\bauthor{\bsnm{Agarwal}, \binits{N.}},
\bauthor{\bsnm{Lo}, \binits{S.-Y.}},
\bauthor{\bsnm{Lee}, \binits{K.}}:
\bctitle{Can't make an omelette without breaking some eggs: Plausible action anticipation using large video-language models}.
In: \bbtitle{Proceedings of the IEEE/CVF Conference on Computer Vision and Pattern Recognition},
pp. \bfpage{18580}--\blpage{18590}
(\byear{2024})
\end{bchapter}
\endbibitem

\bibitem{kazakos_epicfusion_2019}
\begin{bchapter}
\bauthor{\bsnm{Kazakos}, \binits{E.}},
\bauthor{\bsnm{Nagrani}, \binits{A.}},
\bauthor{\bsnm{Zisserman}, \binits{A.}},
\bauthor{\bsnm{Damen}, \binits{D.}}:
\bctitle{Epic-fusion: Audio-visual temporal binding for egocentric action recognition}.
In: \bbtitle{Proceedings of the IEEE/CVF International Conference on Computer Vision},
pp. \bfpage{5492}--\blpage{5501}
(\byear{2019})
\end{bchapter}
\endbibitem

\bibitem{min_integrating_2021}
\begin{bchapter}
\bauthor{\bsnm{Min}, \binits{K.}},
\bauthor{\bsnm{Corso}, \binits{J.J.}}:
\bctitle{Integrating human gaze into attention for egocentric activity recognition}.
In: \bbtitle{Proceedings of the IEEE/CVF Winter Conference on Applications of Computer Vision},
pp. \bfpage{1069}--\blpage{1078}
(\byear{2021})
\end{bchapter}
\endbibitem

\bibitem{thapar_sharing_2020}
\begin{bchapter}
\bauthor{\bsnm{Thapar}, \binits{D.}},
\bauthor{\bsnm{Arora}, \binits{C.}},
\bauthor{\bsnm{Nigam}, \binits{A.}}:
\bctitle{Is sharing of egocentric video giving away your biometric signature?}
In: \bbtitle{Computer Vision--ECCV 2020: 16th European Conference, Glasgow, UK, August 23--28, 2020, Proceedings, Part XVII 16},
pp. \bfpage{399}--\blpage{416}
(\byear{2020}).
\bcomment{Springer}
\end{bchapter}
\endbibitem

\bibitem{plizzari_e2_2022}
\begin{bchapter}
\bauthor{\bsnm{Plizzari}, \binits{C.}},
\bauthor{\bsnm{Planamente}, \binits{M.}},
\bauthor{\bsnm{Goletto}, \binits{G.}},
\bauthor{\bsnm{Cannici}, \binits{M.}},
\bauthor{\bsnm{Gusso}, \binits{E.}},
\bauthor{\bsnm{Matteucci}, \binits{M.}},
\bauthor{\bsnm{Caputo}, \binits{B.}}:
\bctitle{E2 (go) motion: Motion augmented event stream for egocentric action recognition}.
In: \bbtitle{Proceedings of the IEEE/CVF Conference on Computer Vision and Pattern Recognition},
pp. \bfpage{19935}--\blpage{19947}
(\byear{2022})
\end{bchapter}
\endbibitem

\bibitem{li_egoexo_2021}
\begin{bchapter}
\bauthor{\bsnm{Li}, \binits{Y.}},
\bauthor{\bsnm{Nagarajan}, \binits{T.}},
\bauthor{\bsnm{Xiong}, \binits{B.}},
\bauthor{\bsnm{Grauman}, \binits{K.}}:
\bctitle{Ego-exo: Transferring visual representations from third-person to first-person videos}.
In: \bbtitle{Proceedings of the IEEE/CVF Conference on Computer Vision and Pattern Recognition},
pp. \bfpage{6943}--\blpage{6953}
(\byear{2021})
\end{bchapter}
\endbibitem

\bibitem{tan_egodistill_2023}
\begin{barticle}
\bauthor{\bsnm{Tan}, \binits{S.}},
\bauthor{\bsnm{Nagarajan}, \binits{T.}},
\bauthor{\bsnm{Grauman}, \binits{K.}}:
\batitle{Egodistill: Egocentric head motion distillation for efficient video understanding}.
\bjtitle{Advances in Neural Information Processing Systems}
\bvolume{36},
\bfpage{33485}--\blpage{33498}
(\byear{2023})
\end{barticle}
\endbibitem

\bibitem{gong_mmgego4d_2023}
\begin{bchapter}
\bauthor{\bsnm{Gong}, \binits{X.}},
\bauthor{\bsnm{Mohan}, \binits{S.}},
\bauthor{\bsnm{Dhingra}, \binits{N.}},
\bauthor{\bsnm{Bazin}, \binits{J.-C.}},
\bauthor{\bsnm{Li}, \binits{Y.}},
\bauthor{\bsnm{Wang}, \binits{Z.}},
\bauthor{\bsnm{Ranjan}, \binits{R.}}:
\bctitle{Mmg-ego4d: Multimodal generalization in egocentric action recognition}.
In: \bbtitle{Proceedings of the IEEE/CVF Conference on Computer Vision and Pattern Recognition},
pp. \bfpage{6481}--\blpage{6491}
(\byear{2023})
\end{bchapter}
\endbibitem

\bibitem{wu_memvit_2022}
\begin{bchapter}
\bauthor{\bsnm{Wu}, \binits{C.-Y.}},
\bauthor{\bsnm{Li}, \binits{Y.}},
\bauthor{\bsnm{Mangalam}, \binits{K.}},
\bauthor{\bsnm{Fan}, \binits{H.}},
\bauthor{\bsnm{Xiong}, \binits{B.}},
\bauthor{\bsnm{Malik}, \binits{J.}},
\bauthor{\bsnm{Feichtenhofer}, \binits{C.}}:
\bctitle{Memvit: Memory-augmented multiscale vision transformer for efficient long-term video recognition}.
In: \bbtitle{Proceedings of the IEEE/CVF Conference on Computer Vision and Pattern Recognition},
pp. \bfpage{13587}--\blpage{13597}
(\byear{2022})
\end{bchapter}
\endbibitem

\bibitem{yan_multiview_2022}
\begin{bchapter}
\bauthor{\bsnm{Yan}, \binits{S.}},
\bauthor{\bsnm{Xiong}, \binits{X.}},
\bauthor{\bsnm{Arnab}, \binits{A.}},
\bauthor{\bsnm{Lu}, \binits{Z.}},
\bauthor{\bsnm{Zhang}, \binits{M.}},
\bauthor{\bsnm{Sun}, \binits{C.}},
\bauthor{\bsnm{Schmid}, \binits{C.}}:
\bctitle{Multiview transformers for video recognition}.
In: \bbtitle{Proceedings of the IEEE/CVF Conference on Computer Vision and Pattern Recognition},
pp. \bfpage{3333}--\blpage{3343}
(\byear{2022})
\end{bchapter}
\endbibitem

\bibitem{lu_mixed_2025}
\begin{barticle}
\bauthor{\bsnm{Lu}, \binits{X.}},
\bauthor{\bsnm{Hao}, \binits{Y.}},
\bauthor{\bsnm{Cheng}, \binits{L.}},
\bauthor{\bsnm{Zhao}, \binits{S.}},
\bauthor{\bsnm{Liu}, \binits{Y.}},
\bauthor{\bsnm{Song}, \binits{M.}}:
\batitle{Mixed {{Attention}} and {{Channel Shift Transformer}} for {{Efficient Action Recognition}}}.
\bjtitle{ACM Trans. Multimedia Comput. Commun. Appl.}
\bvolume{21}(\bissue{3}),
\bfpage{93}--\blpage{19320}
(\byear{2025})
\end{barticle}
\endbibitem

\bibitem{chen_skeletonbased_2025}
\begin{bchapter}
\bauthor{\bsnm{Chen}, \binits{H.}},
\bauthor{\bsnm{Yang}, \binits{Y.}},
\bauthor{\bsnm{Lyu}, \binits{Y.}}:
\bctitle{Skeleton-based {{Action Recognition}} with {{Non-linear Dependency Modeling}} and {{Hilbert-Schmidt Independence Criterion}}}.
In: \bbtitle{Proceedings of the {{AAAI Conference}} on {{Artificial Intelligence}}},
vol. \bseriesno{39},
pp. \bfpage{2043}--\blpage{2051}
(\byear{2025})
\end{bchapter}
\endbibitem

\bibitem{geng_hierarchical_2024}
\begin{barticle}
\bauthor{\bsnm{Geng}, \binits{P.}},
\bauthor{\bsnm{Lu}, \binits{X.}},
\bauthor{\bsnm{Li}, \binits{W.}},
\bauthor{\bsnm{Lyu}, \binits{L.}}:
\batitle{Hierarchical {{Aggregated Graph Neural Network}} for {{Skeleton-Based Action Recognition}}}.
\bjtitle{IEEE Transactions on Multimedia}
\bvolume{26},
\bfpage{11003}--\blpage{11017}
(\byear{2024})
\end{barticle}
\endbibitem

\bibitem{liu_knowledgebased_2024}
\begin{barticle}
\bauthor{\bsnm{Liu}, \binits{Y.}},
\bauthor{\bsnm{Liu}, \binits{F.}},
\bauthor{\bsnm{Jiao}, \binits{L.}},
\bauthor{\bsnm{Bao}, \binits{Q.}},
\bauthor{\bsnm{Li}, \binits{L.}},
\bauthor{\bsnm{Guo}, \binits{Y.}},
\bauthor{\bsnm{Chen}, \binits{P.}}:
\batitle{A {{Knowledge-Based Hierarchical Causal Inference Network}} for {{Video Action Recognition}}}.
\bjtitle{IEEE Transactions on Multimedia}
\bvolume{26},
\bfpage{9135}--\blpage{9149}
(\byear{2024})
\end{barticle}
\endbibitem

\bibitem{liu_knowledgedriven_2025}
\begin{barticle}
\bauthor{\bsnm{Liu}, \binits{Y.}},
\bauthor{\bsnm{Liu}, \binits{F.}},
\bauthor{\bsnm{Jiao}, \binits{L.}},
\bauthor{\bsnm{Bao}, \binits{Q.}},
\bauthor{\bsnm{Li}, \binits{S.}},
\bauthor{\bsnm{Li}, \binits{L.}},
\bauthor{\bsnm{Liu}, \binits{X.}}:
\batitle{Knowledge-{{Driven Compositional Action Recognition}}}.
\bjtitle{Pattern Recognition}
\bvolume{163},
\bfpage{111452}
(\byear{2025})
\end{barticle}
\endbibitem

\bibitem{jiao_braininspired_2025}
\begin{barticle}
\bauthor{\bsnm{Jiao}, \binits{L.}},
\bauthor{\bsnm{Ma}, \binits{M.}},
\bauthor{\bsnm{He}, \binits{P.}},
\bauthor{\bsnm{Geng}, \binits{X.}},
\bauthor{\bsnm{Liu}, \binits{X.}},
\bauthor{\bsnm{Liu}, \binits{F.}},
\bauthor{\bsnm{Ma}, \binits{W.}},
\bauthor{\bsnm{Yang}, \binits{S.}},
\bauthor{\bsnm{Hou}, \binits{B.}},
\bauthor{\bsnm{Tang}, \binits{X.}}:
\batitle{Brain-{{Inspired Learning}}, {{Perception}}, and {{Cognition}}: {{A Comprehensive Review}}}.
\bjtitle{IEEE Transactions on Neural Networks and Learning Systems}
\bvolume{36}(\bissue{4}),
\bfpage{5921}--\blpage{5941}
(\byear{2025})
\end{barticle}
\endbibitem

\bibitem{plizzari_what_2023}
\begin{bchapter}
\bauthor{\bsnm{Plizzari}, \binits{C.}},
\bauthor{\bsnm{Perrett}, \binits{T.}},
\bauthor{\bsnm{Caputo}, \binits{B.}},
\bauthor{\bsnm{Damen}, \binits{D.}}:
\bctitle{What can a cook in italy teach a mechanic in india? action recognition generalisation over scenarios and locations}.
In: \bbtitle{Proceedings of the IEEE/CVF International Conference on Computer Vision},
pp. \bfpage{13656}--\blpage{13666}
(\byear{2023})
\end{bchapter}
\endbibitem

\bibitem{kundu_discovering_2024}
\begin{bchapter}
\bauthor{\bsnm{Kundu}, \binits{S.}},
\bauthor{\bsnm{Trehan}, \binits{S.}},
\bauthor{\bsnm{Aakur}, \binits{S.N.}}:
\bctitle{Discovering novel actions from open world egocentric videos with object-grounded visual commonsense reasoning}.
In: \bbtitle{Proceedings of the European Conference on Computer Vision (ECCV)},
pp. \bfpage{39}--\blpage{56}
(\byear{2024}).
\bcomment{Springer Nature Switzerland}
\end{bchapter}
\endbibitem

\bibitem{hatano_multimodal_2024}
\begin{bchapter}
\bauthor{\bsnm{Hatano}, \binits{M.}},
\bauthor{\bsnm{Hachiuma}, \binits{R.}},
\bauthor{\bsnm{Fujii}, \binits{R.}},
\bauthor{\bsnm{Saito}, \binits{H.}}:
\bctitle{Multimodal cross-domain few-shot learning for egocentric action recognition}.
In: \bbtitle{European Conference on Computer Vision},
pp. \bfpage{182}--\blpage{199}
(\byear{2024}).
\bcomment{Springer}
\end{bchapter}
\endbibitem

\bibitem{huang_improving_2020}
\begin{bchapter}
\bauthor{\bsnm{Huang}, \binits{Y.}},
\bauthor{\bsnm{Sugano}, \binits{Y.}},
\bauthor{\bsnm{Sato}, \binits{Y.}}:
\bctitle{Improving action segmentation via graph-based temporal reasoning}.
In: \bbtitle{Proceedings of the IEEE/CVF Conference on Computer Vision and Pattern Recognition},
pp. \bfpage{14024}--\blpage{14034}
(\byear{2020})
\end{bchapter}
\endbibitem

\bibitem{lin_egocentric_2022}
\begin{barticle}
\bauthor{\bsnm{Lin}, \binits{K.Q.}},
\bauthor{\bsnm{Wang}, \binits{J.}},
\bauthor{\bsnm{Soldan}, \binits{M.}},
\bauthor{\bsnm{Wray}, \binits{M.}},
\bauthor{\bsnm{Yan}, \binits{R.}},
\bauthor{\bsnm{Xu}, \binits{E.Z.}},
\bauthor{\bsnm{Gao}, \binits{D.}},
\bauthor{\bsnm{Tu}, \binits{R.-C.}},
\bauthor{\bsnm{Zhao}, \binits{W.}},
\bauthor{\bsnm{Kong}, \binits{W.}}, \betal:
\batitle{Egocentric video-language pretraining}.
\bjtitle{Advances in Neural Information Processing Systems}
\bvolume{35},
\bfpage{7575}--\blpage{7586}
(\byear{2022})
\end{barticle}
\endbibitem

\bibitem{quattrocchi_synchronization_2024}
\begin{bchapter}
\bauthor{\bsnm{Quattrocchi}, \binits{C.}},
\bauthor{\bsnm{Furnari}, \binits{A.}},
\bauthor{\bsnm{Di~Mauro}, \binits{D.}},
\bauthor{\bsnm{Giuffrida}, \binits{M.V.}},
\bauthor{\bsnm{Farinella}, \binits{G.M.}}:
\bctitle{Synchronization is all you need: Exocentric-to-egocentric transfer for temporal action segmentation with unlabeled synchronized video pairs}.
In: \bbtitle{European Conference on Computer Vision},
pp. \bfpage{253}--\blpage{270}
(\byear{2024}).
\bcomment{Springer}
\end{bchapter}
\endbibitem

\bibitem{shih-polee_error_2024}
\begin{bchapter}
\bauthor{\bsnm{Lee}, \binits{S.-P.}},
\bauthor{\bsnm{Lu}, \binits{Z.}},
\bauthor{\bsnm{Zhang}, \binits{Z.}},
\bauthor{\bsnm{Hoai}, \binits{M.}},
\bauthor{\bsnm{Elhamifar}, \binits{E.}}:
\bctitle{Error detection in egocentric procedural task videos}.
In: \bbtitle{Proceedings of the IEEE/CVF Conference on Computer Vision and Pattern Recognition},
pp. \bfpage{18655}--\blpage{18666}
(\byear{2024})
\end{bchapter}
\endbibitem

\bibitem{reza_hat_2024}
\begin{bchapter}
\bauthor{\bsnm{Reza}, \binits{S.}},
\bauthor{\bsnm{Zhang}, \binits{Y.}},
\bauthor{\bsnm{Moghaddam}, \binits{M.}},
\bauthor{\bsnm{Camps}, \binits{O.}}:
\bctitle{Hat: History-augmented anchor transformer for online temporal action localization}.
In: \bbtitle{European Conference on Computer Vision},
pp. \bfpage{205}--\blpage{222}
(\byear{2024}).
\bcomment{Springer}
\end{bchapter}
\endbibitem

\bibitem{kilgour_frechet_2018}
\begin{botherref}
\oauthor{\bsnm{Kilgour}, \binits{K.}},
\oauthor{\bsnm{Zuluaga}, \binits{M.}},
\oauthor{\bsnm{Roblek}, \binits{D.}},
\oauthor{\bsnm{Sharifi}, \binits{M.}}:
Fr$\backslash$'echet audio distance: A metric for evaluating music enhancement algorithms.
arXiv preprint arXiv:1812.08466
(2018)
\end{botherref}
\endbibitem

\bibitem{wu_largescale_2023}
\begin{bchapter}
\bauthor{\bsnm{Wu}, \binits{Y.}},
\bauthor{\bsnm{Chen}, \binits{K.}},
\bauthor{\bsnm{Zhang}, \binits{T.}},
\bauthor{\bsnm{Hui}, \binits{Y.}},
\bauthor{\bsnm{Berg-Kirkpatrick}, \binits{T.}},
\bauthor{\bsnm{Dubnov}, \binits{S.}}:
\bctitle{Large-scale contrastive language-audio pretraining with feature fusion and keyword-to-caption augmentation}.
In: \bbtitle{ICASSP 2023-2023 IEEE International Conference on Acoustics, Speech and Signal Processing (ICASSP)},
pp. \bfpage{1}--\blpage{5}
(\byear{2023}).
\bcomment{IEEE}
\end{bchapter}
\endbibitem

\bibitem{luo_difffoley_2023}
\begin{botherref}
\oauthor{\bsnm{Luo}, \binits{S.}},
\oauthor{\bsnm{Yan}, \binits{C.}},
\oauthor{\bsnm{Hu}, \binits{C.}},
\oauthor{\bsnm{Zhao}, \binits{H.}}:
Diff-foley: Synchronized video-to-audio synthesis with latent diffusion models.
Advances in Neural Information Processing Systems
\textbf{36}
(2024)
\end{botherref}
\endbibitem

\bibitem{huh_epicsounds_2023}
\begin{bchapter}
\bauthor{\bsnm{Huh}, \binits{J.}},
\bauthor{\bsnm{Chalk}, \binits{J.}},
\bauthor{\bsnm{Kazakos}, \binits{E.}},
\bauthor{\bsnm{Damen}, \binits{D.}},
\bauthor{\bsnm{Zisserman}, \binits{A.}}:
\bctitle{Epic-sounds: A large-scale dataset of actions that sound}.
In: \bbtitle{ICASSP 2023-2023 IEEE International Conference on Acoustics, Speech and Signal Processing (ICASSP)},
pp. \bfpage{1}--\blpage{5}
(\byear{2023}).
\bcomment{IEEE}
\end{bchapter}
\endbibitem

\bibitem{oncescu_sound_2024}
\begin{bchapter}
\bauthor{\bsnm{Oncescu}, \binits{A.-M.}},
\bauthor{\bsnm{Henriques}, \binits{J.F.}},
\bauthor{\bsnm{Zisserman}, \binits{A.}},
\bauthor{\bsnm{Albanie}, \binits{S.}},
\bauthor{\bsnm{Koepke}, \binits{A.S.}}:
\bctitle{A sound approach: Using large language models to generate audio descriptions for egocentric text-audio retrieval}.
In: \bbtitle{ICASSP 2024-2024 IEEE International Conference on Acoustics, Speech and Signal Processing (ICASSP)},
pp. \bfpage{7300}--\blpage{7304}
(\byear{2024}).
\bcomment{IEEE}
\end{bchapter}
\endbibitem

\bibitem{chen_action2sound_2024}
\begin{bchapter}
\bauthor{\bsnm{Chen}, \binits{C.}},
\bauthor{\bsnm{Peng}, \binits{P.}},
\bauthor{\bsnm{Baid}, \binits{A.}},
\bauthor{\bsnm{Xue}, \binits{S.}},
\bauthor{\bsnm{Hsu}, \binits{W.-N.}},
\bauthor{\bsnm{Harwath}, \binits{D.}},
\bauthor{\bsnm{Grauman}, \binits{K.}}:
\bctitle{Action2sound: Ambient-aware generation of action sounds from egocentric videos}.
In: \bbtitle{Proceedings of the European Conference on Computer Vision (ECCV)}
(\byear{2024})
\end{bchapter}
\endbibitem

\bibitem{chen_soundingactions_2024}
\begin{bchapter}
\bauthor{\bsnm{Chen}, \binits{C.}},
\bauthor{\bsnm{Ashutosh}, \binits{K.}},
\bauthor{\bsnm{Girdhar}, \binits{R.}},
\bauthor{\bsnm{Harwath}, \binits{D.}},
\bauthor{\bsnm{Grauman}, \binits{K.}}:
\bctitle{Soundingactions: Learning how actions sound from narrated egocentric videos}.
In: \bbtitle{Proceedings of the IEEE/CVF Conference on Computer Vision and Pattern Recognition},
pp. \bfpage{27252}--\blpage{27262}
(\byear{2024})
\end{bchapter}
\endbibitem

\bibitem{liu_forecasting_2020}
\begin{bchapter}
\bauthor{\bsnm{Liu}, \binits{M.}},
\bauthor{\bsnm{Tang}, \binits{S.}},
\bauthor{\bsnm{Li}, \binits{Y.}},
\bauthor{\bsnm{Rehg}, \binits{J.M.}}:
\bctitle{Forecasting human-object interaction: joint prediction of motor attention and actions in first person video}.
In: \bbtitle{Computer Vision--ECCV 2020: 16th European Conference, Glasgow, UK, August 23--28, 2020, Proceedings, Part I 16},
pp. \bfpage{704}--\blpage{721}
(\byear{2020}).
\bcomment{Springer}
\end{bchapter}
\endbibitem

\bibitem{jia_generative_2022}
\begin{bchapter}
\bauthor{\bsnm{Jia}, \binits{W.}},
\bauthor{\bsnm{Liu}, \binits{M.}},
\bauthor{\bsnm{Rehg}, \binits{J.M.}}:
\bctitle{Generative adversarial network for future hand segmentation from egocentric video}.
In: \bbtitle{European Conference on Computer Vision},
pp. \bfpage{639}--\blpage{656}
(\byear{2022}).
\bcomment{Springer}
\end{bchapter}
\endbibitem

\bibitem{bao_uncertaintyaware_2023}
\begin{bchapter}
\bauthor{\bsnm{Bao}, \binits{W.}},
\bauthor{\bsnm{Chen}, \binits{L.}},
\bauthor{\bsnm{Zeng}, \binits{L.}},
\bauthor{\bsnm{Li}, \binits{Z.}},
\bauthor{\bsnm{Xu}, \binits{Y.}},
\bauthor{\bsnm{Yuan}, \binits{J.}},
\bauthor{\bsnm{Kong}, \binits{Y.}}:
\bctitle{Uncertainty-aware state space transformer for egocentric 3d hand trajectory forecasting}.
In: \bbtitle{Proceedings of the IEEE/CVF International Conference on Computer Vision},
pp. \bfpage{13702}--\blpage{13711}
(\byear{2023})
\end{bchapter}
\endbibitem

\bibitem{abilkassov_augmenting_2024}
\begin{barticle}
\bauthor{\bsnm{Abilkassov}, \binits{S.}},
\bauthor{\bsnm{Gentner}, \binits{M.}},
\bauthor{\bsnm{Popa}, \binits{M.}}:
\batitle{Augmenting human-robot collaboration task by human hand position forecasting}.
\bjtitle{Proceedings Copyright}
\bvolume{262},
\bfpage{269}
(\byear{2024})
\end{barticle}
\endbibitem

\bibitem{li_uniformer_2023}
\begin{barticle}
\bauthor{\bsnm{Li}, \binits{K.}},
\bauthor{\bsnm{Wang}, \binits{Y.}},
\bauthor{\bsnm{Zhang}, \binits{J.}},
\bauthor{\bsnm{Gao}, \binits{P.}},
\bauthor{\bsnm{Song}, \binits{G.}},
\bauthor{\bsnm{Liu}, \binits{Y.}},
\bauthor{\bsnm{Li}, \binits{H.}},
\bauthor{\bsnm{Qiao}, \binits{Y.}}:
\batitle{Uniformer: Unifying convolution and self-attention for visual recognition}.
\bjtitle{IEEE Transactions on Pattern Analysis and Machine Intelligence}
\bvolume{45}(\bissue{10}),
\bfpage{12581}--\blpage{12600}
(\byear{2023})
\end{barticle}
\endbibitem

\bibitem{tang_prompting_2024}
\begin{bchapter}
\bauthor{\bsnm{Tang}, \binits{B.}},
\bauthor{\bsnm{Zhang}, \binits{K.}},
\bauthor{\bsnm{Luo}, \binits{W.}},
\bauthor{\bsnm{Liu}, \binits{W.}},
\bauthor{\bsnm{Li}, \binits{H.}}:
\bctitle{Prompting future driven diffusion model for hand motion prediction}.
In: \bbtitle{European Conference on Computer Vision},
pp. \bfpage{169}--\blpage{186}
(\byear{2024}).
\bcomment{Springer}
\end{bchapter}
\endbibitem

\bibitem{garcia-hernando_firstperson_2018}
\begin{bchapter}
\bauthor{\bsnm{Garcia-Hernando}, \binits{G.}},
\bauthor{\bsnm{Yuan}, \binits{S.}},
\bauthor{\bsnm{Baek}, \binits{S.}},
\bauthor{\bsnm{Kim}, \binits{T.-K.}}:
\bctitle{First-person hand action benchmark with rgb-d videos and 3d hand pose annotations}.
In: \bbtitle{Proceedings of the IEEE Conference on Computer Vision and Pattern Recognition},
pp. \bfpage{409}--\blpage{419}
(\byear{2018})
\end{bchapter}
\endbibitem

\bibitem{wu_learning_2021}
\begin{barticle}
\bauthor{\bsnm{Wu}, \binits{Y.}},
\bauthor{\bsnm{Zhu}, \binits{L.}},
\bauthor{\bsnm{Wang}, \binits{X.}},
\bauthor{\bsnm{Yang}, \binits{Y.}},
\bauthor{\bsnm{Wu}, \binits{F.}}:
\batitle{Learning to anticipate egocentric actions by imagination}.
\bjtitle{IEEE Transactions on Image Processing}
\bvolume{30},
\bfpage{1143}--\blpage{1152}
(\byear{2020})
\end{barticle}
\endbibitem

\bibitem{roy_action_2022}
\begin{bchapter}
\bauthor{\bsnm{Roy}, \binits{D.}},
\bauthor{\bsnm{Fernando}, \binits{B.}}:
\bctitle{Action anticipation using latent goal learning}.
In: \bbtitle{Proceedings of the IEEE/CVF Winter Conference on Applications of Computer Vision},
pp. \bfpage{2745}--\blpage{2753}
(\byear{2022})
\end{bchapter}
\endbibitem

\bibitem{nawhal_rethinking_2022}
\begin{bchapter}
\bauthor{\bsnm{Nawhal}, \binits{M.}},
\bauthor{\bsnm{Jyothi}, \binits{A.A.}},
\bauthor{\bsnm{Mori}, \binits{G.}}:
\bctitle{Rethinking learning approaches for long-term action anticipation}.
In: \bbtitle{European Conference on Computer Vision},
pp. \bfpage{558}--\blpage{576}
(\byear{2022}).
\bcomment{Springer}
\end{bchapter}
\endbibitem

\bibitem{girdhar_anticipative_2021}
\begin{bchapter}
\bauthor{\bsnm{Girdhar}, \binits{R.}},
\bauthor{\bsnm{Grauman}, \binits{K.}}:
\bctitle{Anticipative video transformer}.
In: \bbtitle{Proceedings of the IEEE/CVF International Conference on Computer Vision},
pp. \bfpage{13505}--\blpage{13515}
(\byear{2021})
\end{bchapter}
\endbibitem

\bibitem{ashutosh_hiervl_2023}
\begin{bchapter}
\bauthor{\bsnm{Ashutosh}, \binits{K.}},
\bauthor{\bsnm{Girdhar}, \binits{R.}},
\bauthor{\bsnm{Torresani}, \binits{L.}},
\bauthor{\bsnm{Grauman}, \binits{K.}}:
\bctitle{Hiervl: Learning hierarchical video-language embeddings}.
In: \bbtitle{Proceedings of the IEEE/CVF Conference on Computer Vision and Pattern Recognition},
pp. \bfpage{23066}--\blpage{23078}
(\byear{2023})
\end{bchapter}
\endbibitem

\bibitem{zhao_diverse_2020}
\begin{bchapter}
\bauthor{\bsnm{Zhao}, \binits{H.}},
\bauthor{\bsnm{Wildes}, \binits{R.P.}}:
\bctitle{On diverse asynchronous activity anticipation}.
In: \bbtitle{Computer Vision--ECCV 2020: 16th European Conference, Glasgow, UK, August 23--28, 2020, Proceedings, Part XXIX 16},
pp. \bfpage{781}--\blpage{799}
(\byear{2020}).
\bcomment{Springer}
\end{bchapter}
\endbibitem

\bibitem{abdelslam_gepsan_2023}
\begin{bchapter}
\bauthor{\bsnm{Abdelsalam}, \binits{M.A.}},
\bauthor{\bsnm{Rangrej}, \binits{S.B.}},
\bauthor{\bsnm{Hadji}, \binits{I.}},
\bauthor{\bsnm{Dvornik}, \binits{N.}},
\bauthor{\bsnm{Derpanis}, \binits{K.G.}},
\bauthor{\bsnm{Fazly}, \binits{A.}}:
\bctitle{Gepsan: Generative procedure step anticipation in cooking videos}.
In: \bbtitle{Proceedings of the IEEE/CVF International Conference on Computer Vision},
pp. \bfpage{2988}--\blpage{2997}
(\byear{2023})
\end{bchapter}
\endbibitem

\bibitem{ho_denoising_2020}
\begin{barticle}
\bauthor{\bsnm{Ho}, \binits{J.}},
\bauthor{\bsnm{Jain}, \binits{A.}},
\bauthor{\bsnm{Abbeel}, \binits{P.}}:
\batitle{Denoising diffusion probabilistic models}.
\bjtitle{Advances in neural information processing systems}
\bvolume{33},
\bfpage{6840}--\blpage{6851}
(\byear{2020})
\end{barticle}
\endbibitem

\bibitem{lai_lego_2024}
\begin{bchapter}
\bauthor{\bsnm{Lai}, \binits{B.}},
\bauthor{\bsnm{Dai}, \binits{X.}},
\bauthor{\bsnm{Chen}, \binits{L.}},
\bauthor{\bsnm{Pang}, \binits{G.}},
\bauthor{\bsnm{Rehg}, \binits{J.M.}},
\bauthor{\bsnm{Liu}, \binits{M.}}:
\bctitle{Lego: Learning egocentric action frame generation via visual instruction tuning}.
In: \bbtitle{European Conference on Computer Vision},
pp. \bfpage{135}--\blpage{155}
(\byear{2024}).
\bcomment{Springer}
\end{bchapter}
\endbibitem

\bibitem{rombach_highresolution_2022}
\begin{bchapter}
\bauthor{\bsnm{Rombach}, \binits{R.}},
\bauthor{\bsnm{Blattmann}, \binits{A.}},
\bauthor{\bsnm{Lorenz}, \binits{D.}},
\bauthor{\bsnm{Esser}, \binits{P.}},
\bauthor{\bsnm{Ommer}, \binits{B.}}:
\bctitle{High-resolution image synthesis with latent diffusion models}.
In: \bbtitle{Proceedings of the IEEE/CVF Conference on Computer Vision and Pattern Recognition},
pp. \bfpage{10684}--\blpage{10695}
(\byear{2022})
\end{bchapter}
\endbibitem

\bibitem{donley_easycom_2021}
\begin{botherref}
\oauthor{\bsnm{Donley}, \binits{J.}},
\oauthor{\bsnm{Tourbabin}, \binits{V.}},
\oauthor{\bsnm{Lee}, \binits{J.-S.}},
\oauthor{\bsnm{Broyles}, \binits{M.}},
\oauthor{\bsnm{Jiang}, \binits{H.}},
\oauthor{\bsnm{Shen}, \binits{J.}},
\oauthor{\bsnm{Pantic}, \binits{M.}},
\oauthor{\bsnm{Ithapu}, \binits{V.K.}},
\oauthor{\bsnm{Mehra}, \binits{R.}}:
Easycom: An augmented reality dataset to support algorithms for easy communication in noisy environments.
arXiv preprint arXiv:2107.04174
(2021)
\end{botherref}
\endbibitem

\bibitem{tran_ex2egmae_2024}
\begin{bchapter}
\bauthor{\bsnm{Tran}, \binits{M.}},
\bauthor{\bsnm{Kim}, \binits{Y.}},
\bauthor{\bsnm{Su}, \binits{C.-C.}},
\bauthor{\bsnm{Kuo}, \binits{C.-H.}},
\bauthor{\bsnm{Sun}, \binits{M.}},
\bauthor{\bsnm{Soleymani}, \binits{M.}}:
\bctitle{Ex2eg-mae: A framework for adaptation of exocentric video masked autoencoders for egocentric social role understanding}.
In: \bbtitle{European Conference on Computer Vision},
pp. \bfpage{1}--\blpage{19}
(\byear{2024}).
\bcomment{Springer}
\end{bchapter}
\endbibitem

\bibitem{kong_longterm_2024}
\begin{bchapter}
\bauthor{\bsnm{Kong}, \binits{D.}},
\bauthor{\bsnm{Khan}, \binits{F.}},
\bauthor{\bsnm{Zhang}, \binits{X.}},
\bauthor{\bsnm{Singhal}, \binits{P.}},
\bauthor{\bsnm{Wu}, \binits{Y.N.}}:
\bctitle{Long-term social interaction context: The key to egocentric addressee detection}.
In: \bbtitle{ICASSP 2024-2024 IEEE International Conference on Acoustics, Speech and Signal Processing (ICASSP)},
pp. \bfpage{8250}--\blpage{8254}
(\byear{2024}).
\bcomment{IEEE}
\end{bchapter}
\endbibitem

\bibitem{northcutt_egocom_2023}
\begin{botherref}
\oauthor{\bsnm{Northcutt}, \binits{C.}},
\oauthor{\bsnm{Zha}, \binits{S.}},
\oauthor{\bsnm{Lovegrove}, \binits{S.}},
\oauthor{\bsnm{Newcombe}, \binits{R.}}:
Egocom: A multi-person multi-modal egocentric communications dataset.
IEEE Transactions on Pattern Analysis and Machine Intelligence
(2020)
\end{botherref}
\endbibitem

\bibitem{majumder_learning_2024}
\begin{bchapter}
\bauthor{\bsnm{Majumder}, \binits{S.}},
\bauthor{\bsnm{Al-Halah}, \binits{Z.}},
\bauthor{\bsnm{Grauman}, \binits{K.}}:
\bctitle{Learning spatial features from audio-visual correspondence in egocentric videos}.
In: \bbtitle{Proceedings of the IEEE/CVF Conference on Computer Vision and Pattern Recognition},
pp. \bfpage{27058}--\blpage{27068}
(\byear{2024})
\end{bchapter}
\endbibitem

\bibitem{lertniphonphan_pcie_lam_2024}
\begin{botherref}
\oauthor{\bsnm{Lertniphonphan}, \binits{K.}},
\oauthor{\bsnm{Xie}, \binits{J.}},
\oauthor{\bsnm{Meng}, \binits{Y.}},
\oauthor{\bsnm{Wang}, \binits{S.}},
\oauthor{\bsnm{Chen}, \binits{F.}},
\oauthor{\bsnm{Wang}, \binits{Z.}}:
Pcie\_lam solution for ego4d looking at me challenge.
arXiv preprint arXiv:2406.12211
(2024)
\end{botherref}
\endbibitem

\bibitem{duarte_action_2018}
\begin{barticle}
\bauthor{\bsnm{Duarte}, \binits{N.F.}},
\bauthor{\bsnm{Rakovi{\'c}}, \binits{M.}},
\bauthor{\bsnm{Tasevski}, \binits{J.}},
\bauthor{\bsnm{Coco}, \binits{M.I.}},
\bauthor{\bsnm{Billard}, \binits{A.}},
\bauthor{\bsnm{Santos-Victor}, \binits{J.}}:
\batitle{Action anticipation: Reading the intentions of humans and robots}.
\bjtitle{IEEE Robotics and Automation Letters}
\bvolume{3}(\bissue{4}),
\bfpage{4132}--\blpage{4139}
(\byear{2018})
\end{barticle}
\endbibitem

\bibitem{li_deep_2019}
\begin{bchapter}
\bauthor{\bsnm{Li}, \binits{H.}},
\bauthor{\bsnm{Cai}, \binits{Y.}},
\bauthor{\bsnm{Zheng}, \binits{W.-S.}}:
\bctitle{Deep dual relation modeling for egocentric interaction recognition}.
In: \bbtitle{Proceedings of the IEEE/CVF Conference on Computer Vision and Pattern Recognition},
pp. \bfpage{7932}--\blpage{7941}
(\byear{2019})
\end{bchapter}
\endbibitem

\bibitem{lai_werewolf_2023}
\begin{botherref}
\oauthor{\bsnm{Lai}, \binits{B.}},
\oauthor{\bsnm{Zhang}, \binits{H.}},
\oauthor{\bsnm{Liu}, \binits{M.}},
\oauthor{\bsnm{Pariani}, \binits{A.}},
\oauthor{\bsnm{Ryan}, \binits{F.}},
\oauthor{\bsnm{Jia}, \binits{W.}},
\oauthor{\bsnm{Hayati}, \binits{S.A.}},
\oauthor{\bsnm{Rehg}, \binits{J.}},
\oauthor{\bsnm{Yang}, \binits{D.}}:
Werewolf among us: Multimodal resources for modeling persuasion behaviors in social deduction games.
Association for Computational Linguistics: ACL 2023
(2023)
\end{botherref}
\endbibitem

\bibitem{grimaldi_am_2024}
\begin{bchapter}
\bauthor{\bsnm{Grimaldi}, \binits{C.}},
\bauthor{\bsnm{Rossi}, \binits{A.}},
\bauthor{\bsnm{Rossi}, \binits{S.}}:
\bctitle{I am part of the robot’s group: Evaluating engagement and group membership from egocentric views}.
In: \bbtitle{2024 33rd IEEE International Conference on Robot and Human Interactive Communication (ROMAN)},
pp. \bfpage{1774}--\blpage{1779}
(\byear{2024}).
\bcomment{IEEE}
\end{bchapter}
\endbibitem

\bibitem{lu_sound_2022}
\begin{barticle}
\bauthor{\bsnm{Lu}, \binits{H.}},
\bauthor{\bsnm{Brimijoin}, \binits{W.O.}}:
\batitle{Sound source selection based on head movements in natural group conversation}.
\bjtitle{Trends in Hearing}
\bvolume{26},
\bfpage{23312165221097789}
(\byear{2022})
\end{barticle}
\endbibitem

\bibitem{yin_hearing_2024}
\begin{bchapter}
\bauthor{\bsnm{Yin}, \binits{Y.}},
\bauthor{\bsnm{Ananthabhotla}, \binits{I.}},
\bauthor{\bsnm{Ithapu}, \binits{V.K.}},
\bauthor{\bsnm{Petridis}, \binits{S.}},
\bauthor{\bsnm{Wu}, \binits{Y.-H.}},
\bauthor{\bsnm{Miller}, \binits{C.}}:
\bctitle{Hearing loss detection from facial expressions in one-on-one conversations}.
In: \bbtitle{ICASSP 2024-2024 IEEE International Conference on Acoustics, Speech and Signal Processing (ICASSP)},
pp. \bfpage{5460}--\blpage{5464}
(\byear{2024}).
\bcomment{IEEE}
\end{bchapter}
\endbibitem

\bibitem{jiang_egocentric_2022}
\begin{bchapter}
\bauthor{\bsnm{Jiang}, \binits{H.}},
\bauthor{\bsnm{Murdock}, \binits{C.}},
\bauthor{\bsnm{Ithapu}, \binits{V.K.}}:
\bctitle{Egocentric deep multi-channel audio-visual active speaker localization}.
In: \bbtitle{Proceedings of the IEEE/CVF Conference on Computer Vision and Pattern Recognition},
pp. \bfpage{10544}--\blpage{10552}
(\byear{2022})
\end{bchapter}
\endbibitem

\bibitem{chong_detection_2020}
\begin{barticle}
\bauthor{\bsnm{Chong}, \binits{E.}},
\bauthor{\bsnm{Clark-Whitney}, \binits{E.}},
\bauthor{\bsnm{Southerland}, \binits{A.}},
\bauthor{\bsnm{Stubbs}, \binits{E.}},
\bauthor{\bsnm{Miller}, \binits{C.}},
\bauthor{\bsnm{Ajodan}, \binits{E.L.}},
\bauthor{\bsnm{Silverman}, \binits{M.R.}},
\bauthor{\bsnm{Lord}, \binits{C.}},
\bauthor{\bsnm{Rozga}, \binits{A.}},
\bauthor{\bsnm{Jones}, \binits{R.M.}}, \betal:
\batitle{Detection of eye contact with deep neural networks is as accurate as human experts}.
\bjtitle{Nature communications}
\bvolume{11}(\bissue{1}),
\bfpage{6386}
(\byear{2020})
\end{barticle}
\endbibitem

\bibitem{xue_egocentric_2023}
\begin{bchapter}
\bauthor{\bsnm{Xue}, \binits{Z.}},
\bauthor{\bsnm{Song}, \binits{Y.}},
\bauthor{\bsnm{Grauman}, \binits{K.}},
\bauthor{\bsnm{Torresani}, \binits{L.}}:
\bctitle{Egocentric video task translation}.
In: \bbtitle{Proceedings of the IEEE/CVF Conference on Computer Vision and Pattern Recognition},
pp. \bfpage{2310}--\blpage{2320}
(\byear{2023})
\end{bchapter}
\endbibitem

\bibitem{choudhary_domain_2021}
\begin{bchapter}
\bauthor{\bsnm{Choudhary}, \binits{A.}},
\bauthor{\bsnm{Mishra}, \binits{D.}},
\bauthor{\bsnm{Karmakar}, \binits{A.}}:
\bctitle{Domain adaptive egocentric person re-identification}.
In: \bbtitle{Computer Vision and Image Processing: 5th International Conference, CVIP 2020, Prayagraj, India, December 4-6, 2020, Revised Selected Papers, Part III 5},
pp. \bfpage{81}--\blpage{92}
(\byear{2021}).
\bcomment{Springer}
\end{bchapter}
\endbibitem

\bibitem{yagi_future_2018}
\begin{bchapter}
\bauthor{\bsnm{Yagi}, \binits{T.}},
\bauthor{\bsnm{Mangalam}, \binits{K.}},
\bauthor{\bsnm{Yonetani}, \binits{R.}},
\bauthor{\bsnm{Sato}, \binits{Y.}}:
\bctitle{Future person localization in first-person videos}.
In: \bbtitle{Proceedings of the IEEE Conference on Computer Vision and Pattern Recognition},
pp. \bfpage{7593}--\blpage{7602}
(\byear{2018})
\end{bchapter}
\endbibitem

\bibitem{chen_future_2023}
\begin{barticle}
\bauthor{\bsnm{Chen}, \binits{K.}},
\bauthor{\bsnm{Zhu}, \binits{H.}},
\bauthor{\bsnm{Tang}, \binits{D.}},
\bauthor{\bsnm{Zheng}, \binits{K.}}:
\batitle{Future pedestrian location prediction in first-person videos for autonomous vehicles and social robots}.
\bjtitle{Image and Vision Computing}
\bvolume{134},
\bfpage{104671}
(\byear{2023})
\end{barticle}
\endbibitem

\bibitem{zhu_egoobjects_2023}
\begin{bchapter}
\bauthor{\bsnm{Zhu}, \binits{C.}},
\bauthor{\bsnm{Xiao}, \binits{F.}},
\bauthor{\bsnm{Alvarado}, \binits{A.}},
\bauthor{\bsnm{Babaei}, \binits{Y.}},
\bauthor{\bsnm{Hu}, \binits{J.}},
\bauthor{\bsnm{El-Mohri}, \binits{H.}},
\bauthor{\bsnm{Culatana}, \binits{S.}},
\bauthor{\bsnm{Sumbaly}, \binits{R.}},
\bauthor{\bsnm{Yan}, \binits{Z.}}:
\bctitle{Egoobjects: A large-scale egocentric dataset for fine-grained object understanding}.
In: \bbtitle{Proceedings of the IEEE/CVF International Conference on Computer Vision},
pp. \bfpage{20110}--\blpage{20120}
(\byear{2023})
\end{bchapter}
\endbibitem

\bibitem{akiva_selfsupervised_2023}
\begin{bchapter}
\bauthor{\bsnm{Akiva}, \binits{P.}},
\bauthor{\bsnm{Huang}, \binits{J.}},
\bauthor{\bsnm{Liang}, \binits{K.J.}},
\bauthor{\bsnm{Kovvuri}, \binits{R.}},
\bauthor{\bsnm{Chen}, \binits{X.}},
\bauthor{\bsnm{Feiszli}, \binits{M.}},
\bauthor{\bsnm{Dana}, \binits{K.}},
\bauthor{\bsnm{Hassner}, \binits{T.}}:
\bctitle{Self-supervised object detection from egocentric videos}.
In: \bbtitle{Proceedings of the IEEE/CVF International Conference on Computer Vision},
pp. \bfpage{5225}--\blpage{5237}
(\byear{2023})
\end{bchapter}
\endbibitem

\bibitem{wu_labelefficient_2023}
\begin{bchapter}
\bauthor{\bsnm{Wu}, \binits{J.Z.}},
\bauthor{\bsnm{Zhang}, \binits{D.J.}},
\bauthor{\bsnm{Hsu}, \binits{W.}},
\bauthor{\bsnm{Zhang}, \binits{M.}},
\bauthor{\bsnm{Shou}, \binits{M.Z.}}:
\bctitle{Label-efficient online continual object detection in streaming video}.
In: \bbtitle{Proceedings of the IEEE/CVF International Conference on Computer Vision},
pp. \bfpage{19246}--\blpage{19255}
(\byear{2023})
\end{bchapter}
\endbibitem

\bibitem{darkhalil_epickitchens_2022}
\begin{barticle}
\bauthor{\bsnm{Darkhalil}, \binits{A.}},
\bauthor{\bsnm{Shan}, \binits{D.}},
\bauthor{\bsnm{Zhu}, \binits{B.}},
\bauthor{\bsnm{Ma}, \binits{J.}},
\bauthor{\bsnm{Kar}, \binits{A.}},
\bauthor{\bsnm{Higgins}, \binits{R.}},
\bauthor{\bsnm{Fidler}, \binits{S.}},
\bauthor{\bsnm{Fouhey}, \binits{D.}},
\bauthor{\bsnm{Damen}, \binits{D.}}:
\batitle{Epic-kitchens visor benchmark: Video segmentations and object relations}.
\bjtitle{Advances in Neural Information Processing Systems}
\bvolume{35},
\bfpage{13745}--\blpage{13758}
(\byear{2022})
\end{barticle}
\endbibitem

\bibitem{tokmakov_breaking_2023}
\begin{bchapter}
\bauthor{\bsnm{Tokmakov}, \binits{P.}},
\bauthor{\bsnm{Li}, \binits{J.}},
\bauthor{\bsnm{Gaidon}, \binits{A.}}:
\bctitle{Breaking the" object" in video object segmentation}.
In: \bbtitle{Proceedings of the IEEE/CVF Conference on Computer Vision and Pattern Recognition},
pp. \bfpage{22836}--\blpage{22845}
(\byear{2023})
\end{bchapter}
\endbibitem

\bibitem{jiang_singlestage_2023}
\begin{barticle}
\bauthor{\bsnm{Jiang}, \binits{H.}},
\bauthor{\bsnm{Ramakrishnan}, \binits{S.K.}},
\bauthor{\bsnm{Grauman}, \binits{K.}}:
\batitle{Single-stage visual query localization in egocentric videos}.
\bjtitle{Advances in Neural Information Processing Systems}
\bvolume{36},
\bfpage{24143}--\blpage{24157}
(\byear{2023})
\end{barticle}
\endbibitem

\bibitem{xu_where_2023}
\begin{bchapter}
\bauthor{\bsnm{Xu}, \binits{M.}},
\bauthor{\bsnm{Li}, \binits{Y.}},
\bauthor{\bsnm{Fu}, \binits{C.-Y.}},
\bauthor{\bsnm{Ghanem}, \binits{B.}},
\bauthor{\bsnm{Xiang}, \binits{T.}},
\bauthor{\bsnm{P{\'e}rez-R{\'u}a}, \binits{J.-M.}}:
\bctitle{Where is my wallet? modeling object proposal sets for egocentric visual query localization}.
In: \bbtitle{Proceedings of the IEEE/CVF Conference on Computer Vision and Pattern Recognition},
pp. \bfpage{2593}--\blpage{2603}
(\byear{2023})
\end{bchapter}
\endbibitem

\bibitem{khosla_relocate_2024}
\begin{botherref}
\oauthor{\bsnm{Khosla}, \binits{S.}},
\oauthor{\bsnm{TV}, \binits{S.}},
\oauthor{\bsnm{Schwing}, \binits{A.}},
\oauthor{\bsnm{Hoiem}, \binits{D.}}:
Relocate: A simple training-free baseline for visual query localization using region-based representations.
arXiv preprint arXiv:2412.01826
(2024)
\end{botherref}
\endbibitem

\bibitem{mai_egoloc_2023}
\begin{bchapter}
\bauthor{\bsnm{Mai}, \binits{J.}},
\bauthor{\bsnm{Hamdi}, \binits{A.}},
\bauthor{\bsnm{Giancola}, \binits{S.}},
\bauthor{\bsnm{Zhao}, \binits{C.}},
\bauthor{\bsnm{Ghanem}, \binits{B.}}:
\bctitle{Egoloc: Revisiting 3d object localization from egocentric videos with visual queries}.
In: \bbtitle{Proceedings of the IEEE/CVF International Conference on Computer Vision},
pp. \bfpage{45}--\blpage{57}
(\byear{2023})
\end{bchapter}
\endbibitem

\bibitem{huang_egocentric_2023}
\begin{bchapter}
\bauthor{\bsnm{Huang}, \binits{C.}},
\bauthor{\bsnm{Tian}, \binits{Y.}},
\bauthor{\bsnm{Kumar}, \binits{A.}},
\bauthor{\bsnm{Xu}, \binits{C.}}:
\bctitle{Egocentric audio-visual object localization}.
In: \bbtitle{Proceedings of the IEEE/CVF Conference on Computer Vision and Pattern Recognition},
pp. \bfpage{22910}--\blpage{22921}
(\byear{2023})
\end{bchapter}
\endbibitem

\bibitem{shi_crossmodal_2025}
\begin{barticle}
\bauthor{\bsnm{Shi}, \binits{Z.}},
\bauthor{\bsnm{Wu}, \binits{Q.}},
\bauthor{\bsnm{Meng}, \binits{F.}},
\bauthor{\bsnm{Xu}, \binits{L.}},
\bauthor{\bsnm{Li}, \binits{H.}}:
\batitle{Cross-modal cognitive consensus guided audio–visual segmentation}.
\bjtitle{IEEE Transactions on Multimedia}
\bvolume{27},
\bfpage{209}--\blpage{223}
(\byear{2025}).
\doiurl{10.1109/TMM.2024.3521746}
\end{barticle}
\endbibitem

\bibitem{zhang_helping_2023}
\begin{bchapter}
\bauthor{\bsnm{Zhang}, \binits{C.}},
\bauthor{\bsnm{Gupta}, \binits{A.}},
\bauthor{\bsnm{Zisserman}, \binits{A.}}:
\bctitle{Helping hands: An object-aware ego-centric video recognition model}.
In: \bbtitle{Proceedings of the IEEE/CVF International Conference on Computer Vision},
pp. \bfpage{13901}--\blpage{13912}
(\byear{2023})
\end{bchapter}
\endbibitem

\bibitem{yu_video_2023}
\begin{bchapter}
\bauthor{\bsnm{Yu}, \binits{J.}},
\bauthor{\bsnm{Li}, \binits{X.}},
\bauthor{\bsnm{Zhao}, \binits{X.}},
\bauthor{\bsnm{Zhang}, \binits{H.}},
\bauthor{\bsnm{Wang}, \binits{Y.-X.}}:
\bctitle{Video state-changing object segmentation}.
In: \bbtitle{Proceedings of the IEEE/CVF International Conference on Computer Vision},
pp. \bfpage{20439}--\blpage{20448}
(\byear{2023})
\end{bchapter}
\endbibitem

\bibitem{tschernezki_neuraldiff_2021}
\begin{bchapter}
\bauthor{\bsnm{Tschernezki}, \binits{V.}},
\bauthor{\bsnm{Larlus}, \binits{D.}},
\bauthor{\bsnm{Vedaldi}, \binits{A.}}:
\bctitle{Neuraldiff: Segmenting 3d objects that move in egocentric videos}.
In: \bbtitle{2021 International Conference on 3D Vision (3DV)},
pp. \bfpage{910}--\blpage{919}
(\byear{2021}).
\bcomment{IEEE}
\end{bchapter}
\endbibitem

\bibitem{huang_tracking_2023}
\begin{bchapter}
\bauthor{\bsnm{Huang}, \binits{M.}},
\bauthor{\bsnm{Li}, \binits{X.}},
\bauthor{\bsnm{Hu}, \binits{J.}},
\bauthor{\bsnm{Peng}, \binits{H.}},
\bauthor{\bsnm{Lyu}, \binits{S.}}:
\bctitle{Tracking multiple deformable objects in egocentric videos}.
In: \bbtitle{Proceedings of the IEEE/CVF Conference on Computer Vision and Pattern Recognition},
pp. \bfpage{1461}--\blpage{1471}
(\byear{2023})
\end{bchapter}
\endbibitem

\bibitem{gu_egolifter_2024}
\begin{bchapter}
\bauthor{\bsnm{Gu}, \binits{Q.}},
\bauthor{\bsnm{Lv}, \binits{Z.}},
\bauthor{\bsnm{Frost}, \binits{D.}},
\bauthor{\bsnm{Green}, \binits{S.}},
\bauthor{\bsnm{Straub}, \binits{J.}},
\bauthor{\bsnm{Sweeney}, \binits{C.}}:
\bctitle{Egolifter: Open-world 3d segmentation for egocentric perception}.
In: \bbtitle{European Conference on Computer Vision},
pp. \bfpage{382}--\blpage{400}
(\byear{2024}).
\bcomment{Springer}
\end{bchapter}
\endbibitem

\bibitem{kirillov_segment_2023}
\begin{bchapter}
\bauthor{\bsnm{Kirillov}, \binits{A.}},
\bauthor{\bsnm{Mintun}, \binits{E.}},
\bauthor{\bsnm{Ravi}, \binits{N.}},
\bauthor{\bsnm{Mao}, \binits{H.}},
\bauthor{\bsnm{Rolland}, \binits{C.}},
\bauthor{\bsnm{Gustafson}, \binits{L.}},
\bauthor{\bsnm{Xiao}, \binits{T.}},
\bauthor{\bsnm{Whitehead}, \binits{S.}},
\bauthor{\bsnm{Berg}, \binits{A.C.}},
\bauthor{\bsnm{Lo}, \binits{W.-Y.}}, \betal:
\bctitle{Segment anything}.
In: \bbtitle{Proceedings of the IEEE/CVF International Conference on Computer Vision},
pp. \bfpage{4015}--\blpage{4026}
(\byear{2023})
\end{bchapter}
\endbibitem

\bibitem{hubner_evaluation_2020}
\begin{barticle}
\bauthor{\bsnm{H{\"u}bner}, \binits{P.}},
\bauthor{\bsnm{Clintworth}, \binits{K.}},
\bauthor{\bsnm{Liu}, \binits{Q.}},
\bauthor{\bsnm{Weinmann}, \binits{M.}},
\bauthor{\bsnm{Wursthorn}, \binits{S.}}:
\batitle{Evaluation of hololens tracking and depth sensing for indoor mapping applications}.
\bjtitle{Sensors}
\bvolume{20}(\bissue{4}),
\bfpage{1021}
(\byear{2020})
\end{barticle}
\endbibitem

\bibitem{yi_egolocate_2023}
\begin{barticle}
\bauthor{\bsnm{Yi}, \binits{X.}},
\bauthor{\bsnm{Zhou}, \binits{Y.}},
\bauthor{\bsnm{Habermann}, \binits{M.}},
\bauthor{\bsnm{Golyanik}, \binits{V.}},
\bauthor{\bsnm{Pan}, \binits{S.}},
\bauthor{\bsnm{Theobalt}, \binits{C.}},
\bauthor{\bsnm{Xu}, \binits{F.}}:
\batitle{Egolocate: Real-time motion capture, localization, and mapping with sparse body-mounted sensors}.
\bjtitle{ACM Transactions on Graphics (TOG)}
\bvolume{42}(\bissue{4}),
\bfpage{1}--\blpage{17}
(\byear{2023})
\end{barticle}
\endbibitem

\bibitem{rosinol_nerfslam_2023}
\begin{bchapter}
\bauthor{\bsnm{Rosinol}, \binits{A.}},
\bauthor{\bsnm{Leonard}, \binits{J.J.}},
\bauthor{\bsnm{Carlone}, \binits{L.}}:
\bctitle{Nerf-slam: Real-time dense monocular slam with neural radiance fields}.
In: \bbtitle{2023 IEEE/RSJ International Conference on Intelligent Robots and Systems (IROS)},
pp. \bfpage{3437}--\blpage{3444}
(\byear{2023}).
\bcomment{IEEE}
\end{bchapter}
\endbibitem

\bibitem{mildenhall_nerf_2021}
\begin{barticle}
\bauthor{\bsnm{Mildenhall}, \binits{B.}},
\bauthor{\bsnm{Srinivasan}, \binits{P.P.}},
\bauthor{\bsnm{Tancik}, \binits{M.}},
\bauthor{\bsnm{Barron}, \binits{J.T.}},
\bauthor{\bsnm{Ramamoorthi}, \binits{R.}},
\bauthor{\bsnm{Ng}, \binits{R.}}:
\batitle{Nerf: Representing scenes as neural radiance fields for view synthesis}.
\bjtitle{Communications of the ACM}
\bvolume{65}(\bissue{1}),
\bfpage{99}--\blpage{106}
(\byear{2021})
\end{barticle}
\endbibitem

\bibitem{yin_egohdm_2024}
\begin{barticle}
\bauthor{\bsnm{Yin}, \binits{H.}},
\bauthor{\bsnm{Liu}, \binits{B.}},
\bauthor{\bsnm{Kaufmann}, \binits{M.}},
\bauthor{\bsnm{He}, \binits{J.}},
\bauthor{\bsnm{Christen}, \binits{S.}},
\bauthor{\bsnm{Song}, \binits{J.}},
\bauthor{\bsnm{Hui}, \binits{P.}}:
\batitle{Egohdm: A real-time egocentric-inertial human motion capture, localization, and dense mapping system}.
\bjtitle{ACM Transactions on Graphics (TOG)}
\bvolume{43}(\bissue{6}),
\bfpage{1}--\blpage{12}
(\byear{2024})
\end{barticle}
\endbibitem

\bibitem{trumble_total_2017}
\begin{bchapter}
\bauthor{\bsnm{Trumble}, \binits{M.}},
\bauthor{\bsnm{Gilbert}, \binits{A.}},
\bauthor{\bsnm{Malleson}, \binits{C.}},
\bauthor{\bsnm{Hilton}, \binits{A.}},
\bauthor{\bsnm{Collomosse}, \binits{J.P.}}:
\bctitle{Total capture: 3d human pose estimation fusing video and inertial sensors.}
In: \bbtitle{BMVC},
vol. \bseriesno{2},
pp. \bfpage{1}--\blpage{13}
(\byear{2017}).
\bcomment{London, UK}
\end{bchapter}
\endbibitem

\bibitem{berton_rethinking_2022}
\begin{bchapter}
\bauthor{\bsnm{Berton}, \binits{G.}},
\bauthor{\bsnm{Masone}, \binits{C.}},
\bauthor{\bsnm{Caputo}, \binits{B.}}:
\bctitle{Rethinking visual geo-localization for large-scale applications}.
In: \bbtitle{Proceedings of the IEEE/CVF Conference on Computer Vision and Pattern Recognition},
pp. \bfpage{4878}--\blpage{4888}
(\byear{2022})
\end{bchapter}
\endbibitem

\bibitem{ali-bey_mixvpr_2023}
\begin{bchapter}
\bauthor{\bsnm{Ali-Bey}, \binits{A.}},
\bauthor{\bsnm{Chaib-Draa}, \binits{B.}},
\bauthor{\bsnm{Giguere}, \binits{P.}}:
\bctitle{Mixvpr: Feature mixing for visual place recognition}.
In: \bbtitle{Proceedings of the IEEE/CVF Winter Conference on Applications of Computer Vision},
pp. \bfpage{2998}--\blpage{3007}
(\byear{2023})
\end{bchapter}
\endbibitem

\bibitem{suveges_unsupervised_2025}
\begin{barticle}
\bauthor{\bsnm{Suveges}, \binits{T.}},
\bauthor{\bsnm{McKenna}, \binits{S.}}:
\batitle{Unsupervised mapping and semantic user localisation from first-person monocular video}.
\bjtitle{Pattern Recognition}
\bvolume{158},
\bfpage{110923}
(\byear{2025})
\end{barticle}
\endbibitem

\bibitem{huang_automatic_2024}
\begin{bchapter}
\bauthor{\bsnm{Huang}, \binits{Y.}},
\bauthor{\bsnm{A~Hassan}, \binits{M.}},
\bauthor{\bsnm{He}, \binits{J.}},
\bauthor{\bsnm{Higgins}, \binits{J.}},
\bauthor{\bsnm{McCrory}, \binits{M.}},
\bauthor{\bsnm{Eicher-Miller}, \binits{H.}},
\bauthor{\bsnm{Thomas}, \binits{J.G.}},
\bauthor{\bsnm{Sazonov}, \binits{E.}},
\bauthor{\bsnm{Zhu}, \binits{F.}}:
\bctitle{Automatic recognition of food ingestion environment from the aim-2 wearable sensor}.
In: \bbtitle{Proceedings of the IEEE/CVF Conference on Computer Vision and Pattern Recognition},
pp. \bfpage{3685}--\blpage{3694}
(\byear{2024})
\end{bchapter}
\endbibitem

\bibitem{blanton_extending_2020}
\begin{bchapter}
\bauthor{\bsnm{Blanton}, \binits{H.}},
\bauthor{\bsnm{Greenwell}, \binits{C.}},
\bauthor{\bsnm{Workman}, \binits{S.}},
\bauthor{\bsnm{Jacobs}, \binits{N.}}:
\bctitle{Extending absolute pose regression to multiple scenes}.
In: \bbtitle{Proceedings of the IEEE/CVF Conference on Computer Vision and Pattern Recognition Workshops},
pp. \bfpage{38}--\blpage{39}
(\byear{2020})
\end{bchapter}
\endbibitem

\bibitem{shavit_learning_2021}
\begin{bchapter}
\bauthor{\bsnm{Shavit}, \binits{Y.}},
\bauthor{\bsnm{Ferens}, \binits{R.}},
\bauthor{\bsnm{Keller}, \binits{Y.}}:
\bctitle{Learning multi-scene absolute pose regression with transformers}.
In: \bbtitle{Proceedings of the IEEE/CVF International Conference on Computer Vision},
pp. \bfpage{2733}--\blpage{2742}
(\byear{2021})
\end{bchapter}
\endbibitem

\bibitem{do_learning_2022}
\begin{bchapter}
\bauthor{\bsnm{Do}, \binits{T.}},
\bauthor{\bsnm{Miksik}, \binits{O.}},
\bauthor{\bsnm{DeGol}, \binits{J.}},
\bauthor{\bsnm{Park}, \binits{H.S.}},
\bauthor{\bsnm{Sinha}, \binits{S.N.}}:
\bctitle{Learning to detect scene landmarks for camera localization}.
In: \bbtitle{Proceedings of the IEEE/CVF Conference on Computer Vision and Pattern Recognition},
pp. \bfpage{11132}--\blpage{11142}
(\byear{2022})
\end{bchapter}
\endbibitem

\bibitem{panek_meshloc_2022}
\begin{bchapter}
\bauthor{\bsnm{Panek}, \binits{V.}},
\bauthor{\bsnm{Kukelova}, \binits{Z.}},
\bauthor{\bsnm{Sattler}, \binits{T.}}:
\bctitle{Meshloc: Mesh-based visual localization}.
In: \bbtitle{European Conference on Computer Vision},
pp. \bfpage{589}--\blpage{609}
(\byear{2022}).
\bcomment{Springer}
\end{bchapter}
\endbibitem

\bibitem{zhu_r2_2023}
\begin{bchapter}
\bauthor{\bsnm{Zhu}, \binits{S.}},
\bauthor{\bsnm{Yang}, \binits{L.}},
\bauthor{\bsnm{Chen}, \binits{C.}},
\bauthor{\bsnm{Shah}, \binits{M.}},
\bauthor{\bsnm{Shen}, \binits{X.}},
\bauthor{\bsnm{Wang}, \binits{H.}}:
\bctitle{R2former: Unified retrieval and reranking transformer for place recognition}.
In: \bbtitle{Proceedings of the IEEE/CVF Conference on Computer Vision and Pattern Recognition},
pp. \bfpage{19370}--\blpage{19380}
(\byear{2023})
\end{bchapter}
\endbibitem

\bibitem{lin_exploring_2024}
\begin{bchapter}
\bauthor{\bsnm{Lin}, \binits{H.}},
\bauthor{\bsnm{Long}, \binits{C.}},
\bauthor{\bsnm{Fei}, \binits{Y.}},
\bauthor{\bsnm{Xia}, \binits{Q.}},
\bauthor{\bsnm{Yin}, \binits{E.}},
\bauthor{\bsnm{Yin}, \binits{B.}},
\bauthor{\bsnm{Yang}, \binits{X.}}:
\bctitle{Exploring matching rates: From keypoint selection to camera relocalization}.
In: \bbtitle{Proceedings of the 32nd ACM International Conference on Multimedia},
pp. \bfpage{506}--\blpage{514}
(\byear{2024})
\end{bchapter}
\endbibitem

\bibitem{xiong2025adaptively}
\begin{barticle}
\bauthor{\bsnm{Xiong}, \binits{H.}},
\bauthor{\bsnm{Wang}, \binits{L.}},
\bauthor{\bsnm{Qiu}, \binits{H.}},
\bauthor{\bsnm{Zhao}, \binits{T.}},
\bauthor{\bsnm{Qiu}, \binits{B.}},
\bauthor{\bsnm{Li}, \binits{H.}}:
\batitle{Adaptively forget with crossmodal and textual distillation for class-incremental video captioning}.
\bjtitle{Neurocomputing}
\bvolume{624},
\bfpage{129388}
(\byear{2025})
\end{barticle}
\endbibitem

\bibitem{nagar_generating_2021}
\begin{barticle}
\bauthor{\bsnm{Nagar}, \binits{P.}},
\bauthor{\bsnm{Rathore}, \binits{A.}},
\bauthor{\bsnm{Jawahar}, \binits{C.}},
\bauthor{\bsnm{Arora}, \binits{C.}}:
\batitle{Generating personalized summaries of day long egocentric videos}.
\bjtitle{IEEE Transactions on Pattern Analysis and Machine Intelligence}
\bvolume{45}(\bissue{6}),
\bfpage{6832}--\blpage{6845}
(\byear{2021})
\end{barticle}
\endbibitem

\bibitem{elfeki_multistream_2022}
\begin{bchapter}
\bauthor{\bsnm{Elfeki}, \binits{M.}},
\bauthor{\bsnm{Wang}, \binits{L.}},
\bauthor{\bsnm{Borji}, \binits{A.}}:
\bctitle{Multi-stream dynamic video summarization}.
In: \bbtitle{Proceedings of the IEEE/CVF Winter Conference on Applications of Computer Vision},
pp. \bfpage{339}--\blpage{349}
(\byear{2022})
\end{bchapter}
\endbibitem

\bibitem{furlan_fast_2018}
\begin{bchapter}
\bauthor{\bsnm{Bajcsy}, \binits{R.}},
\bauthor{\bsnm{Nascimento}, \binits{E.R.}}, \betal:
\bctitle{Fast forwarding egocentric videos by listening and watching}.
In: \bbtitle{Proceedings of the IEEE Conference on Computer Vision and Pattern Recognition Workshops},
pp. \bfpage{2504}--\blpage{2507}
(\byear{2018})
\end{bchapter}
\endbibitem

\bibitem{ramos_personalizing_2020}
\begin{bchapter}
\bauthor{\bsnm{Ramos}, \binits{W.}},
\bauthor{\bsnm{Silva}, \binits{M.}},
\bauthor{\bsnm{Araujo}, \binits{E.}},
\bauthor{\bsnm{Neves}, \binits{A.}},
\bauthor{\bsnm{Nascimento}, \binits{E.}}:
\bctitle{Personalizing fast-forward videos based on visual and textual features from social network}.
In: \bbtitle{Proceedings of the IEEE/CVF Winter Conference on Applications of Computer Vision},
pp. \bfpage{3271}--\blpage{3280}
(\byear{2020})
\end{bchapter}
\endbibitem

\bibitem{wu_intentvizor_2022}
\begin{bchapter}
\bauthor{\bsnm{Wu}, \binits{G.}},
\bauthor{\bsnm{Lin}, \binits{J.}},
\bauthor{\bsnm{Silva}, \binits{C.T.}}:
\bctitle{Intentvizor: Towards generic query guided interactive video summarization}.
In: \bbtitle{Proceedings of the IEEE/CVF Conference on Computer Vision and Pattern Recognition},
pp. \bfpage{10503}--\blpage{10512}
(\byear{2022})
\end{bchapter}
\endbibitem

\bibitem{sahu_egocentric_2023}
\begin{barticle}
\bauthor{\bsnm{Sahu}, \binits{A.}},
\bauthor{\bsnm{Chowdhury}, \binits{A.S.}}:
\batitle{Egocentric video co-summarization using transfer learning and refined random walk on a constrained graph}.
\bjtitle{Pattern Recognition}
\bvolume{134},
\bfpage{109128}
(\byear{2023})
\end{barticle}
\endbibitem

\bibitem{dai_egocap_2024}
\begin{barticle}
\bauthor{\bsnm{Dai}, \binits{Z.}},
\bauthor{\bsnm{Tran}, \binits{V.}},
\bauthor{\bsnm{Markham}, \binits{A.}},
\bauthor{\bsnm{Trigoni}, \binits{N.}},
\bauthor{\bsnm{Rahman}, \binits{M.A.}},
\bauthor{\bsnm{Wijayasingha}, \binits{L.N.}},
\bauthor{\bsnm{Stankovic}, \binits{J.}},
\bauthor{\bsnm{Li}, \binits{C.}}:
\batitle{Egocap and egoformer: First-person image captioning with context fusion}.
\bjtitle{Pattern Recognition Letters}
\bvolume{181},
\bfpage{50}--\blpage{56}
(\byear{2024})
\end{barticle}
\endbibitem

\bibitem{qiu_egocentric_2024}
\begin{barticle}
\bauthor{\bsnm{Qiu}, \binits{J.}},
\bauthor{\bsnm{Lo}, \binits{F.P.-W.}},
\bauthor{\bsnm{Gu}, \binits{X.}},
\bauthor{\bsnm{Jobarteh}, \binits{M.L.}},
\bauthor{\bsnm{Jia}, \binits{W.}},
\bauthor{\bsnm{Baranowski}, \binits{T.}},
\bauthor{\bsnm{Steiner-Asiedu}, \binits{M.}},
\bauthor{\bsnm{Anderson}, \binits{A.K.}},
\bauthor{\bsnm{McCrory}, \binits{M.A.}},
\bauthor{\bsnm{Sazonov}, \binits{E.}}, \betal:
\batitle{Egocentric image captioning for privacy-preserved passive dietary intake monitoring}.
\bjtitle{IEEE Transactions on Cybernetics}
\bvolume{54}(\bissue{2}),
\bfpage{679}--\blpage{692}
(\byear{2023})
\end{barticle}
\endbibitem

\bibitem{parikh_echoguide_2024}
\begin{bchapter}
\bauthor{\bsnm{Parikh}, \binits{V.}},
\bauthor{\bsnm{Mahmud}, \binits{S.}},
\bauthor{\bsnm{Agarwal}, \binits{D.}},
\bauthor{\bsnm{Li}, \binits{K.}},
\bauthor{\bsnm{Guimbreti{\`e}re}, \binits{F.}},
\bauthor{\bsnm{Zhang}, \binits{C.}}:
\bctitle{Echoguide: Active acoustic guidance for llm-based eating event analysis from egocentric videos}.
In: \bbtitle{Proceedings of the 2024 ACM International Symposium on Wearable Computers},
pp. \bfpage{40}--\blpage{47}
(\byear{2024})
\end{bchapter}
\endbibitem

\bibitem{he_align_2023}
\begin{bchapter}
\bauthor{\bsnm{He}, \binits{B.}},
\bauthor{\bsnm{Wang}, \binits{J.}},
\bauthor{\bsnm{Qiu}, \binits{J.}},
\bauthor{\bsnm{Bui}, \binits{T.}},
\bauthor{\bsnm{Shrivastava}, \binits{A.}},
\bauthor{\bsnm{Wang}, \binits{Z.}}:
\bctitle{Align and attend: Multimodal summarization with dual contrastive losses}.
In: \bbtitle{Proceedings of the IEEE/CVF Conference on Computer Vision and Pattern Recognition},
pp. \bfpage{14867}--\blpage{14878}
(\byear{2023})
\end{bchapter}
\endbibitem

\bibitem{xu_retrievalaugmented_2024}
\begin{bchapter}
\bauthor{\bsnm{Xu}, \binits{J.}},
\bauthor{\bsnm{Huang}, \binits{Y.}},
\bauthor{\bsnm{Hou}, \binits{J.}},
\bauthor{\bsnm{Chen}, \binits{G.}},
\bauthor{\bsnm{Zhang}, \binits{Y.}},
\bauthor{\bsnm{Feng}, \binits{R.}},
\bauthor{\bsnm{Xie}, \binits{W.}}:
\bctitle{Retrieval-augmented egocentric video captioning}.
In: \bbtitle{Proceedings of the IEEE/CVF Conference on Computer Vision and Pattern Recognition},
pp. \bfpage{13525}--\blpage{13536}
(\byear{2024})
\end{bchapter}
\endbibitem

\bibitem{han_benchmarking_2024}
\begin{barticle}
\bauthor{\bsnm{Han}, \binits{R.}},
\bauthor{\bsnm{Feng}, \binits{W.}},
\bauthor{\bsnm{Wang}, \binits{F.}},
\bauthor{\bsnm{Qian}, \binits{Z.}},
\bauthor{\bsnm{Yan}, \binits{H.}},
\bauthor{\bsnm{Wang}, \binits{S.}}:
\batitle{Benchmarking the complementary-view multi-human association and tracking}.
\bjtitle{International Journal of Computer Vision}
\bvolume{132}(\bissue{1}),
\bfpage{118}--\blpage{136}
(\byear{2024})
\end{barticle}
\endbibitem

\bibitem{fanyang_yowo_2024}
\begin{botherref}
\oauthor{\bsnm{Yang}, \binits{F.}},
\oauthor{\bsnm{Yamao}, \binits{S.}},
\oauthor{\bsnm{Kusajima}, \binits{I.}},
\oauthor{\bsnm{Moteki}, \binits{A.}},
\oauthor{\bsnm{Masui}, \binits{S.}},
\oauthor{\bsnm{Jiang}, \binits{S.}}:
Yowo: You only walk once to jointly map an indoor scene and register ceiling-mounted cameras.
IEEE Transactions on Circuits and Systems for Video Technology
(2024)
\end{botherref}
\endbibitem

\bibitem{xue_learning_2023}
\begin{barticle}
\bauthor{\bsnm{Xue}, \binits{Z.S.}},
\bauthor{\bsnm{Grauman}, \binits{K.}}:
\batitle{Learning fine-grained view-invariant representations from unpaired ego-exo videos via temporal alignment}.
\bjtitle{Advances in Neural Information Processing Systems}
\bvolume{36},
\bfpage{53688}--\blpage{53710}
(\byear{2023})
\end{barticle}
\endbibitem

\bibitem{jang_intra_2024}
\begin{bchapter}
\bauthor{\bsnm{Jang}, \binits{J.H.}},
\bauthor{\bsnm{Seo}, \binits{H.}},
\bauthor{\bsnm{Chun}, \binits{S.Y.}}:
\bctitle{Intra: Interaction relationship-aware weakly supervised affordance grounding}.
In: \bbtitle{European Conference on Computer Vision},
pp. \bfpage{18}--\blpage{34}
(\byear{2024}).
\bcomment{Springer}
\end{bchapter}
\endbibitem

\bibitem{truong_crossview_2025}
\begin{barticle}
\bauthor{\bsnm{Truong}, \binits{T.-D.}},
\bauthor{\bsnm{Luu}, \binits{K.}}:
\batitle{Cross-view action recognition understanding from exocentric to egocentric perspective}.
\bjtitle{Neurocomputing}
\bvolume{614},
\bfpage{128731}
(\byear{2025})
\end{barticle}
\endbibitem

\bibitem{sigurdsson_actor_2018}
\begin{bchapter}
\bauthor{\bsnm{Sigurdsson}, \binits{G.A.}},
\bauthor{\bsnm{Gupta}, \binits{A.}},
\bauthor{\bsnm{Schmid}, \binits{C.}},
\bauthor{\bsnm{Farhadi}, \binits{A.}},
\bauthor{\bsnm{Alahari}, \binits{K.}}:
\bctitle{Actor and observer: Joint modeling of first and third-person videos}.
In: \bbtitle{Proceedings of the IEEE Conference on Computer Vision and Pattern Recognition},
pp. \bfpage{7396}--\blpage{7404}
(\byear{2018})
\end{bchapter}
\endbibitem

\bibitem{cheng_4diff_2024}
\begin{bchapter}
\bauthor{\bsnm{Cheng}, \binits{F.}},
\bauthor{\bsnm{Luo}, \binits{M.}},
\bauthor{\bsnm{Wang}, \binits{H.}},
\bauthor{\bsnm{Dimakis}, \binits{A.}},
\bauthor{\bsnm{Torresani}, \binits{L.}},
\bauthor{\bsnm{Bertasius}, \binits{G.}},
\bauthor{\bsnm{Grauman}, \binits{K.}}:
\bctitle{4diff: 3d-aware diffusion model for third-to-first viewpoint translation}.
In: \bbtitle{European Conference on Computer Vision},
pp. \bfpage{409}--\blpage{427}
(\byear{2024}).
\bcomment{Springer}
\end{bchapter}
\endbibitem

\bibitem{miluo_put_2024}
\begin{bchapter}
\bauthor{\bsnm{Luo}, \binits{M.}},
\bauthor{\bsnm{Xue}, \binits{Z.}},
\bauthor{\bsnm{Dimakis}, \binits{A.}},
\bauthor{\bsnm{Grauman}, \binits{K.}}:
\bctitle{Put myself in your shoes: Lifting the egocentric perspective from exocentric videos}.
In: \bbtitle{European Conference on Computer Vision},
pp. \bfpage{407}--\blpage{425}
(\byear{2024}).
\bcomment{Springer}
\end{bchapter}
\endbibitem

\bibitem{fan_egovqa_2019}
\begin{bchapter}
\bauthor{\bsnm{Fan}, \binits{C.}}:
\bctitle{Egovqa-an egocentric video question answering benchmark dataset}.
In: \bbtitle{Proceedings of the IEEE/CVF International Conference on Computer Vision Workshops},
pp. \bfpage{0}--\blpage{0}
(\byear{2019})
\end{bchapter}
\endbibitem

\bibitem{gao_envqa_2021}
\begin{bchapter}
\bauthor{\bsnm{Gao}, \binits{D.}},
\bauthor{\bsnm{Wang}, \binits{R.}},
\bauthor{\bsnm{Bai}, \binits{Z.}},
\bauthor{\bsnm{Chen}, \binits{X.}}:
\bctitle{Env-qa: A video question answering benchmark for comprehensive understanding of dynamic environments}.
In: \bbtitle{Proceedings of the IEEE/CVF International Conference on Computer Vision},
pp. \bfpage{1675}--\blpage{1685}
(\byear{2021})
\end{bchapter}
\endbibitem

\bibitem{jia_egotaskqa_2022}
\begin{barticle}
\bauthor{\bsnm{Jia}, \binits{B.}},
\bauthor{\bsnm{Lei}, \binits{T.}},
\bauthor{\bsnm{Zhu}, \binits{S.-C.}},
\bauthor{\bsnm{Huang}, \binits{S.}}:
\batitle{Egotaskqa: Understanding human tasks in egocentric videos}.
\bjtitle{Advances in Neural Information Processing Systems}
\bvolume{35},
\bfpage{3343}--\blpage{3360}
(\byear{2022})
\end{barticle}
\endbibitem

\bibitem{gao_mist_2023}
\begin{bchapter}
\bauthor{\bsnm{Gao}, \binits{D.}},
\bauthor{\bsnm{Zhou}, \binits{L.}},
\bauthor{\bsnm{Ji}, \binits{L.}},
\bauthor{\bsnm{Zhu}, \binits{L.}},
\bauthor{\bsnm{Yang}, \binits{Y.}},
\bauthor{\bsnm{Shou}, \binits{M.Z.}}:
\bctitle{Mist: Multi-modal iterative spatial-temporal transformer for long-form video question answering}.
In: \bbtitle{Proceedings of the IEEE/CVF Conference on Computer Vision and Pattern Recognition},
pp. \bfpage{14773}--\blpage{14783}
(\byear{2023})
\end{bchapter}
\endbibitem

\bibitem{goletto_amego_2024}
\begin{bchapter}
\bauthor{\bsnm{Goletto}, \binits{G.}},
\bauthor{\bsnm{Nagarajan}, \binits{T.}},
\bauthor{\bsnm{Averta}, \binits{G.}},
\bauthor{\bsnm{Damen}, \binits{D.}}:
\bctitle{Amego: Active memory from long egocentric videos}.
In: \bbtitle{European Conference on Computer Vision},
pp. \bfpage{92}--\blpage{110}
(\byear{2024}).
\bcomment{Springer}
\end{bchapter}
\endbibitem

\bibitem{yu_multitarget_2019}
\begin{bchapter}
\bauthor{\bsnm{Yu}, \binits{L.}},
\bauthor{\bsnm{Chen}, \binits{X.}},
\bauthor{\bsnm{Gkioxari}, \binits{G.}},
\bauthor{\bsnm{Bansal}, \binits{M.}},
\bauthor{\bsnm{Berg}, \binits{T.L.}},
\bauthor{\bsnm{Batra}, \binits{D.}}:
\bctitle{Multi-target embodied question answering}.
In: \bbtitle{Proceedings of the IEEE/CVF Conference on Computer Vision and Pattern Recognition},
pp. \bfpage{6309}--\blpage{6318}
(\byear{2019})
\end{bchapter}
\endbibitem

\bibitem{ma_sqa3d_2023}
\begin{botherref}
\oauthor{\bsnm{Ma}, \binits{X.}},
\oauthor{\bsnm{Yong}, \binits{S.}},
\oauthor{\bsnm{Zheng}, \binits{Z.}},
\oauthor{\bsnm{Li}, \binits{Q.}},
\oauthor{\bsnm{Liang}, \binits{Y.}},
\oauthor{\bsnm{Zhu}, \binits{S.-C.}},
\oauthor{\bsnm{Huang}, \binits{S.}}:
Sqa3d: Situated question answering in 3d scenes.
arXiv preprint arXiv:2210.07474
(2022)
\end{botherref}
\endbibitem

\bibitem{zhu_excalibur_2023}
\begin{bchapter}
\bauthor{\bsnm{Zhu}, \binits{H.}},
\bauthor{\bsnm{Kapoor}, \binits{R.}},
\bauthor{\bsnm{Min}, \binits{S.Y.}},
\bauthor{\bsnm{Han}, \binits{W.}},
\bauthor{\bsnm{Li}, \binits{J.}},
\bauthor{\bsnm{Geng}, \binits{K.}},
\bauthor{\bsnm{Neubig}, \binits{G.}},
\bauthor{\bsnm{Bisk}, \binits{Y.}},
\bauthor{\bsnm{Kembhavi}, \binits{A.}},
\bauthor{\bsnm{Weihs}, \binits{L.}}:
\bctitle{Excalibur: Encouraging and evaluating embodied exploration}.
In: \bbtitle{Proceedings of the IEEE/CVF Conference on Computer Vision and Pattern Recognition},
pp. \bfpage{14931}--\blpage{14942}
(\byear{2023})
\end{bchapter}
\endbibitem

\bibitem{wong_assistq_2022}
\begin{bchapter}
\bauthor{\bsnm{Wong}, \binits{B.}},
\bauthor{\bsnm{Chen}, \binits{J.}},
\bauthor{\bsnm{Wu}, \binits{Y.}},
\bauthor{\bsnm{Lei}, \binits{S.W.}},
\bauthor{\bsnm{Mao}, \binits{D.}},
\bauthor{\bsnm{Gao}, \binits{D.}},
\bauthor{\bsnm{Shou}, \binits{M.Z.}}:
\bctitle{Assistq: Affordance-centric question-driven task completion for egocentric assistant}.
In: \bbtitle{European Conference on Computer Vision},
pp. \bfpage{485}--\blpage{501}
(\byear{2022}).
\bcomment{Springer}
\end{bchapter}
\endbibitem

\bibitem{hummel_egocvr_2024}
\begin{bchapter}
\bauthor{\bsnm{Hummel}, \binits{T.}},
\bauthor{\bsnm{Karthik}, \binits{S.}},
\bauthor{\bsnm{Georgescu}, \binits{M.-I.}},
\bauthor{\bsnm{Akata}, \binits{Z.}}:
\bctitle{Egocvr: An egocentric benchmark for fine-grained composed video retrieval}.
In: \bbtitle{European Conference on Computer Vision},
pp. \bfpage{1}--\blpage{17}
(\byear{2024}).
\bcomment{Springer}
\end{bchapter}
\endbibitem

\bibitem{cheng_egothink_2024}
\begin{bchapter}
\bauthor{\bsnm{Cheng}, \binits{S.}},
\bauthor{\bsnm{Guo}, \binits{Z.}},
\bauthor{\bsnm{Wu}, \binits{J.}},
\bauthor{\bsnm{Fang}, \binits{K.}},
\bauthor{\bsnm{Li}, \binits{P.}},
\bauthor{\bsnm{Liu}, \binits{H.}},
\bauthor{\bsnm{Liu}, \binits{Y.}}:
\bctitle{Egothink: Evaluating first-person perspective thinking capability of vision-language models}.
In: \bbtitle{Proceedings of the IEEE/CVF Conference on Computer Vision and Pattern Recognition},
pp. \bfpage{14291}--\blpage{14302}
(\byear{2024})
\end{bchapter}
\endbibitem

\bibitem{radford_learning_2021}
\begin{bchapter}
\bauthor{\bsnm{Radford}, \binits{A.}},
\bauthor{\bsnm{Kim}, \binits{J.W.}},
\bauthor{\bsnm{Hallacy}, \binits{C.}},
\bauthor{\bsnm{Ramesh}, \binits{A.}},
\bauthor{\bsnm{Goh}, \binits{G.}},
\bauthor{\bsnm{Agarwal}, \binits{S.}},
\bauthor{\bsnm{Sastry}, \binits{G.}},
\bauthor{\bsnm{Askell}, \binits{A.}},
\bauthor{\bsnm{Mishkin}, \binits{P.}},
\bauthor{\bsnm{Clark}, \binits{J.}}, \betal:
\bctitle{Learning transferable visual models from natural language supervision}.
In: \bbtitle{International Conference on Machine Learning},
pp. \bfpage{8748}--\blpage{8763}
(\byear{2021}).
\bcomment{PMLR}
\end{bchapter}
\endbibitem

\bibitem{tong_videomae_2022}
\begin{barticle}
\bauthor{\bsnm{Tong}, \binits{Z.}},
\bauthor{\bsnm{Song}, \binits{Y.}},
\bauthor{\bsnm{Wang}, \binits{J.}},
\bauthor{\bsnm{Wang}, \binits{L.}}:
\batitle{Videomae: Masked autoencoders are data-efficient learners for self-supervised video pre-training}.
\bjtitle{Advances in neural information processing systems}
\bvolume{35},
\bfpage{10078}--\blpage{10093}
(\byear{2022})
\end{barticle}
\endbibitem

\bibitem{pramanick_egovlpv2_2023}
\begin{bchapter}
\bauthor{\bsnm{Pramanick}, \binits{S.}},
\bauthor{\bsnm{Song}, \binits{Y.}},
\bauthor{\bsnm{Nag}, \binits{S.}},
\bauthor{\bsnm{Lin}, \binits{K.Q.}},
\bauthor{\bsnm{Shah}, \binits{H.}},
\bauthor{\bsnm{Shou}, \binits{M.Z.}},
\bauthor{\bsnm{Chellappa}, \binits{R.}},
\bauthor{\bsnm{Zhang}, \binits{P.}}:
\bctitle{Egovlpv2: Egocentric video-language pre-training with fusion in the backbone}.
In: \bbtitle{Proceedings of the IEEE/CVF International Conference on Computer Vision},
pp. \bfpage{5285}--\blpage{5297}
(\byear{2023})
\end{bchapter}
\endbibitem

\bibitem{pirsiavash_detecting_2012}
\begin{bchapter}
\bauthor{\bsnm{Pirsiavash}, \binits{H.}},
\bauthor{\bsnm{Ramanan}, \binits{D.}}:
\bctitle{Detecting activities of daily living in first-person camera views}.
In: \bbtitle{2012 IEEE Conference on Computer Vision and Pattern Recognition},
pp. \bfpage{2847}--\blpage{2854}
(\byear{2012}).
\bcomment{IEEE}
\end{bchapter}
\endbibitem

\bibitem{liu_hoi4d_2022}
\begin{bchapter}
\bauthor{\bsnm{Liu}, \binits{Y.}},
\bauthor{\bsnm{Liu}, \binits{Y.}},
\bauthor{\bsnm{Jiang}, \binits{C.}},
\bauthor{\bsnm{Lyu}, \binits{K.}},
\bauthor{\bsnm{Wan}, \binits{W.}},
\bauthor{\bsnm{Shen}, \binits{H.}},
\bauthor{\bsnm{Liang}, \binits{B.}},
\bauthor{\bsnm{Fu}, \binits{Z.}},
\bauthor{\bsnm{Wang}, \binits{H.}},
\bauthor{\bsnm{Yi}, \binits{L.}}:
\bctitle{Hoi4d: A 4d egocentric dataset for category-level human-object interaction}.
In: \bbtitle{Proceedings of the IEEE/CVF Conference on Computer Vision and Pattern Recognition},
pp. \bfpage{21013}--\blpage{21022}
(\byear{2022})
\end{bchapter}
\endbibitem

\bibitem{bar_egopet_2024}
\begin{bchapter}
\bauthor{\bsnm{Bar}, \binits{A.}},
\bauthor{\bsnm{Bakhtiar}, \binits{A.}},
\bauthor{\bsnm{Tran}, \binits{D.}},
\bauthor{\bsnm{Loquercio}, \binits{A.}},
\bauthor{\bsnm{Rajasegaran}, \binits{J.}},
\bauthor{\bsnm{LeCun}, \binits{Y.}},
\bauthor{\bsnm{Globerson}, \binits{A.}},
\bauthor{\bsnm{Darrell}, \binits{T.}}:
\bctitle{Egopet: Egomotion and interaction data from an animal's perspective}.
In: \bbtitle{Proceedings of the European Conference on Computer Vision (ECCV)},
pp. \bfpage{377}--\blpage{394}
(\byear{2024}).
\bcomment{Springer Nature Switzerland}
\end{bchapter}
\endbibitem

\bibitem{bock_wear_2024}
\begin{barticle}
\bauthor{\bsnm{Bock}, \binits{M.}},
\bauthor{\bsnm{Kuehne}, \binits{H.}},
\bauthor{\bsnm{Van~Laerhoven}, \binits{K.}},
\bauthor{\bsnm{Moeller}, \binits{M.}}:
\batitle{Wear: An outdoor sports dataset for wearable and egocentric activity recognition}.
\bjtitle{Proceedings of the ACM on Interactive, Mobile, Wearable and Ubiquitous Technologies}
\bvolume{8}(\bissue{4}),
\bfpage{1}--\blpage{21}
(\byear{2024})
\end{barticle}
\endbibitem

\bibitem{grauman_egoexo4d_2024}
\begin{bchapter}
\bauthor{\bsnm{Grauman}, \binits{K.}},
\bauthor{\bsnm{Westbury}, \binits{A.}},
\bauthor{\bsnm{Torresani}, \binits{L.}},
\bauthor{\bsnm{Kitani}, \binits{K.}},
\bauthor{\bsnm{Malik}, \binits{J.}},
\bauthor{\bsnm{Afouras}, \binits{T.}},
\bauthor{\bsnm{Ashutosh}, \binits{K.}},
\bauthor{\bsnm{Baiyya}, \binits{V.}},
\bauthor{\bsnm{Bansal}, \binits{S.}},
\bauthor{\bsnm{Boote}, \binits{B.}}, \betal:
\bctitle{Ego-exo4d: Understanding skilled human activity from first-and third-person perspectives}.
In: \bbtitle{Proceedings of the IEEE/CVF Conference on Computer Vision and Pattern Recognition},
pp. \bfpage{19383}--\blpage{19400}
(\byear{2024})
\end{bchapter}
\endbibitem

\bibitem{huang_egoexolearn_2024}
\begin{bchapter}
\bauthor{\bsnm{Huang}, \binits{Y.}},
\bauthor{\bsnm{Chen}, \binits{G.}},
\bauthor{\bsnm{Xu}, \binits{J.}},
\bauthor{\bsnm{Zhang}, \binits{M.}},
\bauthor{\bsnm{Yang}, \binits{L.}},
\bauthor{\bsnm{Pei}, \binits{B.}},
\bauthor{\bsnm{Zhang}, \binits{H.}},
\bauthor{\bsnm{Dong}, \binits{L.}},
\bauthor{\bsnm{Wang}, \binits{Y.}},
\bauthor{\bsnm{Wang}, \binits{L.}}, \betal:
\bctitle{Egoexolearn: A dataset for bridging asynchronous ego-and exo-centric view of procedural activities in real world}.
In: \bbtitle{Proceedings of the IEEE/CVF Conference on Computer Vision and Pattern Recognition},
pp. \bfpage{22072}--\blpage{22086}
(\byear{2024})
\end{bchapter}
\endbibitem

\bibitem{li_egoexofitness_2024}
\begin{bchapter}
\bauthor{\bsnm{Li}, \binits{Y.-M.}},
\bauthor{\bsnm{Huang}, \binits{W.-J.}},
\bauthor{\bsnm{Wang}, \binits{A.-L.}},
\bauthor{\bsnm{Zeng}, \binits{L.-A.}},
\bauthor{\bsnm{Meng}, \binits{J.-K.}},
\bauthor{\bsnm{Zheng}, \binits{W.-S.}}:
\bctitle{Egoexo-fitness: towards egocentric and exocentric full-body action understanding}.
In: \bbtitle{European Conference on Computer Vision},
pp. \bfpage{363}--\blpage{382}
(\byear{2024}).
\bcomment{Springer}
\end{bchapter}
\endbibitem

\bibitem{liu_taco_2024}
\begin{bchapter}
\bauthor{\bsnm{Liu}, \binits{Y.}},
\bauthor{\bsnm{Yang}, \binits{H.}},
\bauthor{\bsnm{Si}, \binits{X.}},
\bauthor{\bsnm{Liu}, \binits{L.}},
\bauthor{\bsnm{Li}, \binits{Z.}},
\bauthor{\bsnm{Zhang}, \binits{Y.}},
\bauthor{\bsnm{Liu}, \binits{Y.}},
\bauthor{\bsnm{Yi}, \binits{L.}}:
\bctitle{Taco: Benchmarking generalizable bimanual tool-action-object understanding}.
In: \bbtitle{Proceedings of the IEEE/CVF Conference on Computer Vision and Pattern Recognition},
pp. \bfpage{21740}--\blpage{21751}
(\byear{2024})
\end{bchapter}
\endbibitem

\bibitem{ma_nymeria_2024}
\begin{bchapter}
\bauthor{\bsnm{Ma}, \binits{L.}},
\bauthor{\bsnm{Ye}, \binits{Y.}},
\bauthor{\bsnm{Hong}, \binits{F.}},
\bauthor{\bsnm{Guzov}, \binits{V.}},
\bauthor{\bsnm{Jiang}, \binits{Y.}},
\bauthor{\bsnm{Postyeni}, \binits{R.}},
\bauthor{\bsnm{Pesqueira}, \binits{L.}},
\bauthor{\bsnm{Gamino}, \binits{A.}},
\bauthor{\bsnm{Baiyya}, \binits{V.}},
\bauthor{\bsnm{Kim}, \binits{H.J.}}, \betal:
\bctitle{Nymeria: A massive collection of multimodal egocentric daily motion in the wild}.
In: \bbtitle{European Conference on Computer Vision},
pp. \bfpage{445}--\blpage{465}
(\byear{2024}).
\bcomment{Springer}
\end{bchapter}
\endbibitem

\bibitem{schoonbeek_industreal_2024}
\begin{bchapter}
\bauthor{\bsnm{Schoonbeek}, \binits{T.J.}},
\bauthor{\bsnm{Houben}, \binits{T.}},
\bauthor{\bsnm{Onvlee}, \binits{H.}},
\bauthor{\bparticle{Van~der} \bsnm{Sommen}, \binits{F.}}, \betal:
\bctitle{Industreal: A dataset for procedure step recognition handling execution errors in egocentric videos in an industrial-like setting}.
In: \bbtitle{Proceedings of the IEEE/CVF Winter Conference on Applications of Computer Vision},
pp. \bfpage{4365}--\blpage{4374}
(\byear{2024})
\end{bchapter}
\endbibitem

\bibitem{tang_egotracks_2024}
\begin{botherref}
\oauthor{\bsnm{Tang}, \binits{H.}},
\oauthor{\bsnm{Liang}, \binits{K.J.}},
\oauthor{\bsnm{Grauman}, \binits{K.}},
\oauthor{\bsnm{Feiszli}, \binits{M.}},
\oauthor{\bsnm{Wang}, \binits{W.}}:
Egotracks: A long-term egocentric visual object tracking dataset.
Advances in Neural Information Processing Systems
\textbf{36}
(2024)
\end{botherref}
\endbibitem

\bibitem{qiu_egome_2025}
\begin{botherref}
\oauthor{\bsnm{Qiu}, \binits{H.}},
\oauthor{\bsnm{Shi}, \binits{Z.}},
\oauthor{\bsnm{Wang}, \binits{L.}},
\oauthor{\bsnm{Xiong}, \binits{H.}},
\oauthor{\bsnm{Li}, \binits{X.}},
\oauthor{\bsnm{Li}, \binits{H.}}:
Egome: Follow me via egocentric view in real world.
arXiv preprint arXiv:2501.19061
(2025)
\end{botherref}
\endbibitem

\bibitem{luo_grounded_2024}
\begin{barticle}
\bauthor{\bsnm{Luo}, \binits{H.}},
\bauthor{\bsnm{Zhai}, \binits{W.}},
\bauthor{\bsnm{Zhang}, \binits{J.}},
\bauthor{\bsnm{Cao}, \binits{Y.}},
\bauthor{\bsnm{Tao}, \binits{D.}}:
\batitle{Grounded affordance from exocentric view}.
\bjtitle{International Journal of Computer Vision}
\bvolume{132}(\bissue{6}),
\bfpage{1945}--\blpage{1969}
(\byear{2024})
\end{barticle}
\endbibitem

\bibitem{dai_hisc4d_2024}
\begin{botherref}
\oauthor{\bsnm{Dai}, \binits{Y.}},
\oauthor{\bsnm{Wang}, \binits{Z.}},
\oauthor{\bsnm{Lin}, \binits{X.}},
\oauthor{\bsnm{Wen}, \binits{C.}},
\oauthor{\bsnm{Xu}, \binits{L.}},
\oauthor{\bsnm{Shen}, \binits{S.}},
\oauthor{\bsnm{Ma}, \binits{Y.}},
\oauthor{\bsnm{Wang}, \binits{C.}}:
Hisc4d: Human-centered interaction and 4d scene capture in large-scale space using wearable imus and lidar.
IEEE Transactions on Pattern Analysis and Machine Intelligence
(2024)
\end{botherref}
\endbibitem

\bibitem{he_masked_2022}
\begin{bchapter}
\bauthor{\bsnm{He}, \binits{K.}},
\bauthor{\bsnm{Chen}, \binits{X.}},
\bauthor{\bsnm{Xie}, \binits{S.}},
\bauthor{\bsnm{Li}, \binits{Y.}},
\bauthor{\bsnm{Doll{\'a}r}, \binits{P.}},
\bauthor{\bsnm{Girshick}, \binits{R.}}:
\bctitle{Masked autoencoders are scalable vision learners}.
In: \bbtitle{Proceedings of the IEEE/CVF Conference on Computer Vision and Pattern Recognition},
pp. \bfpage{16000}--\blpage{16009}
(\byear{2022})
\end{bchapter}
\endbibitem

\end{thebibliography}

\end{document}